\documentclass[12pt]{article}
\usepackage{arxiv}

\usepackage[utf8]{inputenc} % allow utf-8 input
\usepackage[T1]{fontenc}    % use 8-bit T1 fonts
\usepackage{hyperref}       % hyperlinks
\usepackage{url}            % simple URL typesetting
\usepackage{booktabs}       % professional-quality tables
\usepackage{amsfonts}       % blackboard math symbols
\usepackage{nicefrac}       % compact symbols for 1/2, etc.
\usepackage{microtype}      % microtypography
\usepackage{lipsum}		% Can be removed after putting your text content
\usepackage{graphicx}
\usepackage{natbib}
\usepackage{doi}

\usepackage{url}
\usepackage{xcolor}

\usepackage{hyperref}

\definecolor{newcolor}{rgb}{.8,.349,.1}

\usepackage{float}

\usepackage{algorithmic}
\usepackage[ruled,vlined]{algorithm2e}

\SetCommentSty{mycommfont}

\SetKwInput{KwInput}{Input}
\SetKwInput{KwOutput}{Output}

\usepackage{amsmath}

\usepackage[justification=centering]{caption}

\newtheorem{definition}{Definition}

\title{Dilated Filters for Edge Detection algorithms}

\author{ \href{https://orcid.org/0000-0002-0071-958X}{\includegraphics[scale=0.06]{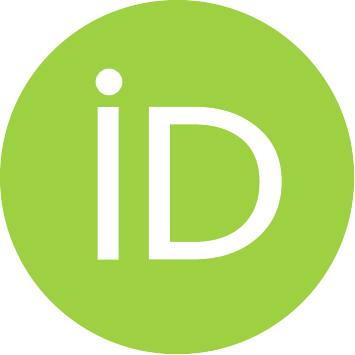}\hspace{1mm}Ciprian ORHEI} \\
	Politehnica University of Timi\c{s}oara, \\ Timi\c{s}oara, RO-300223, Romania. \\
	\texttt{ciprian.orhei@cm.upt.ro} \\
	\And
	\hspace{1mm}Victor BOGDAN \\
	West University of Timi\c{s}oara \\
    Timi\c{s}oara, RO-300223, Romania.\\
	\texttt{victor.bogdan97@e-uvt.ro} \\
	\AND
	\href{https://orcid.org/0000-0001-6660-282X}{\includegraphics[scale=0.06]{orcid.pdf}\hspace{1mm}Cosmin BONCHI\c{S}} \\
	West University of Timi\c{s}oara and \\
	The eAustria Research Institute, \\
    Bd. V. P\^arvan 4, 045B, \\
    Timi\c{s}oara, RO-300223, Romania.\\
	\texttt{cosmin.bonchis@e-uvt.ro} \\
}

\renewcommand{\shorttitle}{Dilated Filters for Edge Detection algorithms}

\hypersetup{
pdftitle={Dilated Filters for Edge Detection algorithms},
pdfsubject={cs.CV, eess.IV},
pdfauthor={Ciprian Orhei \textit{et~al.}},
pdfkeywords={Dilated filters, Edge detection operator, Edge detection, First order edge detection, Canny algorithm, Laplace algorithm, Laplace of Gaussian, Marr–Hildreth algorithm, Shen-Castan algorithm, Edge Drawing},
}

\begin{document}
\maketitle

\begin{abstract}
	Edges are a basic and fundamental feature in image processing, that are used directly or indirectly in huge amount of applications. Inspired by the expansion of image resolution and processing power dilated convolution techniques appeared. Dilated convolution have impressive results in machine learning, we discuss here the idea of dilating the standard filters which are used in edge detection algorithms. In this work we try to put together all our previous and current results by using instead of the classical convolution filters a dilated one. We compare the results of the edge detection algorithms using the proposed dilation filters with original filters or custom variants. Experimental results confirm our statement that dilation of filters have positive impact for edge detection algorithms from simple to rather complex algorithms.
\end{abstract}

% keywords can be removed

\section{Introduction}
An edge in an image is the most basic feature and has been intensively researched over time. A huge variety of mathematical methods have been used to identify points in which the image brightness changes sharply or has discontinuities. Edge detection is one low-level technique that is used for the goal of objects boundary detection. This is a fundamental tool in image processing, image analysis, machine vision and computer vision, particularly in the areas of feature detection and feature extraction. 

There are humongous approaches to basic feature detection depending on the pixel properties of the image. The methods have been defined from the gray scale levels to color slicing or from local features (lines, shapes) to the global matching features (the shape of objects, meta object property). The standard local edge detection filters are built for highlighting intensity change boundaries in the near neighborhood image regions. As also previous have been mentioned \cite{SEN2010}, the problem of edge finding has no universally accepted technique and this is a motivation for ongoing researching how to improve the edge detection methods.

The most successful edge detection algorithms have considered local methods, where the closest neighbourhood of a pixel is considered to be important for the pixel itself. Nowadays, images are containing more information than in the past (due to the sensors technologies) and we could say that the pixel itself is not similar only to its direct neighbours, but a bigger neighbourhood could be important. From early morphological edge detection operators definition \cite{haralick1987morphological} the extension idea was well presented. From another perspective, the dilated convolution methods have been recently proven very beneficial in many highly cited computer-vision papers: for small objects detection \cite{DilatedML2018}, in dense prediction tasks \cite{DilationML22018, DilationML2019}, on prediction without losing resolution \cite{DilationML2016}, for feature classifications in time series \cite{DilationML2019} or beneficial for context aggregation \cite{DilationML2017}.  

Merging those ideas we propose not just to increase the filter size but to simply \textbf{dilate} standard edge detection filters in order to characterize the pixel itself through \textbf{non-direct neighbour} properties. We define the dilation operation for filters that consists by simply adding gapes in the well known classical filters. This method of dilation is neither in the mathematical morphological sense \cite{haralick1987morphological}, nor the geometric extension of the kernels discussed in the literature \cite{Dilatetion2020}.

In our previous work we evaluated the dilation effect on classical first order edge detection algorithms and the classical Canny \cite{Canny1986} algorithm followed, naturally, with an analysis on which approach is better: to expand from lower level or dilate. \cite{Dilatetion2020, DilateionVsExpansion2020}. In this paper we desire to extend our work to other edge detection algorithm that have as a defining step the use of edge detection kernels. The classical edge detection methods Sobel \cite{SobelOriginal1973}, Prewitt \cite{PrewittOriginal1970}, Scharr \cite{Scharr2000}, together with  more complex boundary detection algorithms like Canny \cite{Canny1986}, Marr-Hildreth \cite{marr1980theory} or Shen-Castan \cite{shen1990} are considered for testing with our proposed dilation technique.

To evaluate the dilation benefits over an edge detection filter, we will not limit our analysis to the first order derivative gradient-based edge detection filters but consider the second order filter too. Details regarding the algorithms and steps we used in order to obtain the same edge map format are presented in Section \ref{Sec:preliminaries}. Section \ref{Sec:simulation_results} highlights the results of our hypothesis regarding the dilated filters, rather than expanding them. For a better comparison of the results, we used the challenging BSDS500 boundary detection benchmark tool and image sets from \cite{Bsds2011}. 

From the experiments of our previous work and from this analysis it is clear that in most cases the dilated approach brought benefits to the resulting edge map, idea that is elaborated in Section \ref{sec:conclusions}.

\section{Dilated filters}
\label{Sec:dilated filters}

In our previous work \cite{Dilatetion2020} we explored the idea of dilating the filter, kernels, of edge operators to obtain better edge-maps. In order to benefit from a higher neighbourhood of a pixels to obtain a pixel edge we define dilated filter as in Definition \ref{def:delated_filter}. When we dilate the kernels, we are considering the newly added positions as gaps and we ignore them by setting zeros.

\begin{definition}\label{def:delated_filter}
A \textit{dilated filter} is obtained by expanding the original filter by a dilation factor/size \cite{Dilatetion2020}. 
\end{definition}

By dilating the kernels we increase the distance between important pixels. We consider that this new distance will positively influence the result of the convolution. The bigger region of interest resulted can translate into stronger intensity changes in the image. In order to highlight our definition on a filter, we will use a generic kernel and represent the dilation in Figure \ref{fig:dilaton_example}.

% \begin{figure}[!h]
% \centering
% \small{
%     \begin{tabular}{cccc}
%             $\begin{bmatrix}
%             1&0&-1\\
%             2&0&-2\\
%             1&0&-1\\
%             \end{bmatrix}$
%         &
%             $\begin{bmatrix}
%             1&0&0&0&-1\\
%             0&0&0&0&0\\
%             2&0&0&0&-2\\
%             0&0&0&0&0\\
%             1&0&0&0&-1\\
%             \end{bmatrix}$
%         &
%             $\begin{bmatrix}
%             1&0&0&0&0&0&-1\\
%             0&0&0&0&0&0&0\\
%             0&0&0&0&0&0&0\\
%             2&0&0&0&0&0&-2\\
%             0&0&0&0&0&0&0\\
%             0&0&0&0&0&0&0\\
%             1&0&0&0&0&0&-1\\
%             \end{bmatrix}$
%     \\
%     \\
%             $ factor = 0$
%         &
%             $ factor = 1$
%         &
%             $ factor = 2$        
%     \end{tabular}
% }
% \caption{Sobel kernel dilated with a factor}
% \label{fig:dilaton_example}
% \end{figure}

\begin{figure}[!h]
% \begin{minipage}{.3\textwidth}
\centering
\small{
    \begin{tabular}{cccc}
            $\begin{bmatrix}
            a&b&c\\
            d&e&f\\
            g&h&i\\
            \end{bmatrix}$
        &
            $\begin{bmatrix}
            a&0&b&0&c\\
            0&0&0&0&0\\
            d&0&e&0&f\\
            0&0&0&0&0\\
            g&0&h&0&i\\
            \end{bmatrix}$
        &
            $\begin{bmatrix}
            a&0&0&b&0&0&c\\
            0&0&0&0&0&0&0\\
            0&0&0&0&0&0&0\\
            d&0&0&e&0&0&f\\
            0&0&0&0&0&0&0\\
            0&0&0&0&0&0&0\\
            g&0&0&h&0&0&i\\
            \end{bmatrix}$
    \\
    \\
            $ factor = 0$
        &
            $ factor = 1$
        &
            $ factor = 2$        
    \end{tabular}
}
\caption{General kernel dilated with a factor}
\label{fig:dilaton_example}
% \end{minipage}
\end{figure}

Dilating the filters, rather than extending them, helps in finding more edge pixels than the standard or extended variants of the filters. Another benefit worth mentioning of dilating is the fact that the number of operations does not increase with the dilating factor resulting in the same time cost for the edge detection \cite{Dilatetion2020}.

%\subsection{Expanding versus dilation}

Another considered approach was presented in \cite{DilateionVsExpansion2020}, where we compare and analyze the dilation of filters defined in \cite{Dilatetion2020} with the reconstruction from a lower scale pyramid level. Feature extraction in lower pyramid scale level is a common practice in the domain because of the benefits of lower computation resources which are needed.

The resulting edge map from a dilated $3x3$ filter is equivalent with an edge map calculated in a lower scale pyramid level and expanded back to original size. Dilating a factor of one is similar with applying the same filter in the immediately lower scale pyramid level. Dilating with a factor of two is similar with applying the filter in two scales lower in pyramid level and so on.This hypothesis stands because in both cases the region we take in consideration to find edges is not anymore an $3x3$ matrix but a $5x5$ matrix.

We have examined the equivalence between lower levels processing and dilating in our previous work \cite{DilateionVsExpansion2020}. As expected we obtain similar results when dilating as processing in lower levels. Dilating bring forward benefits regarding the neighborhood we consider and computation time but extracting features in lower levels has benefits of it's on.

The goal of this work is to present the comparison of the results of the \textbf{dilated} filters with \textbf{original} filters results. In the next subsections we will present in short  classical operators and algorithms for edge detection, which we used to show the results of our dilation technique.

\section{Preliminaries}
\label{Sec:preliminaries}

In this section we will present all the edge detection algorithms that are considered in our analysis. We will present the classical steps of each approach and the considered kernels in each case. In all algorithms the input images are converted to gray-scale images. The presentation order of the algorithms is in chronological order.

In order to benchmark our results with BSDS500 \cite{Bsds2011} all the edge maps resulted need to have the values of the pixels between $0$ and $255$ and thickness of $1$ pixel. To achieve this, we will use different techniques for first order gradient operators based algorithms than second order gradient based. Different approaches are needed because of the different raw edge maps that result. All the necessary steps are described in each subsection in the pseudo-code of the corresponding algorithm. In some of the cases multiple algorithms can use the same algorithm to obtain the desired edge map.

\subsection{First Order Derivative Orthogonal Gradient operators}
\label{Sec:preliminari_first_ortho}

First Order Derivative Orthogonal Gradient Operators are the most basic operators and have been extensively  researched over the decades. We will consider in our analysis the following edge detected operators and their extensions: \textbf{Pixel difference operator}\cite{Mlsna2009gradient}, \textbf{Separated pixel difference operator} \cite{Mlsna2009gradient}, \textbf{Sobel operator} \cite{SobelOriginal1973} and the extension to a 5x5 or 7x7 kernel \cite{BandaiSobelExtension2003, SobelExpasion2008, KekreSobelExtend2010, LevkineSobelPrewitScharrExtend2012, GuptaSobelExtension2013, SobelExpasion2008}, \textbf{Prewitt operator} \cite{PrewittOriginal1970} and extension to a 5x5 or 7x7 kernel \cite{SobelExpasion2008, LevkineSobelPrewitScharrExtend2012, SobelExpasion2008}, \textbf{Kirsch operator} \cite{Kirsch1971} and the 5x5 kernel expansion \cite{BandaiSobelExtension2003}, \textbf{Kitchen and Malin Operator} \cite{Kitchen1989}, \textbf{Kayalli Operator} \cite{Kayyali2000}, \textbf{Scharr Operator} \cite{Scharr2000} and the extensions to 5x5 kernel \cite{LevkineSobelPrewitScharrExtend2012, ScharrExtendedChen2017}, \textbf{Kroon Operator} \cite{Kroon2009}, \textbf{Orhei Operator} \cite{Orhei2020}. 

All the kernel masks for the operators are presented in Figure \ref{fig:k3_kernel_masks}, Figure \ref{fig:k5_kernel_masks} and, Figure \ref{fig:k7_kernel_masks} from Appendix \ref{appendix:A}. All those operators are orthogonal discrete isotropic filters so we are going to represent only one of the kernel. To obtain the other kernels we just need to rotate it by a fraction of $\frac{\pi}{2}$.

The gradient is a measure of change in a function and an image can be considered to be an array of samples of some continuous function of image intensity, typically  two-dimensional equivalent of first derivative. The magnitude is calculated using Equation \ref{formula:magnitude_and_aprox}, where $f(x,y)$ is the image and $G_x$, $G_y$ are the components on $x$ and $y$ axis. The Direction of the gradient is calculated using Equation \ref{formula:magnitude_orientation} \cite{haralick1992computer}.

\begin{align}\label{formula:magnitude_and_aprox}
        G[f(x,y)] = \sqrt{G_x^2 + G_y^2} \approx |G_x| + |G_y|
\end{align}
\begin{align}\label{formula:magnitude_orientation}
        \theta = \tan^{-1} \left[\frac{G_x}{G_y}\right]
\end{align}

The result of this algorithm is an edge-map formed by edges that are not topically $1$ pixel width, this aspect can deform the result of our evaluation. So for a better evaluation we choose to thin the resulting edges before hand. To this algorithm we would like to add two second steps that are commonly used, see \cite{Woods2011}: smoothing and thresholding. Smoothing of images is a common practice for enhancing the results as thresholding will eliminate "weak" edges that are found. Steps of our proposed algorithm that will be evaluated are detailed in Algorithm \ref{algorithm:steps_first_order}. 

\begin{algorithm}[!h]
\DontPrintSemicolon
  
  \KwInput{RGB image}
  \KwOutput{Binary edge map}
  \KwData{Testing set BSDS500\cite{Bsds2011}}
  
  \For{edge operator}{
    Convert image to gray-scale image.
    
    \tcc{Needed to enhance the final edge-map}
    Apply Gaussian filter smoothing
    
    \tcc{For gradient magnitude with orthogonal discrete isotropic rotated with $\pi/2$, found in Figure \ref{fig:k3_kernel_masks}}
    \tcc{For compass magnitude with orthogonal discrete isotropic rotate it by a fraction of ${\pi}$, found in Figure \ref{fig:compass_kernel_masks}}
    \tcc{For Frei-Chen with orthogonal discrete isotropic found in Figure \ref{fig:frei_kernel_masks}}
    Apply convolution with kernels
    
    \tcc{For gradient magnitude applying the edge detection operator using Equation \ref{formula:magnitude_and_aprox}}
    \tcc{For compass gradient magnitude applying the edge detection operator using Equation \ref{formula:magnitude_directional}}
    \tcc{For Frei-Chen applying the edge detection operator using Equation \ref{formula:frei_edge} and Equation \ref{formula:frei_line}}
    Calculate gradient magnitude
    
    \tcc{Use global threshold so each pixel, which has an intensity value higher or equal to a \textit{threshold}, will have its value set to a max value(e.g. 255) else to 0}
    Apply global threshold algorithm
    
    \tcc{Apply the Gua-Hall thinning algorithm \cite{GuoHall1989} to remove the excess edge points caused by the convolution and threshold}
    Apply thinning algorithm
  }
\caption{First Order Derivative Operators steps}
\label{algorithm:steps_first_order}
\end{algorithm}

\subsection{First Order Derivative Compass Gradient operators}
\label{Sec:preliminari_first_compass}

Compass gradient operators are commonly used in the edge detection and usually detect the influence of the neighbour pixels in a compass rotating directional components. They are commonly used as a alternative for the  Orthogonal Gradient Operators.

The gradient magnitude is calculated using Equation \ref{formula:magnitude_directional}, where $k$ is the number of kernels and $L$ is the kernel size divided by $2$. The resulting value of intensity is normalized or thresholded for eliminating low confident edges. The local edge orientation is estimated with the orientation of the kernel that yields the maximum response, as in \cite{gonzalez1991, szeliski2010}.

\begin{align}\label{formula:magnitude_directional}
G[f(x,y)] = \max_{z=k}\sum\limits_{i=-L}^L\sum\limits_{j=-L}^L g_{ij}^{(z)} * f(x+i, y+j)
\end{align}

The operators are also orthogonal discrete isotropic filters so we are going to represent and use only $G_{x}$ kernel. To obtain the other kernels for this template gradient we need to rotate it by a fraction of ${\pi}$, different from the previous ones. 

For our analysis we found in literature the following  \textbf{Prewitt Compass Operator} \cite{PrewittOriginal1970}, \textbf{Robinson Compass Operator} \cite{RobisonEdge1977, PrewittOriginal1970}, \textbf{Kirsch operator} \cite{Kirsch1971}. All those kernel masks details can be found in Figure \ref{fig:compass_kernel_masks} from Appendix \ref{appendix:A}.

Similar to Orthogonal Gradient Operators the resulting edge-map is not $1$ pixel width or with the same magnitude so will threshold and thin the results before the evaluation, details in Algorithm \ref{algorithm:steps_first_order}. Similar to other algorithms to obtain the best results we will apply a smoothing filter before-hand.

\subsection{Frei-Chen operator}
\label{Sec:preliminari_frei_chen}

The \textbf{Frei-Chen operator} \cite{ChenFast1977, ParkFourier1990} works on a $3x3$ footprint but applies a total of nine convolution masks to the image. Frei-Chen masks are unique masks, which contain all of the basis vectors. 

The Frei-Chen masks one trough nine, defined on a $3x3$ window span the edge, line, and average subspaces. All the kernel masks for the operators are presented in Figure \ref{fig:frei_kernel_masks} from Appendix. Kernels $G_1$ and $G_2$ are isotropic average gradient basis vectors and kernels $G_3$ and $G_4$ are the ripple vectors. These formulas contribute to the edge detection sub space. Kernels $G_5$ and $G_6$ are the line basis vectors, respective kernels $G_7$ and $G_8$ are the discrete Laplacian vectors and are used for the line subspace detection. Kernel $G_9$ is the average mask \cite{ChenFast1977}. To use the Frei-Chen operator for edge detection we apply the Equation \ref{formula:frei_edge} for the first $4$ masks. To use the Frei-Chen operator for line detection the Equation \ref{formula:frei_line} for the first $4$ masks should be used.

    \begin{align}\label{formula:frei_edge}
        G[f(x,y)]_{edge} = \sqrt{\frac{\sum\limits_{k=1}^4 (G_k)^2}{\sum\limits_{k=1}^9 (G_k)^2}}
    \end{align}

    \begin{align}\label{formula:frei_line}
        G[f(x,y)]_{line} = \sqrt{\frac{\sum\limits_{k=4}^8 (G_k)^2}{\sum\limits_{k=1}^9 (G_k)^2}}
    \end{align}

In some way the Frei-Chen operator is a First Order Derivative Compass Gradient Operator so we will threat it as such when evaluation, see Algorithm \ref{algorithm:steps_first_order}. Even if line detection is not in our analysis scope we will consider the line output this algorithm as is closely coupled with the edge result.

\subsection{Laplacian edge operator}
\label{Sec:preliminari_laplacian}

The Laplacian is a 2-D isotropic measure of the second spatial derivative of an image. The Laplacian of an image highlights regions of rapid intensity change and is therefore often used for edge detection. Another difference between Laplacian and other operators is that Laplacian does not take out edges in any particular direction. The formula for the Laplacian \cite{jainMachine1995} is the second-order derivative by each component of a 2D function as is presented in the Equation \ref{formula:second_order}. 

\begin{align}\label{formula:second_order}
\nabla^2 f(x,y) = \frac{\partial^2{f(x,y)}}{\partial{x^2}} + \frac{\partial^2{f(x,y)}}{\partial{y^2}}
\end{align}

From literature \cite{haralick1992computer, szeliski2010, jainMachine1995, davies1992skimming} we can find different estimation of isotropic kernels for Laplacian operator that we preset in Figure \ref{fig:laplace_kernel_masks} in Appendix \ref{appendix:A}. For our work we will consider all of them so we can highlight changes that appear upon edge map when choosing different approximation of the Laplace function.

The steps we need to take to obtain the desired edge-map format are presented in Algorithm \ref{algorithm:laplace_steps}. We apply the Equation \ref{formula:second_order} to obtain the raw edge map but to able to evaluate we need to transform the edge result from natural range of ($-250$, $+250$) to a ($0$, $250$). Afterwards the scaled edge map is thinned for the final result. We can accept that this isn't the normal usage of the Laplace operator but we wanted to evaluate it separately from Laplace of Gaussian or Marr-Hildreth operators. 

\begin{algorithm}[!h]
    \DontPrintSemicolon
    \KwInput{RGB image}
    \KwOutput{Binary edge map}
    \KwData{Testing set BSDS500\cite{Bsds2011}}
  
    \For{edge operator}{
        Convert image to gray-scale image.
    
        \tcc{For Laplace Operator by applying the edge detection operator using Equation \ref{formula:second_order}}
        \tcc{For LoG by applying the edge detection operator using Equation \ref{formula:log_2_steps}}
        Applying the edge detection operator
    
        \tcc{Use global threshold so each pixel, which has an absolute value of the intensity higher or equal to a \textit{threshold}, will have its value set to a max value(e.g. 255) else to 0.}
        Apply global threshold algorithm
    
        \tcc{Apply the thinning algorithm presented in \cite{GuoHall1989} to remove the excess edge points caused by the convolution and threshold}
        Apply thinning algorithm
    }
\caption{Laplace and LoG Operator steps}
\label{algorithm:laplace_steps}
\end{algorithm}

\subsection{Laplacian of Gaussian - LoG - or Mexican Hat operator}
\label{Sec:preliminari_log}

The Laplacian of Gaussian (LoG) approach defines that the image is convoluted with an Gaussian filter to reduce the noise followed by an Laplacian convolution to expose the edges. The LoG function with the three dimensional plot looks like the Mexican Hat hence the name of the operator \cite{torre1986edge, haralick1992computer}. We can use two mathematical variants for obtaining LoG, (see the Equation \ref{formula:log_2_steps}) : convolve the image with a Gaussian smoothing filter and afterwards compute with the Laplacian Operator, or convolve the image with the linear filter that is the Laplacian of the Gaussian filter (Equation \ref{formula:log}).

\begin{align}\label{formula:log_2_steps}
LoG(x,y) = \nabla^2\left[ (g(x,y) \star f(x,y) ) \right] = \nabla^2\left[ g(x,y) \right] \star f(x,y)
\end{align}

\begin{align}\label{formula:log}
\nabla^2 g(x,y) = -\frac{1}{\pi\sigma^4}\left[1-\frac{x^2 + y^2}{2\sigma^2}\right]e^{-\frac{x^2+y^2}{2\sigma^2}}
\end{align}

Similar to the Laplace operator we need to transform the resulted edge map from range of ($-250$, $+250$) to a ($0$, $250$) so we use the steps from Algorithm \ref{algorithm:laplace_steps}. The determined edge map is scaled followed by thresholding and thinning.

\subsection{Marr–Hildreth algorithm}
\label{Sec:preliminari_marr}

Another method of detecting edges in digital images, is the Marr–Hildreth algorithm, there are used continuous curves where are strong and rapid variations in image brightness. 

The Marr–Hildreth edge detection method is simple and operates by convolving the image with the Laplacian of the Gaussian function, or, as a fast approximation by difference of Gaussian. Zero crossings are detected in the filtered result to obtain the edges. A zero crossing at pixel level implies that the signs of at least two opposite neighboring pixels are different \cite{marr1980theory, HARALICK1987216}.

For our implementation of Zero Crossing algorithm we choose to implement a threshold that will permit us to discriminate better relevant zero crossing according to the difference in intensity \cite{HARALICK1987216, Hildreth1985}. This variant of Zero Crossing will produce better results than the classical version that threshold the results at zero regardless of intensity change.

To obtain the desire edge-map format we have to thin the zero crossing output of the LoG algorithm. All the steps used for simulations are presented in Algorithm \ref{algorithm:marr_steps}.

\begin{algorithm}[!h]
    \DontPrintSemicolon
    \KwInput{RGB image}
    \KwOutput{Binary edge map}
    \KwData{Testing set BSDS500\cite{Bsds2011}}
  
    \For{edge operator}{
        Convert image to gray-scale image.
    
        \tcc{Applying the edge detection operator using Equation \ref{formula:log_2_steps}}
        \tcc{Kernel used is the convolution result of the Laplace kernel and the Gaussian kernel}
        Applying the edge detection operator
    
        \tcc{Zero crossing algorithm to detect where the LoG output changes sign}
        Apply Zero Crossing algorithm
    
        \tcc{Apply the thinning algorithm presented in \cite{GuoHall1989} to remove the excess edge points caused by the convolution and threshold}
        Apply thinning algorithm
    }
\caption{Marr–Hildreth Operator steps}
\label{algorithm:marr_steps}
\end{algorithm}

\subsection{Canny algorithm}
\label{Sec:preliminari_canny}

Canny edge detection algorithm is a classical and robust method for edge detection in gray-scale images. The edge detection algorithm is widely used due to its short operation time, relatively simple calculation process. The two significant features of this method are introduction of Non-Maximum Suppression and double threshold of the gradient image, see \cite{Canny1986}.

The traditional Canny algorithm \cite{Canny1986} has the following steps: 1) smooth the image with a Gaussian function, 2) apply the first order operator, 3) Non-maximum suppression of the magnitude of the gradient, 4) double threshold is used for edge connection and edge connections.

Non-maximum suppression is an important step for Canny algorithm. The purpose of it is to find the local maximum value of the pixels, and set the gray value corresponding to the non-maximum point to zero, so that a large part of non-edge points can be eliminated.

After the non-maximum suppression phase using the two thresholds, we double threshold and edge link by hysteresis the edge points. If an edge pixel’s gradient value is higher than the high threshold it is set as strong edge pixel. Similar if a edge pixel gradient value is smaller than the high threshold value but larger than the low threshold value, it is marked as a weak edge pixel. If an edge pixel's gradient value is smaller than the low threshold value, it will be suppressed. At the end the remaining "weak" and "strong" pixels are connected as long as there is one strong edge pixel that is involved in the blob.

The classic Canny Edge algorithm uses the Sobel Operator during the convolution step in order to compute the gradient. Other operators such as the dilated Sobel can be also used during this step. Thus, the Canny Edge algorithm offers flexibility when choosing the filters for the convolution step.

\begin{algorithm}[!h]
    \DontPrintSemicolon
    \KwInput{RGB image}
    \KwOutput{Binary edge map}
    \KwData{Testing set BSDS500\cite{Bsds2011}}
  
    \For{edge operator}{
        Convert image to gray-scale image.
        
        \tcc{Needed to enhance the results of the following steps}
        Apply Gaussian filter smoothing
        
        \tcc{Applying the filters by convolving the gray-scale image with their kernels on the $x$ and $y$ axes}
        Apply convolution with kernels
        
        \tcc{Applying the filters by convolving the gray-scale image with their kernels on the $x$ and $y$ axes}
        Calculate gradient magnitude
        
        \tcc{Non-maximum suppression, for edge thinning of the obtained results}
        Apply non-maximum suppression
        
        \tcc{Find max value of grey image for double threshold, }
        Find max pixel value of grey image
        
        \tcc{Edge tracking by hysteresis using double threshold}
        \tcc{Double threshold using Equation \ref{ratio_threshold}}
        Apply edge tracking
    }
\caption{Canny Operator steps}
\label{algorithm:canny_steps}
\end{algorithm}

\subsection{Shen-Castan algorithm}
\label{Sec:preliminari_shen_castan}

The unique feature of the Shen-Castan Edge Detection is that it can detect the edge of an image with noise. By using an Infinite Symmetric Exponential Filter, the noise will be eliminated.

Infinite Symmetric Exponential Filter (ISEF) \cite{shen1990}, described as a real continous function by the Equation \ref{formula:isef_continous} and the recursive function by Equation \ref{formula:isef_recursive}, Where b is the Thinning Factor and its values lies in between 0 and 1.

\begin{align}\label{formula:isef_continous}
f(x) = \frac{p}{2}e^{-p|x|}
\end{align}

\begin{align}\label{formula:isef_recursive}
f(x, y) = \frac{(1-b)b^{(|X| + |y|)}}{1+b}
\end{align}

Shen-Castan edge detector has a unique step that overcomes noise in the input image by using the Infinite Symmetric Exponential Filter, see \cite{shen1992optimal}. The algorithm has the following steps: 1) smooth the image with the ISEF filter, 2) convolve the smoothed image with a Laplace binary operator, 3) threshold the edge strong edge pixels using Zero Crossing, 4)  link the edge points by hysteresis and 5) thin the results at the end .

The approximation of Laplacian is computed by subtracting the original image from the smoothed one. The result is a band-limited Laplacian image. Next, a binary Laplacian image is generated by setting all the positive valued pixels to 1 and all others to 0, as in \cite{shen1992optimal}.

For our analysis we will not use the Laplace binary operator, which is considered to be optimal as run time for this algorithm but use the versions of Laplace isotropic kernels presented in subsection \ref{Sec:preliminari_laplacian}. Details of the steps we will use for obtaining the edge map is presented in Algorithm \ref{algorithm:shen-castan_steps}.

\begin{algorithm}[!h]
    \DontPrintSemicolon
    \KwInput{RGB image}
    \KwOutput{Binary edge map}
    \KwData{Testing set BSDS500\cite{Bsds2011}}
  
    \For{edge operator}{
        Convert image to gray-scale image.
        
        \tcc{Recursive ISEF filter on X direction and then Y direction}
        Apply ISEF filter smoothing
        
        \tcc{Original is Laplace binary but we can use any Laplace Operator}
        Apply Laplace Operator
        
        \tcc{replacing the pixel values by 1 for positive values and 0 for negative values. Zero crossing pixels, we need to simply look at the boundaries of the non-zero regions in this binary image}
        Apply Zero Crossing 
        
        \tcc{Adaptive gradient method with fixed width W is centred at candidate edge pixels found after zero crossing}
        Apply Non-Max Suppression
        
        \tcc{Edge tracking by hysteresis using double threshold}
        Apply Hysteresis Threshold
        
        Apply Thinning 
    }
\caption{Shen-Castan Operator steps}
\label{algorithm:shen-castan_steps}
\end{algorithm}

\subsection{Edge Drawing algorithm}
\label{Sec:preliminari_ed}

Edge Drawing (ED) is an edge detection algorithm that works by computing a set of anchor points, which are most likely to be edge elements, and linking them with a predefined set of rules which we call smart routing. \cite{EdgeDrawing2012}

ED algorithm \cite{EdgeDrawing2012}, described in Algorithm \ref{algorithm:ed}, can be summarized in the following steps: suppress the image with a Gaussian filter \cite{GaussianFilter}, calculate the gradient magnitude and orientation using Sobel filter \cite{SobelOriginal1973}, extract the anchor points, connect the anchor points using the smart routing concept. The steps of ED algorithm is presented in Algorithm \ref{algorithm:ed}.

\begin{algorithm}[!h]
    \DontPrintSemicolon
    \KwInput{RGB image}
    \KwOutput{Binary edge map}
    \KwData{Testing set BSDS500\cite{Bsds2011}}
    \For{edge operator}{
        Convert image to gray-scale image.
    
        \tcc{Suppression of noise by Gaussian filtering with $sigma$ parameter}
        Apply Gaussian filter smoothing
        
        \tcc{Using Sobel kernels using $\sqrt{G_x^2 + G_y^2} \approx|G_x| + |G_y|$}
        Calculate gradient map
        
        \tcc{if $|G_x| \geq |G_y|$ vertical edge otherwise horizontal edge}
        Calculate direction map
        
        \tcc{Apply a global threshold scheme by $grad\_thr$ value }
        Threshold gradient map
        
        \tcc{Anchor must be a local peak of the gradient map, using $anchor\_thr$ and $scan\_interval$}
        Extract anchors
    
        \tcc{Three immediate neighbors are considered and the maximum gradient value is picked}
        Smart routing
    }

\caption{Edge Drawing Operator}
\label{algorithm:ed}
\end{algorithm}

The mechanism of connecting anchors is considered the most crucial step of ED. Connecting consecutive anchors is done by passing from one anchor to the next following the cordillera peak of the gradient map mountain. This process, as in \cite{EdgeDrawing2012}, is guided by the gradient magnitude and edge direction maps computed. If a horizontal edge passes through the anchor, we start the connecting process by proceeding to the left and to the right. If a vertical edge passes through the anchor, we start the connection process by proceeding up and down. The process stops if we move out of the edge area or we encounter a previously detected edge.

\subsection*{Benchmarking the edge operators}  

BSDS500 \cite{Bsds2011} is a widely used dataset in the field of computer vision for benchmarking edge detection algorithms. It contains natural images that have been manually segmented and is considered the ground truth in many boundary detection comparisons. The benchmark is used to evaluate the result images generated by a specific algorithm with the segmented images in the dataset. For edge detection evaluation we used the Corresponding Pixel Metric (CPM) algorithm, defined in \cite{CPM2003}. This measure is used for correlating similarities with a small localization error in the detected edges. CPM first finds an optimal matching of the pixels between the edge images and then estimate the error produced by this matching.  

The BSDS500 benchmark offers $500$ images for testing, presented in few different sets. The images are natural images marked for the boundaries and edges of the objects and structures that they represent or contain.

For each benchmark image, three different probability measures are computed: precision ($P$), recall ($R$) and F-measure ($F1$) defined in \cite{F12007}. Precision (Equation \ref{precision}) is the probability that a resulting edge/boundary pixel was labeled as a true edge/boundary pixel. Recall (Equation \ref{recall}) is the probability that a true edge/boundary pixel was detected. F-measure (Equation \ref{f-measure}) is the accuracy measure computed as an average between precision and recall. Also, we need to specify that $TP$ (True Positive) represents the number of matched edge pixels, $FP$ (False Positive) the number of edge pixels which are incorrectly highlighted as edge pixel and $FN$ (False Negative) the number of pixel that have not been detected as edge pixel but in dataset has been labeled as edge pixel.

% \begin{minipage}{.48\linewidth}

\begin{align}\label{precision}
    P = \frac{TP}{TP+FP}.
\end{align}
% \end{minipage}%
% \begin{minipage}{.48\linewidth}

\begin{align}\label{recall}
    R = \frac{TP}{TP+FN}.
\end{align}  
% \end{minipage}

\begin{align}\label{f-measure}
F-measure = 2*\frac{P*R}{P+R}=\frac{2*TP}{2*TP+FP+FN} 
\end{align}
\vspace{\baselineskip}

\section{Experimental results}
\label{Sec:simulation_results}

In this section we will present the results of our analysis for each edge operator presented in Section \ref{Sec:preliminaries}. In our presentation will consist of visual results, statistical results and remarks for each subsection.

For our simulation to be reproducible and easy to use we have used End-to-End Computer Vision Framework\footnote{https://github.com/CipiOrhei/eecvf} -EECVF-\cite{eecvf2020}. EECVF is an adaptable and dynamic framework designed for researching and testing CV concepts which does not require the user to handle the interconnections throughout the system. All the edge operators and algorithms are present in the framework and can be reproduce by running the $main\_dilated\_filters\_for\_edge\_detection\_algorithms$ module.

\subsection{First Order Derivative Orthogonal Gradient operators}

The first category we analyze is the First Order Derivative Orthogonal Gradient Operators that are described in Section \ref{Sec:preliminari_first_ortho}. Using the steps described in Algorithm \ref{algorithm:steps_first_order} we will compare the results we obtain when using the standard kernels, extended kernels and dilated ones.

To obtain the best results from the steps presented in Algorithm \ref{algorithm:steps_first_order} we need to choose a good \textit{threshold} value for Step 3 and of course choose a \textit{sigma} value for Step 2, that will benefit all images. We know from experience and literature that this two aspects can completely change the results of our simulations. For this parameter tuning we will choose the \cite{SobelOriginal1973} operator, being one of the most popular in this category. 

\begin{figure}[H]
    \centering
    \begin{minipage}{0.45\textwidth}
        \centering
        \includegraphics[scale=0.2]{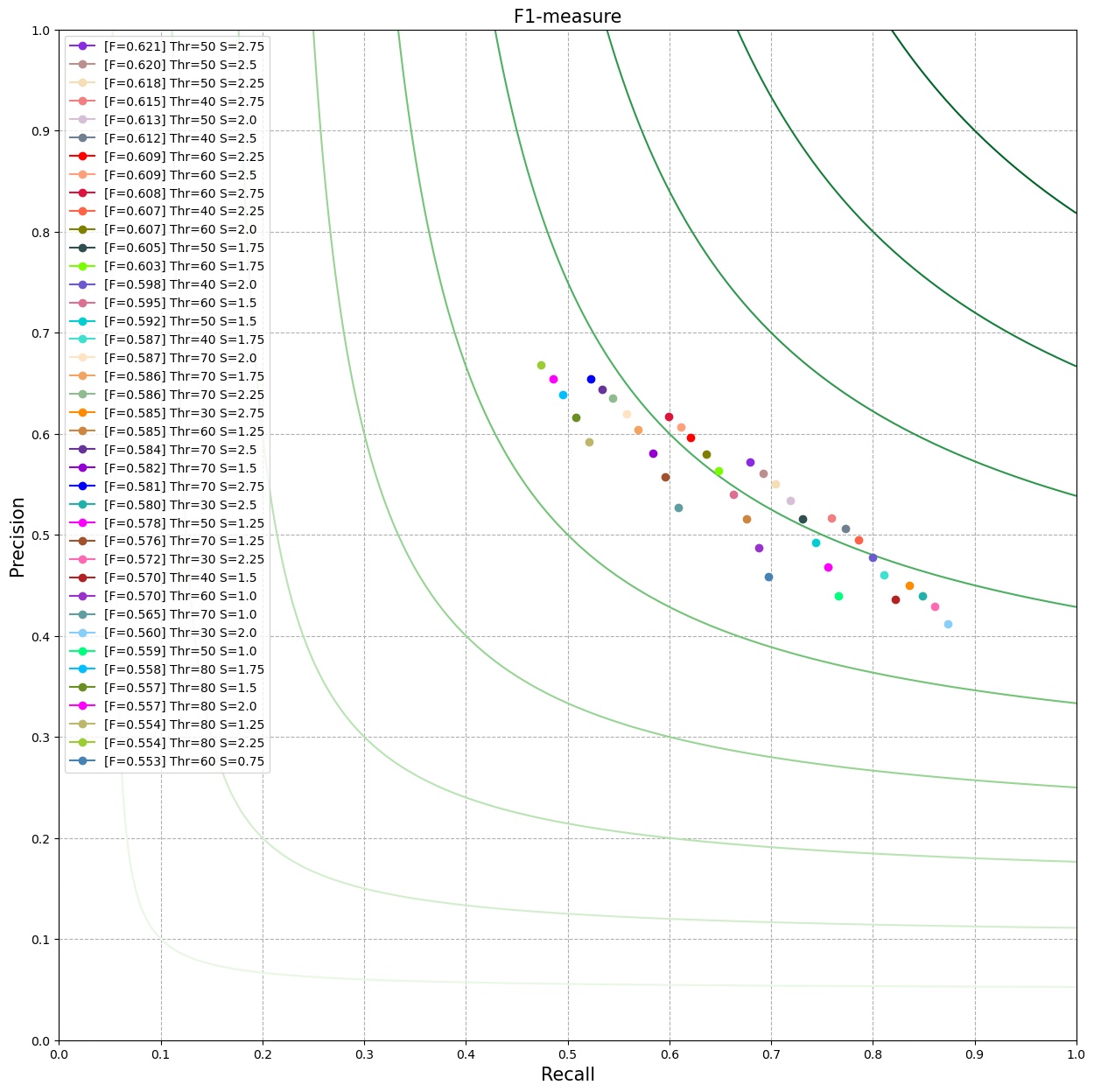} 
        \caption{Parameter tuning for First Order Derivative Orthogonal Gradient Operators using a Sobel 3x3}
        \label{fig:thr_first_thr_sigma_order_tunning}
    \end{minipage}\hfill
    \begin{minipage}{0.45\textwidth}
        \centering
        \includegraphics[scale=0.2]{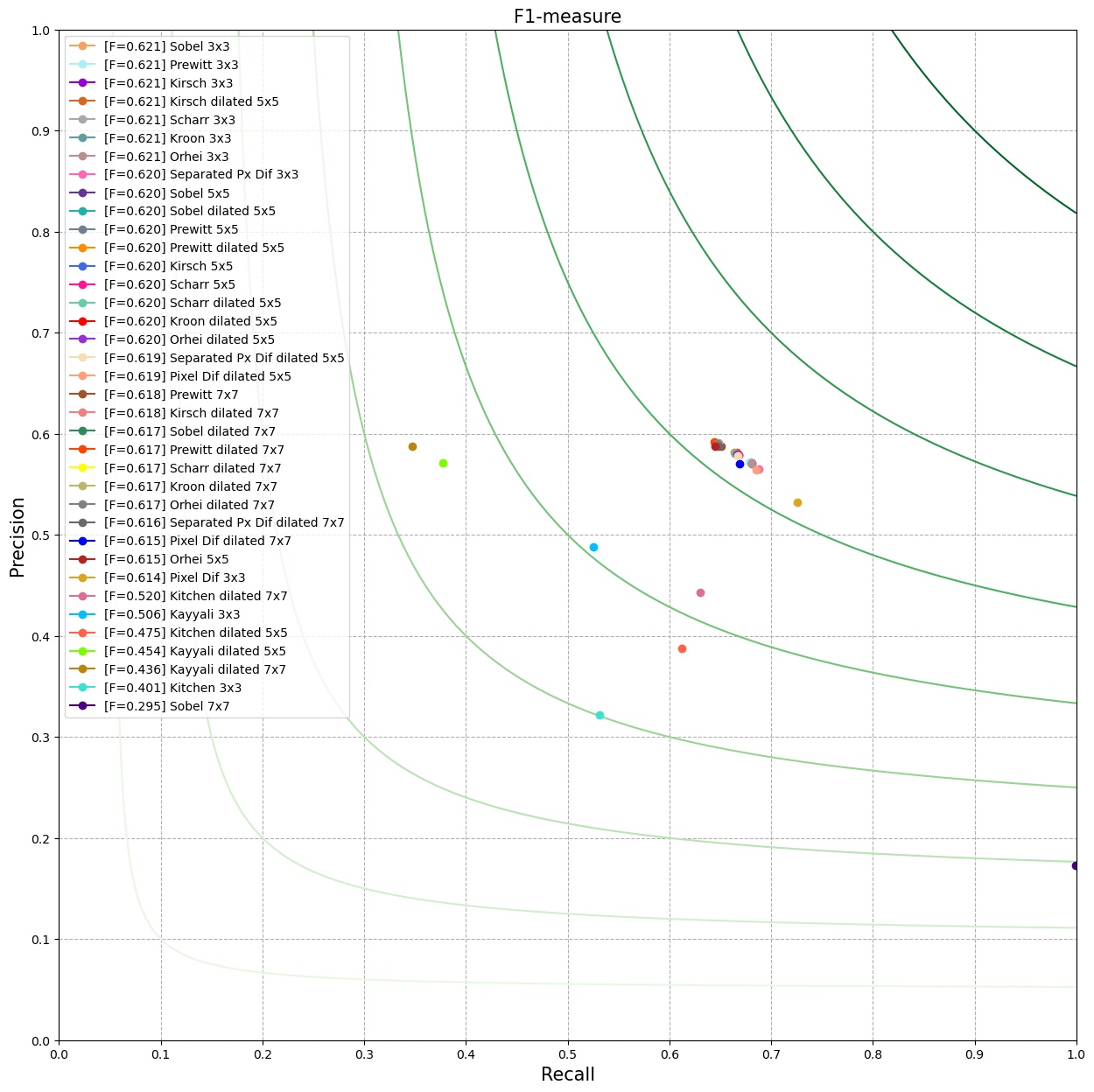} 
        \caption{Results for First Order Derivative Orthogonal Gradient Operators}
        \label{fig:first_order_results}
    \end{minipage}\hfill
\end{figure}

In Figure \ref{fig:thr_first_thr_sigma_order_tunning} we can observe the $F1$ results for a range of the threshold between $30$ and $160$ with a step of $10$ and sigma value between $0.25$ and $3.0$ with a step of $0.25$.

\begin{figure}[H]
% \begin{minipage}{.4\textwidth}
\centering
\setlength{\tabcolsep}{0.5pt}
\begin{tabular}{cccc} 
	    \includegraphics[width=0.17\textwidth]{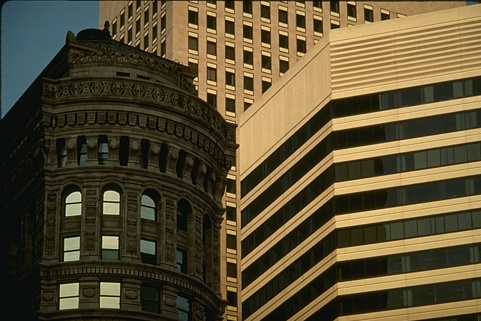} 
	&
        \includegraphics[width=0.17\textwidth]{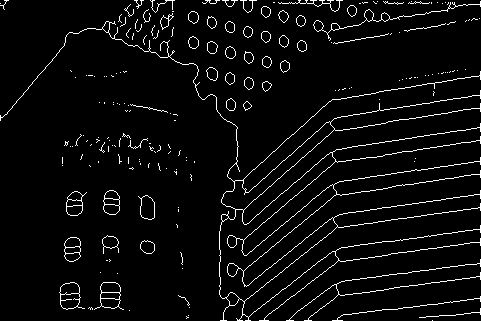} 
    % & 
    %     ~
    % &
    %     ~
    & 
        \includegraphics[width=0.17\textwidth]{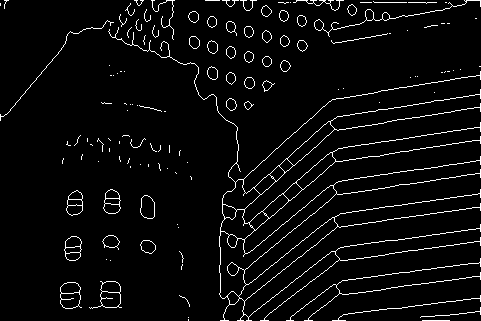}
    & 
        \includegraphics[width=0.17\textwidth]{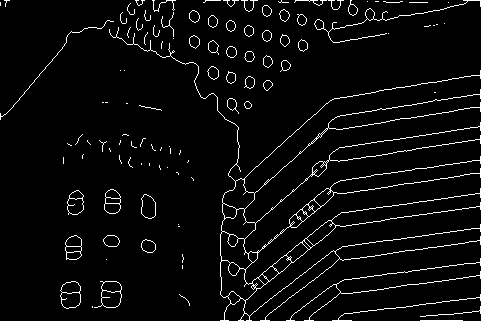}
\\
	    \includegraphics[width=0.17\textwidth]{result_images/first_order_ortho/48017.jpg} 
	&
        \includegraphics[width=0.17\textwidth]{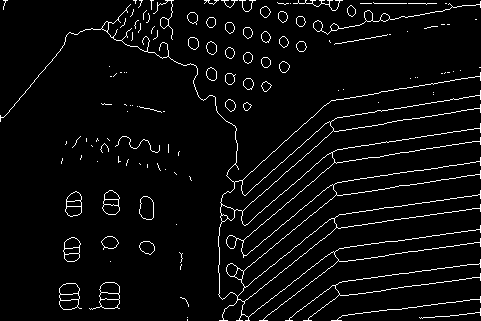} 
    % &
    %     ~
    % & 
    %     ~
    & 
        \includegraphics[width=0.17\textwidth]{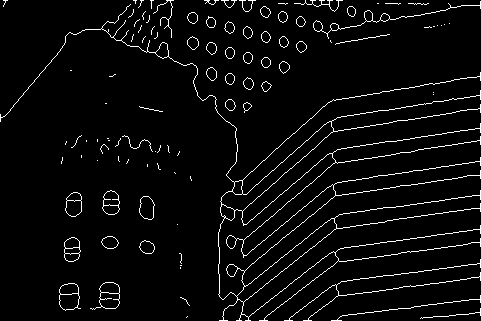}
    & 
        \includegraphics[width=0.17\textwidth]{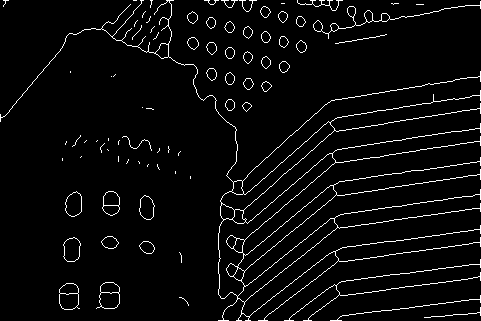}
\\
	    \includegraphics[width=0.17\textwidth]{result_images/first_order_ortho/48017.jpg} 
	&
        \includegraphics[width=0.17\textwidth]{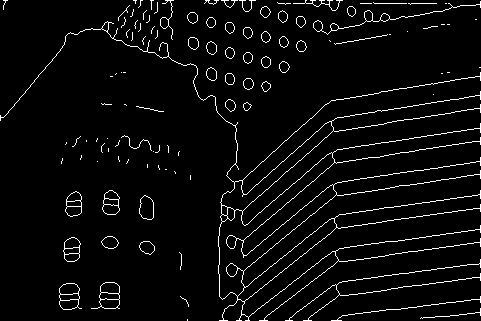} 
    % &
    %     \includegraphics[width=0.17\textwidth]{result_images/first_order_ortho/FINAL_SOBEL_5x5_BLURED_L0_48017.png}
    % & 
    %     \includegraphics[width=0.17\textwidth]{result_images/first_order_ortho/FINAL_SOBEL_7x7_BLURED_L0_48017.png} 
    & 
        \includegraphics[width=0.17\textwidth]{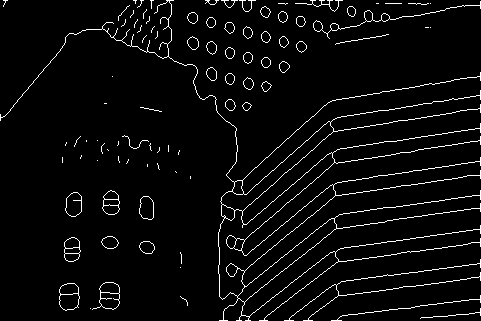}
    & 
        \includegraphics[width=0.17\textwidth]{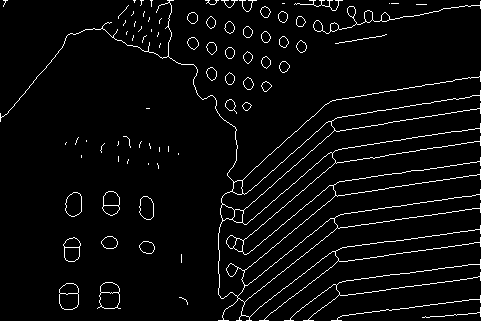}
\\
	    \includegraphics[width=0.17\textwidth]{result_images/first_order_ortho/48017.jpg} 
	&
      \includegraphics[width=0.17\textwidth]{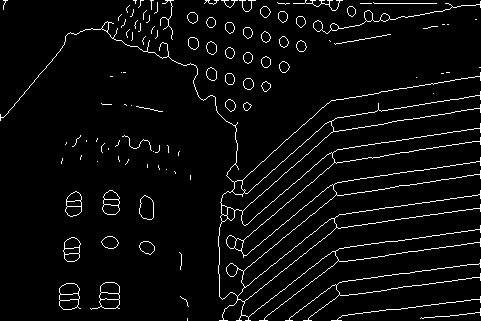} 
    % & 
    %     \includegraphics[width=0.17\textwidth]{result_images/first_order_ortho/FINAL_PREWITT_5x5_BLURED_L0_48017.png} 
    % & 
    %     \includegraphics[width=0.17\textwidth]{result_images/first_order_ortho/FINAL_PREWITT_7x7_BLURED_L0_48017.png} 
    & 
        \includegraphics[width=0.17\textwidth]{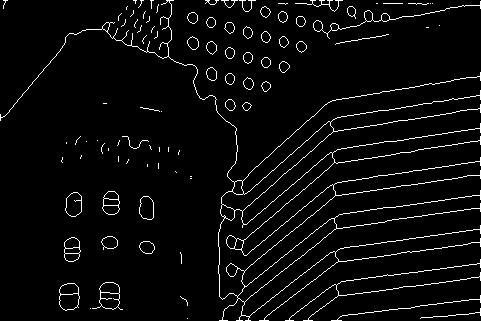}
    & 
        \includegraphics[width=0.17\textwidth]{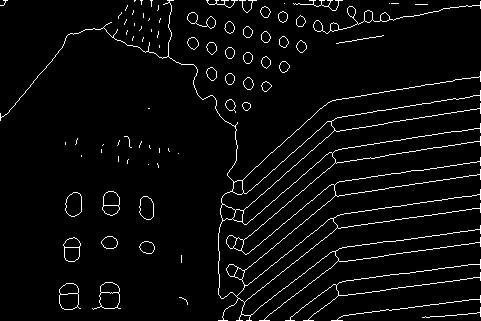}
\\
	    \includegraphics[width=0.17\textwidth]{result_images/first_order_ortho/48017.jpg} 
	&
        \includegraphics[width=0.17\textwidth]{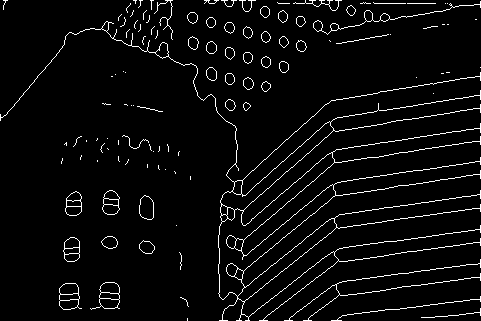} 
    % & 
    %     \includegraphics[width=0.17\textwidth]{result_images/first_order_ortho/FINAL_KIRSCH_5x5_BLURED_L0_48017.png}
    % & 
    %     ~
    & 
        \includegraphics[width=0.17\textwidth]{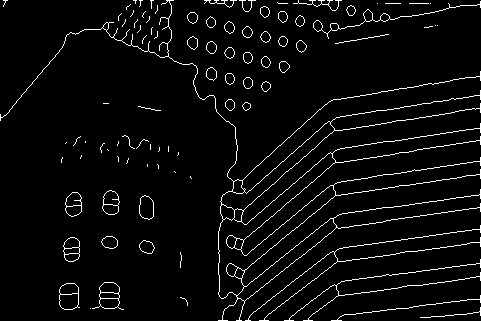}
    & 
        \includegraphics[width=0.17\textwidth]{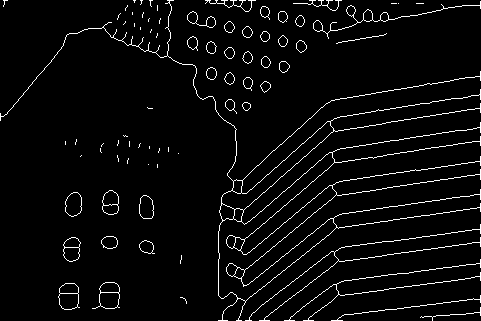}
\\
	    \includegraphics[width=0.17\textwidth]{result_images/first_order_ortho/48017.jpg} 
	&
        \includegraphics[width=0.17\textwidth]{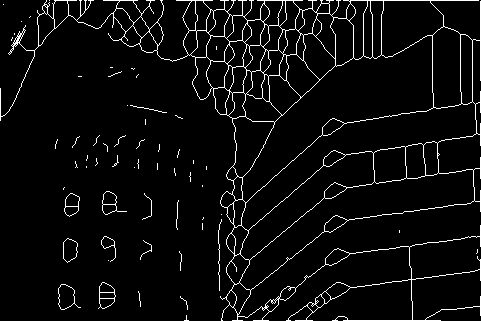} 
    % &
    %     ~
    % &
    %     ~
    & 
        \includegraphics[width=0.17\textwidth]{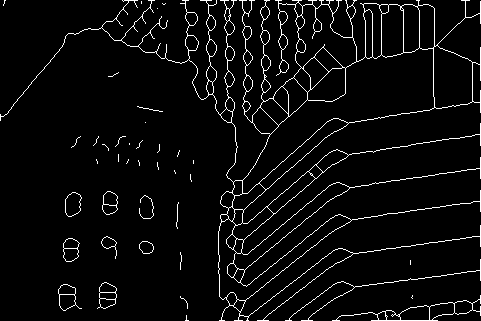}
    & 
        \includegraphics[width=0.17\textwidth]{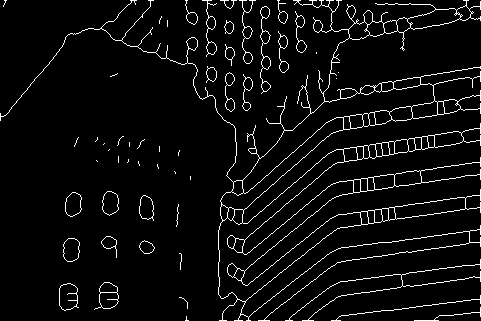}
\\
	    \includegraphics[width=0.17\textwidth]{result_images/first_order_ortho/48017.jpg} 
	&
        \includegraphics[width=0.17\textwidth]{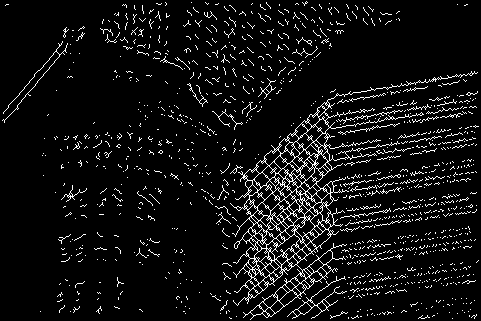} 
    % &
    %     ~
    % &
    %     ~
    & 
        \includegraphics[width=0.17\textwidth]{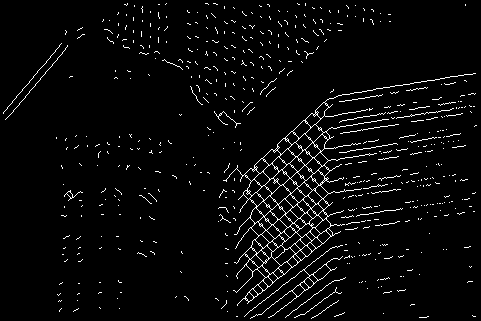}
    & 
        \includegraphics[width=0.17\textwidth]{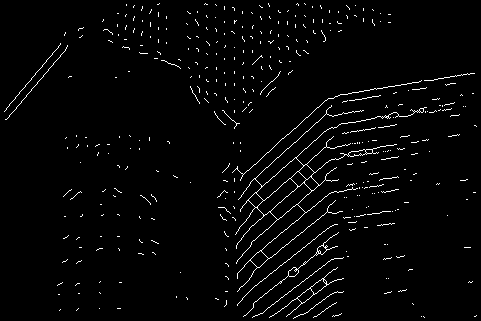}
\\
	    \includegraphics[width=0.17\textwidth]{result_images/first_order_ortho/48017.jpg} 
	&
        \includegraphics[width=0.17\textwidth]{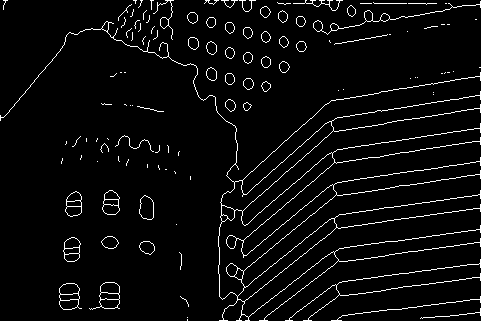} 
    % & 
    %     \includegraphics[width=0.17\textwidth]{result_images/first_order_ortho/FINAL_SCHARR_5x5_BLURED_L0_48017.png}
    % &
    %     ~
    & 
        \includegraphics[width=0.17\textwidth]{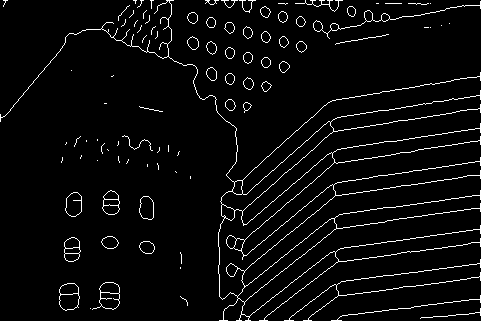}
    & 
        \includegraphics[width=0.17\textwidth]{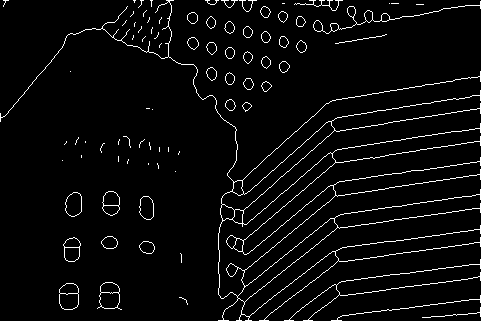}
\\
	    \includegraphics[width=0.17\textwidth]{result_images/first_order_ortho/48017.jpg} 
	&
        \includegraphics[width=0.17\textwidth]{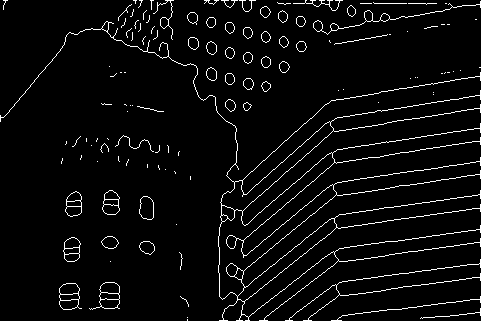} 
    % & 
    %     ~
    % &
    %     ~
    & 
        \includegraphics[width=0.17\textwidth]{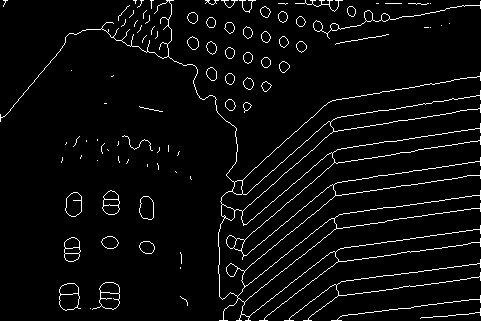}
    & 
        \includegraphics[width=0.17\textwidth]{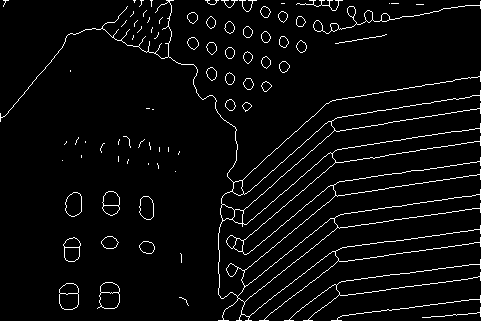}
\\
	    \includegraphics[width=0.17\textwidth]{result_images/first_order_ortho/48017.jpg} 
	&
        \includegraphics[width=0.17\textwidth]{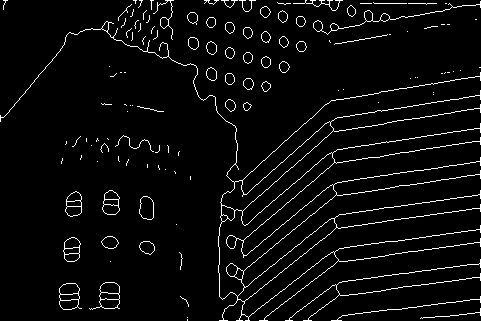} 
    % & 
    %     \includegraphics[width=0.17\textwidth]{result_images/first_order_ortho/FINAL_ORHEI_V1_5x5_BLURED_L0_48017.png} 
    % &
    %     ~
    & 
        \includegraphics[width=0.17\textwidth]{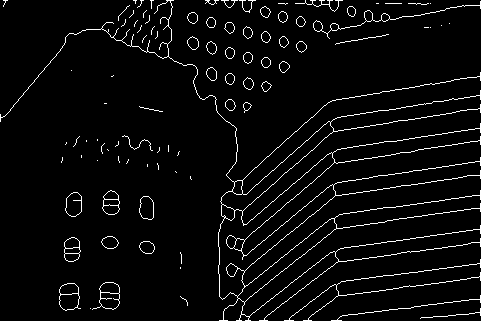}
    & 
        \includegraphics[width=0.17\textwidth]{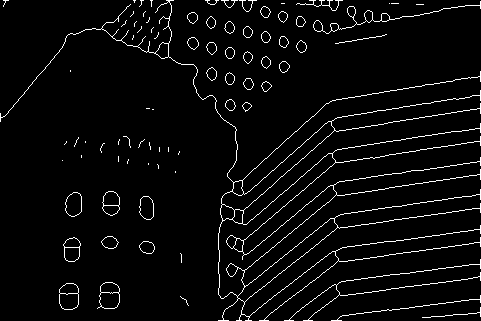}
\end{tabular}
% \caption{First order derivative Orthogonal Gradient results;  Columns: Original, 3x3, 5x5, 7x7, Dilated 5x5, Dilated 7x7; Rows: Pixel Differences, Separated Pixel Difference, Sobel, Prewitt, Kirsch, Kitchen, Kayyali, Scharr, Kroon, Orhei}
\caption{First Order Derivative Orthogonal Gradient Operators results;  Columns: Original, 3x3, Dilated 5x5, Dilated 7x7; Rows: Pixel Differences, Separated Pixel Difference, Sobel, Prewitt, Kirsch, Kitchen, Kayyali, Scharr, Kroon, Orhei}
\label{fig:first_order_img_results}
% \end{minipage}
\end{figure}%

As expected a higher threshold will produce less edge points but with a fair confidence that resulted in a high \textit{R} because of the lack of points. If we look from another perspective, if we choose a small threshold value we see a higher \textit{P} but a very weak \textit{R} caused by the excess edge point generated. Looking over the results obtain in Figure \ref{fig:thr_first_thr_sigma_order_tunning} we ca observe that the best results we obtain when using sigma value of $2.75$ and a threshold value of $50$.

Next step was to change the kernel we use from Sobel to other operators kernels. We keep the parameters we found in the tuning phase so we will have a objective comparison of the results. The annotation we use to represent the results are: $Thr$ for gradient threshold and $S$ for Gaussian Sigma value.

Visual comparison results of the operators are presented in Figures \ref{fig:first_order_img_results}. We can observe (by comparing the Figures \ref{fig:first_order_img_results} columns) from our simulation that the dilation of the kernels doesn't produce a degradation of the edge map. An increase in "noise" edge pixels is a foreseen effect of the extensions of the convoluted area. By dilation we keep the benefits of the increase region of interest but lose the excess "noise" pixels we would obtain by extending.

From Figure \ref{fig:first_order_img_results} and Table \ref{table:first_order_filter_results} we conclude that dilated filters have similar or better results than the classical ones. Even if the $F1$ score is not always better we can clearly see an improvement of $P$ in most cases. We can observe (looking through $F1$ score) that dilation can bring a small degradation of the edge map as for Kayyali \cite{Kayyali2000} but it can bring improvements like for Kitchen operator \cite{Kitchen1989} or Pixel Difference \cite{Mlsna2009gradient}.

\begin{table}[h!]
\begin{minipage}{.99\textwidth}
   \centering
        \setlength{\tabcolsep}{2pt}
        \scalebox{0.65}
        {
            \begin{tabular}{|l|c|ccccc|ccccc|}
            \hline
            \multicolumn{2}{|c|}{\bfseries Operator} &\multicolumn{5}{c|}{\bfseries Magnitude Gradient}&\multicolumn{5}{c|}{\bfseries Canny}\\
            \hline
            						&	&3x3	&5x5	&Dilated 5x5	&7x7	&Dilated 7x7	&3x3	&5x5	&Dilated 5x5	&7x7	&Dilated 7x7\\
            \hline
            	                    &R	&0.726	&-	&0.685	&-	&0.669	                    &0.000	&-	&0.018	&-	&0.124\\
            Pixel Diff	            &P	&0.532	&-	&0.564	&-	&0.570	                    &0.000	&-	&0.861	&-	&0.750\\
            	                    &F1	&0.614	&-	&\textbf{0.619}	&-	&0.615	            &0.000	&-	&0.036	&-	&\textbf{0.212}\\
            \hline
            	                    &R	&0.688	&-	&0.667	&-	&0.648	                    &0.021	&-	&0.236	&-	&0.341\\
            Separated Pixel Diff	&P	&0.565	&-	&0.578	&-	&0.588	                    &0.869	&-	&0.713	&-	&0.691\\
            	                    &F1	&\textbf{0.62}	&-	&0.619	&-	&0.616	            &0.041	&-	&0.355	&-	&\textbf{0.457}\\
            \hline
            	                    &R	&0.679	&0.668	&0.665	&0.999	&0.646	            &0.701	&0.997	&0.857	&0.999	&0.876\\
            Sobel	                &P	&0.572	&0.579	&0.581	&0.173	&0.591	            &0.504	&0.198	&0.425	&0.146	&0.432\\
            	                    &F1	&\textbf{0.621}	&0.620	&0.620	&0.295	&0.617	    &\textbf{0.587}	&0.330	&0.569	&0.255	&0.578\\
            \hline
            	                    &R	&0.678	&0.665	&0.664	&0.651	&0.644	            &0.551	&0.982	&0.758	&0.990	&0.797\\
            Prewitt	                &P	&0.572	&0.581	&0.582	&0.588	&0.592	            &0.582	&0.268	&0.501	&0.249	&0.505\\
            	                    &F1	&\textbf{0.621}	&0.62	&0.62	&0.618	&0.617	    &0.566	&0.421	&0.603	&0.398	&\textbf{0.615}\\
            \hline
            	                    &R	&0.681	&0.665	&0.666	&-	&0.648	                &0.862	&0.996	&0.888	&-	&0.848\\
            Kirsch	                &P	&0.571	&0.581	&0.582	&-	&\textbf{0.591}	        &0.349	&0.212	&0.330	&-	&0.335\\
            	                    &F1	&\textbf{0.621}	&0.62	&\textbf{0.621}	&-	&0.618	&\textbf{0.497}	&0.350	&0.481	&-	&0.481\\
            \hline
            	                    &R	&0.531	&-	&0.612	&-	&0.630	                    &0.959	&-	&0.975	&-	&0.966\\
            Kitchen	                &P	&0.322	&-	&0.388	&-	&\textbf{0.443}	            &0.248	&-	&0.250	&-	&0.268\\
            	                    &F1	&0.401	&-	&0.475	&-	&\textbf{0.520}	            &0.394	&-	&0.398	&-	&\textbf{0.420}\\
            \hline
            	                    &R	&0.525	&-	&0.377	&-	&0.347	                    &0.478	&-	&0.843	&-	&0.894\\
            Kayyali	                &P	&0.488	&-	&0.571	&-	&\textbf{0.588}	            &0.466	&-	&0.338	&-	&0.335\\
            	                    &F1	&\textbf{0.506}	&-	&0.454	&-	&0.436	            &0.472	&-	&0.483	&-	&\textbf{0.487}\\
            \hline
            	                    &R	&0.681	&0.668	&0.665	&-	&0.647	                &0.978	&0.989	&0.988	&-	&0.984\\
            Scharr	                &P	&0.571	&0.579	&0.581	&-	&\textbf{0.590}	        &0.259	&0.244	&0.253	&-	&0.275\\
            	                    &F1	&\textbf{0.621}	&0.62	&0.62	&-	&0.617	        &0.410	&0.392	&0.403	&-	&\textbf{0.430}\\
            \hline
            	                    &R	&0.681	&-	&0.666	&-	&0.647	                    &0.999	&-	&0.997	&-	&0.992\\
            Kroon	                &P	&0.570	&-	&0.580	&-	&\textbf{0.590}	            &0.175	&-	&0.196	&-	&0.218\\
            	                    &F1	&\textbf{0.621}	&-	&0.620	&-	&0.617	            &0.298	&-	&0.328	&-	&\textbf{0.357}\\
            \hline
            	                    &R	&0.681	&0.645	&0.666	&-	&0.647	                &0.850	&0.997	&0.936	&-	&0.942\\
            Orhei	                &P	&0.570	&0.588	&0.580	&-	&\textbf{0.590}	        &0.404	&0.195	&0.342	&-	&0.351\\
            	                    &F1	&\textbf{0.621}	&0.615	&0.620	&-	&0.617	        &\textbf{0.547}	&0.326	&0.500	&-	&0.512\\
            \hline

            \end{tabular}}
        \vspace{1.5pt}
    \caption{F1-measure results for First Order Derivative Orthogonal Gradient Operators and Canny Operator}
    \label{table:first_order_filter_results}
    \label{table:canny_filter_results}
\end{minipage}
\end{table}

\subsection{First Order Derivative Compass Gradient operators}

The following section consists of the analysis of the results of the First Order Derivative Compass or Directional Gradient. Similar to Magnitude Gradient operators we will conduct our analysis using standard and dilated filters presented in Section \ref{Sec:preliminari_first_compass}. 

\begin{figure}[H]
    \centering
    \begin{minipage}{0.45\textwidth}
        \centering
        \includegraphics[scale=0.2]{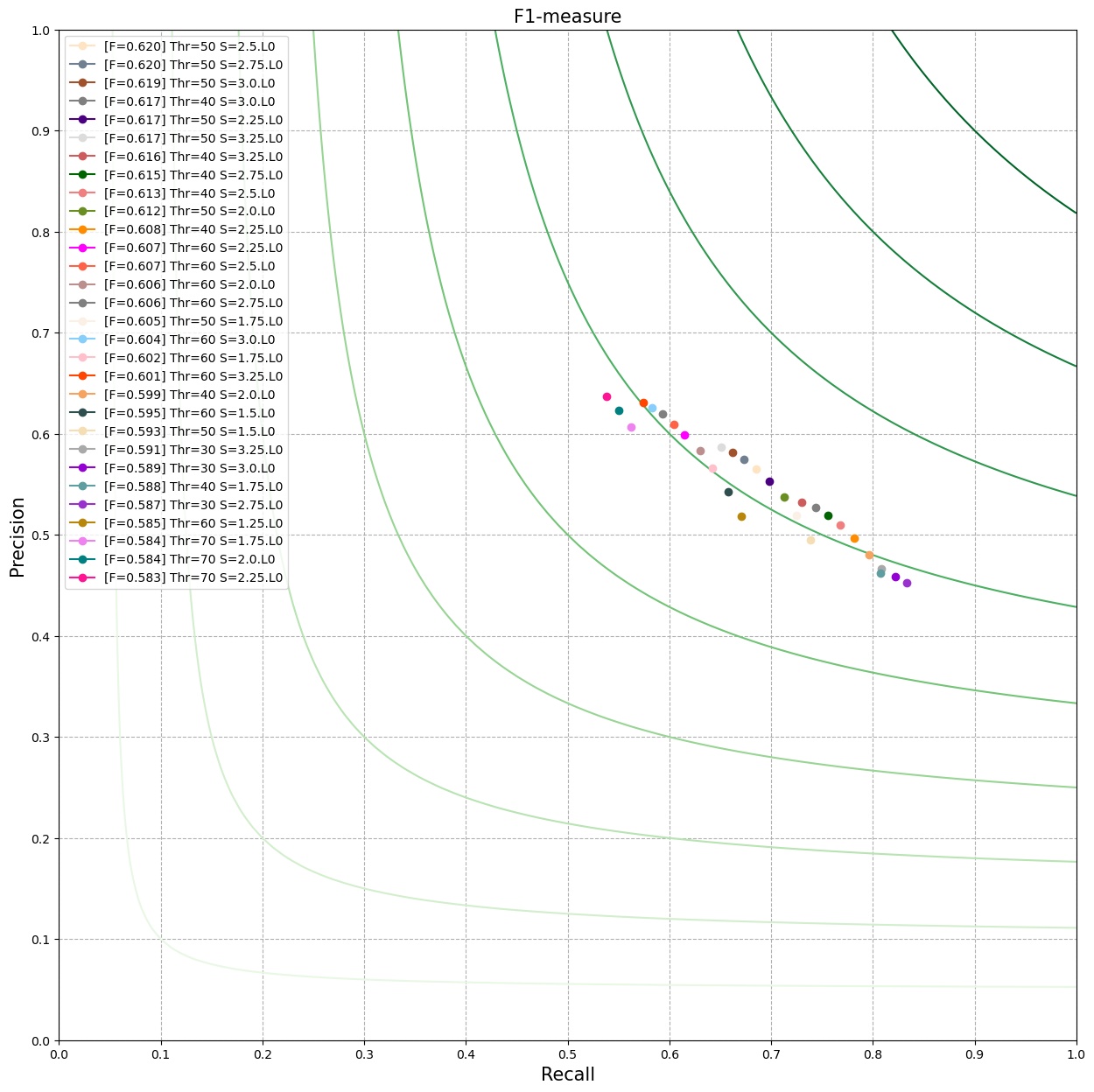} 
        \caption{Parameter tuning for First Order Derivative Compass Gradient Operators using Robinson 3x3}
        \label{fig:compass_thr_sigma_first_order_tunning}
    \end{minipage}\hfill
    \begin{minipage}{0.45\textwidth}
        \centering
        \includegraphics[scale=0.2]{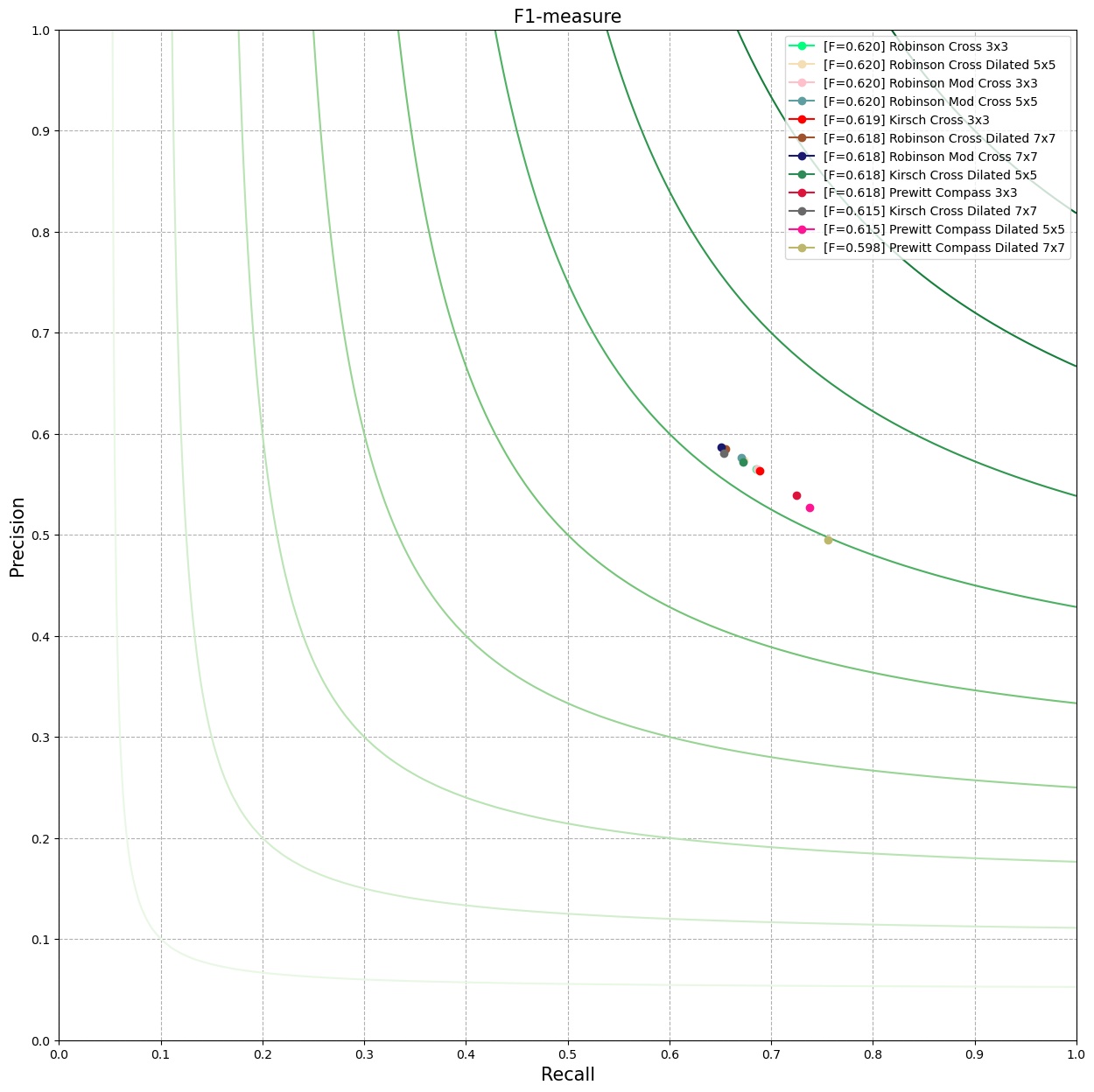} 
        \caption{Results for First Order Derivative Compass Gradient Operators}
        \label{fig:compass_first_order_results}
    \end{minipage}\hfill
\end{figure}

We present first some visual comparison results of the operators, in Figures \ref{fig:compass_thr_sigma_first_order_tunning}. As a first impression we can observe that dilation brings forward a cleaning of artifacts in the overall image. This is an important aspect for this edge operator because it is very affected by "noise" in the input image. 

\begin{figure}[H]
% \begin{minipage}{0.47\textwidth}
\centering
\setlength{\tabcolsep}{0.47pt}
\begin{tabular}{cccccc} 
        \includegraphics[width=0.17\textwidth]{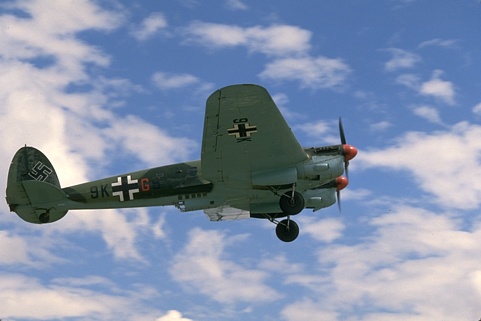}
    &
        \includegraphics[width=0.17\textwidth]{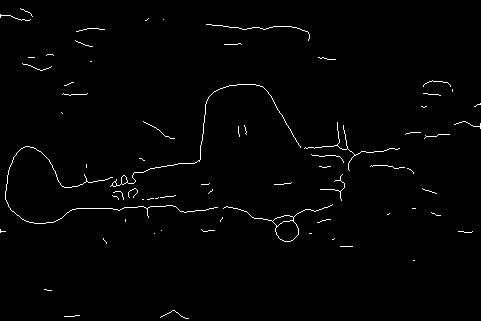} 
    & 
        \includegraphics[width=0.17\textwidth]{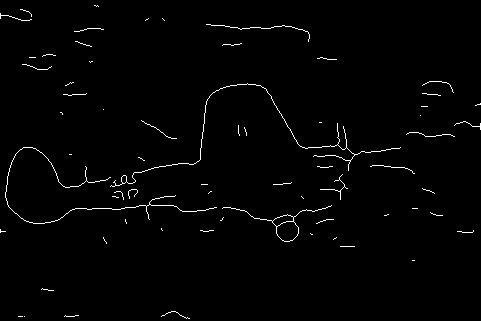} 
    &
        \includegraphics[width=0.17\textwidth]{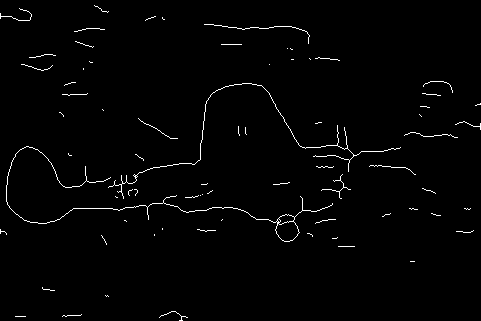} 
\\
        \includegraphics[width=0.17\textwidth]{result_images/first_order_compass/3063.jpg}
    &
        \includegraphics[width=0.17\textwidth]{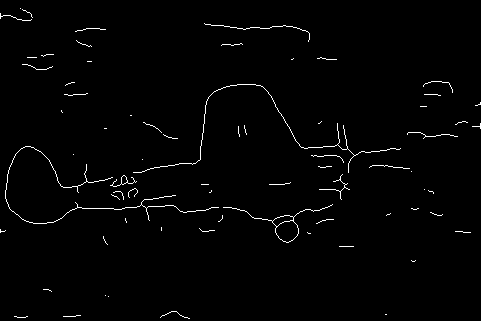} 
    & 
        \includegraphics[width=0.17\textwidth]{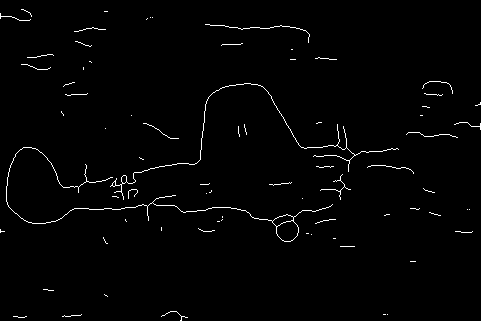} 
    &
        \includegraphics[width=0.17\textwidth]{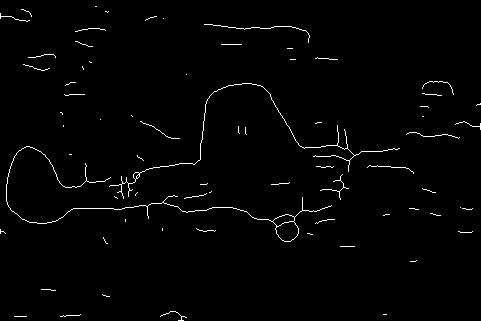} 
\\
        \includegraphics[width=0.17\textwidth]{result_images/first_order_compass/3063.jpg}
    &
        \includegraphics[width=0.17\textwidth]{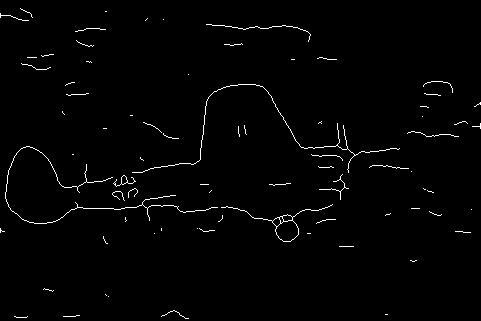} 
    & 
        \includegraphics[width=0.17\textwidth]{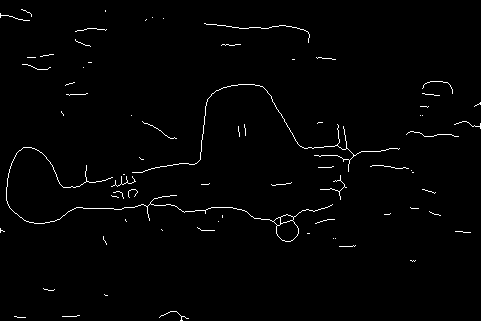} 
    &
        \includegraphics[width=0.17\textwidth]{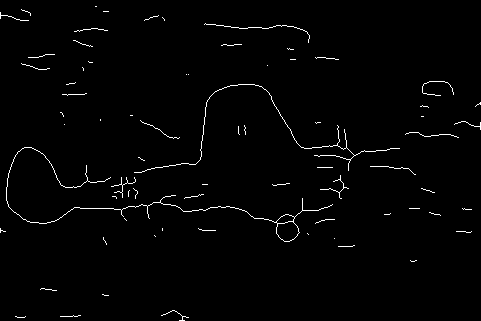} 
\\
        \includegraphics[width=0.17\textwidth]{result_images/first_order_compass/3063.jpg}
    &
        \includegraphics[width=0.17\textwidth]{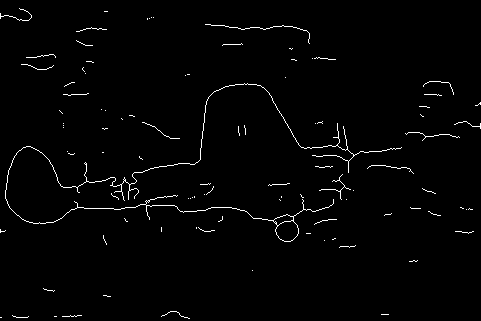} 
    & 
        \includegraphics[width=0.17\textwidth]{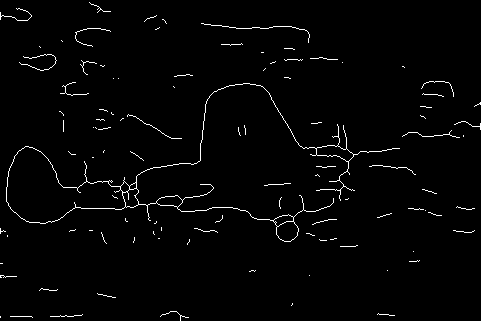} 
    &
        \includegraphics[width=0.17\textwidth]{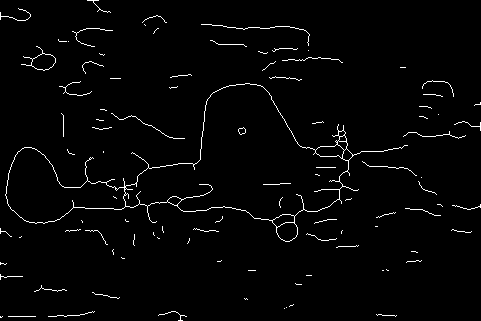} 
\\

        \includegraphics[width=0.17\textwidth]{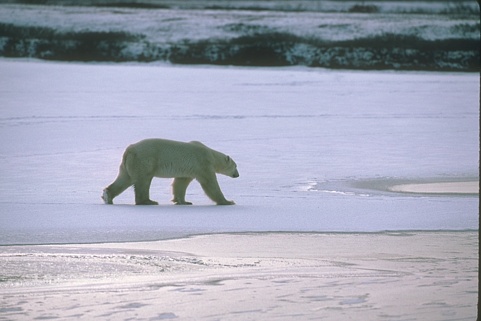}
    &
        \includegraphics[width=0.17\textwidth]{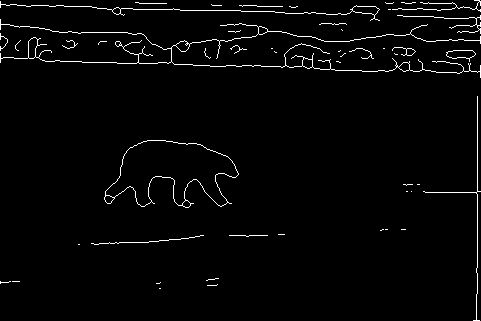} 
    & 
        \includegraphics[width=0.17\textwidth]{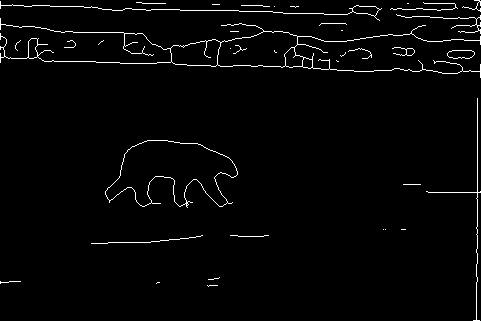} 
    &
        \includegraphics[width=0.17\textwidth]{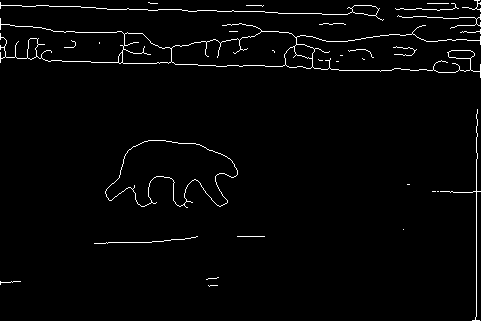} 
\\
        \includegraphics[width=0.17\textwidth]{result_images/frei-chen/100007.jpg}
    &
        \includegraphics[width=0.17\textwidth]{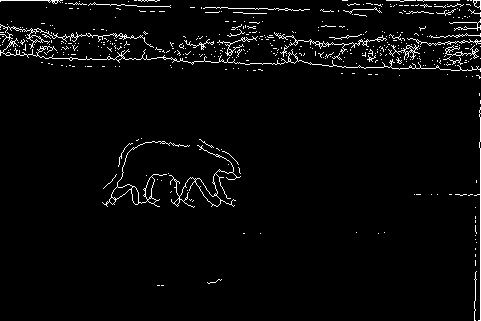} 
    & 
        \includegraphics[width=0.17\textwidth]{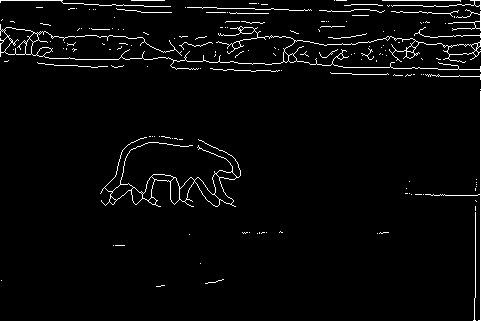} 
    &
        \includegraphics[width=0.17\textwidth]{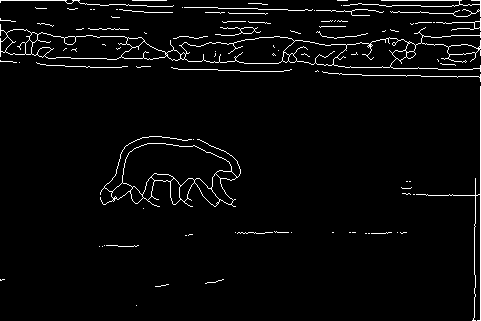} 

\end{tabular}
\caption{First Order Derivative Compass Gradient Operators and Frei-Chen Operator results; Columns: Original, 3x3, Dilated 5x5, Dilated 7x7; Rows:Robinson Cross, Robinson Modified Cross, Kirsch, Prewitt, Frei-Chen Edge, Frei-Chen Line}
\label{fig:first_order_compass_img_results}
\label{fig:frei_chen_img_results}
% \end{minipage}
\end{figure}%

Similar to gradient edge operators first we looked for the best smoothing parameter and threshold value to use. To do so we used the classical Robinson operator \cite{RobisonEdge1977} with a 3x3 kernel. In Figure \ref{fig:compass_thr_sigma_first_order_tunning} we can observe the $F1$ results for a range of the threshold between $30$ and $160$, sigma value between $0.25$ and $3.5$, obtaining the best results when threshold value is $50$ and sigma $2.5$. This is an important step because by choosing a wrong threshold value we can eliminate useful pixels from the edge map.

In Figure \ref{fig:compass_first_order_results} we can observe the \textit{F1-measure} results of all the edge operators presented in Section \ref{Sec:preliminari_first_compass} using the values found from our experiments, for the threshold value and sigma value. For detailed observation of the results we presented them in Table \ref{table:compass_results}.

As we can observe in the case of Compass Gradient operators dilation does not bring in all cases a better $F1$ but it seems that always brings a better \textit{P} metric. In case of Prewitt operator dilating the kernels actually brings overall worse results than the classical one.

\begin{table}[H]
% \begin{minipage}{.47\textwidth}
   \centering
        \setlength{\tabcolsep}{2pt}
        \scalebox{0.65}
        {
        \begin{tabular}{|ll|c|ccc|}
        \hline
        &Operator	&	&3x3	&Dilated 5x5	&Dilated 7x7	\\
        \hline
            &	            &R	&0.685	&0.673	&0.655\\
            &Robinson   	&P	&0.565	&0.574	&\textbf{0.585}\\
            &	            &F1	&0.620	&0.620	&0.618\\
            \hline
            &	            &R	&0.687	&0.670	&0.651\\
            &Robinson Mod.	&P	&0.565	&0.586	&\textbf{0.587}\\
            &	            &F1	&0.620	&0.620	&0.618\\
            \hline
            &	            &R	&0.689	&0.672	&0.653\\
            &Kirsch	        &P	&0.563	&0.572	&\textbf{0.581}\\
            &	            &F1	&0.619	&0.618	&0.615\\
            \hline
            &	            &R	&0.725	&0.738	&0.756\\
            &Prewitt	    &P	&\textbf{0.538}	&0.527	&0.495\\
            &	            &F1	&0.618	&0.615	&0.598\\
            \hline
            \hline
            &	            &R	&0.703	&0.716	&0.718\\
            &Frei-Chen edge	&P	&\textbf{0.53}	&0.523	&0.519\\
            &	            &F1	&0.605	&0.605	&0.603\\
            \hline
            &	            &R	&0.58	&0.597	&0.642\\
            &Frei-Chen line	&P	&0.482	&0.512	&0.506\\
            &	            &F1	&0.527	&0.551	&\textbf{0.566}\\
            \hline
        \end{tabular}}
   \caption{F1-measure results for First order derivative Template Gradients and Frei-Chen Edge Operator}
   \label{table:compass_results}
   \label{table:frei_chen_results}
% \end{minipage}
\end{table}

\subsection{Frei-Chen operator}
\label{Sec:results_frei_chen}

Frei-Chen operator is a special case of the Compass Gradient operators and is presented Section \ref{Sec:preliminari_frei_chen}. Using the Algorithm \ref{algorithm:steps_first_order} we run the experiments for edge and lines generated by this operator. In Figures \ref{fig:frei_chen_img_results} we present the visual results.

\begin{figure}[H]
    \centering
    \begin{minipage}{0.45\textwidth}
        \centering
        \includegraphics[scale=0.2]{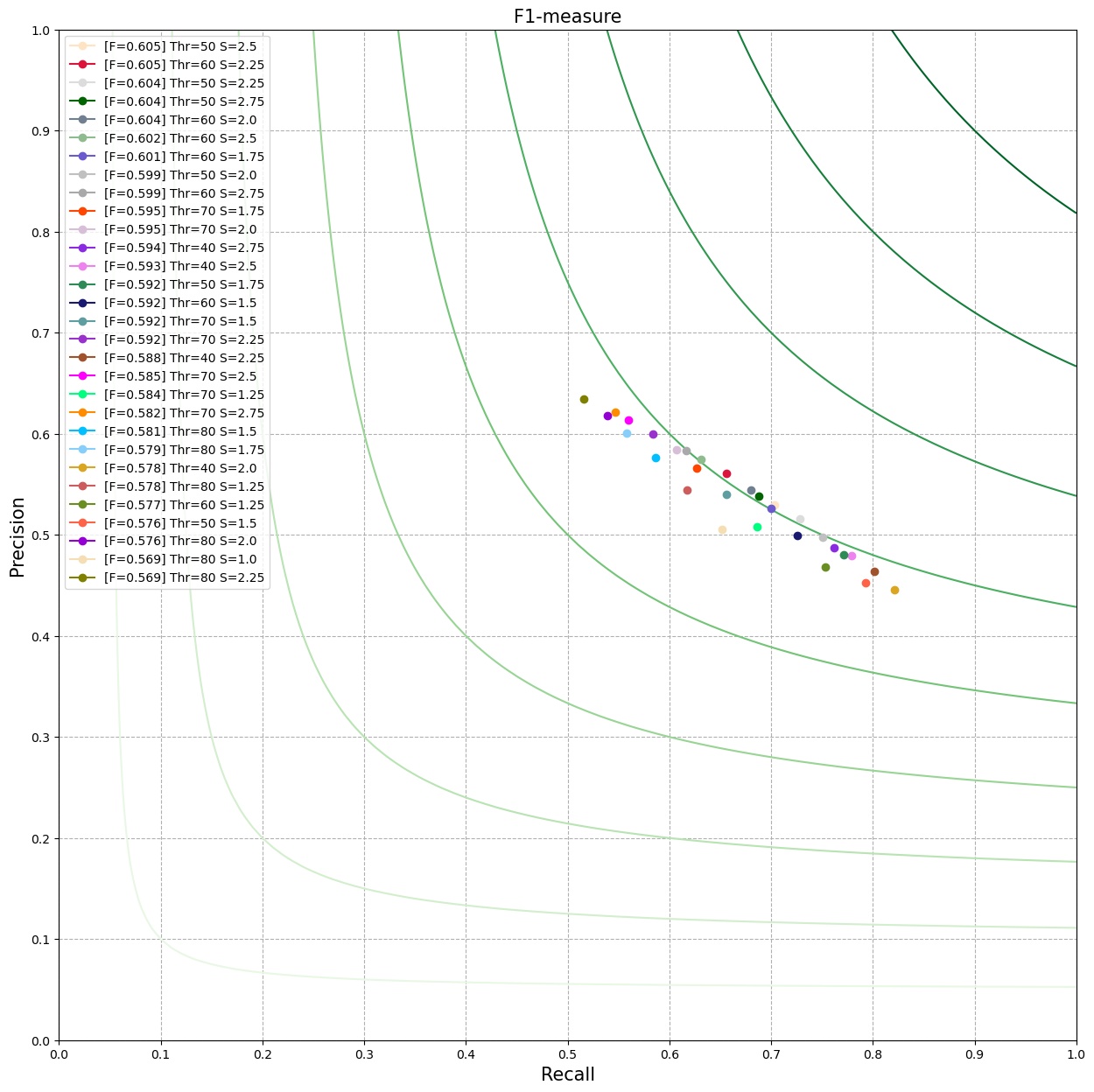} 
        \caption{Parameter tuning for Frei-Chen edge operator}
        \label{fig:compass_sig_thr_frei_chen_tunning}
    \end{minipage}\hfill
    \begin{minipage}{0.45\textwidth}
        \centering
        \includegraphics[scale=0.2]{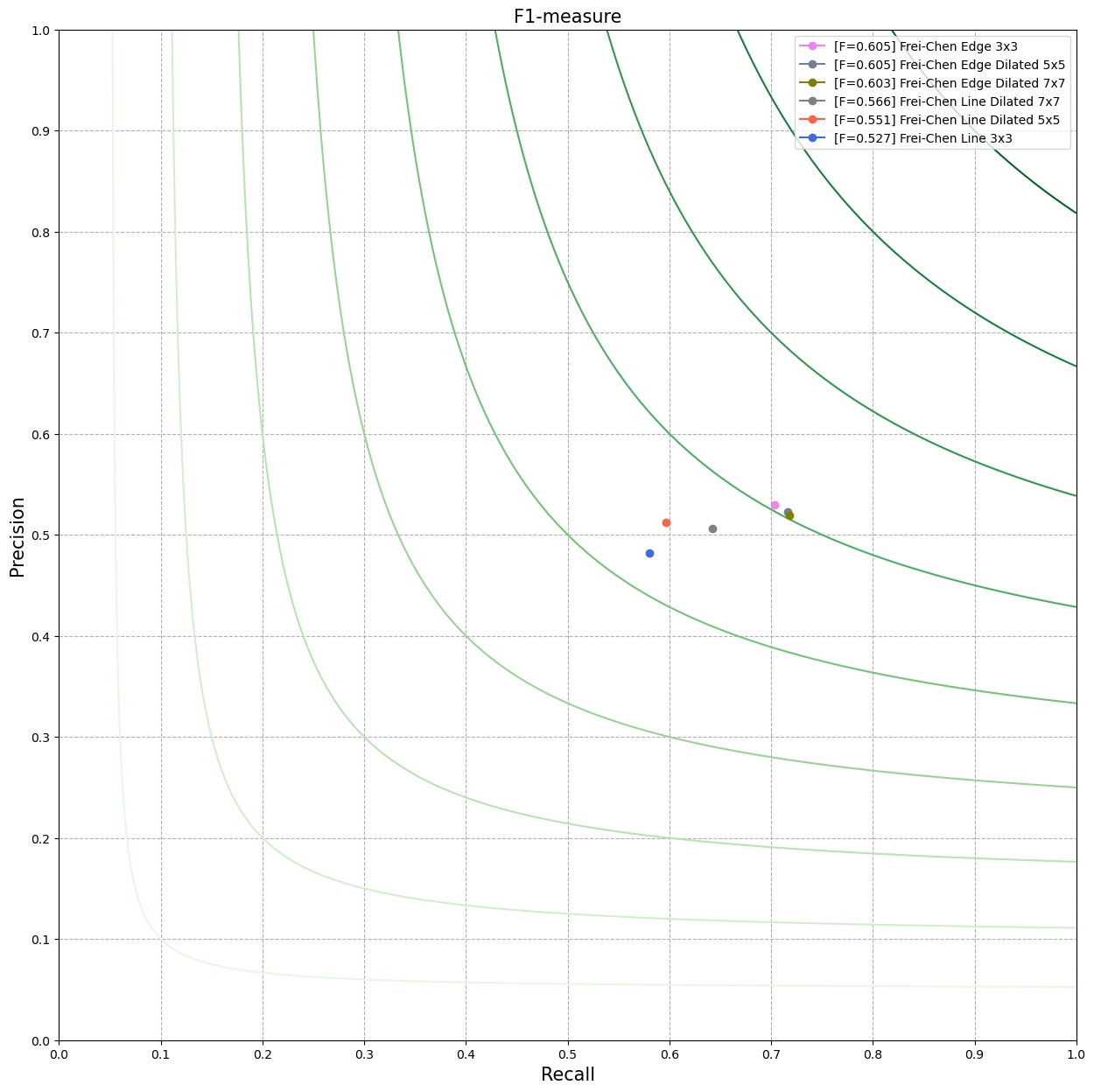} 
        \caption{Results for Frei-Chen edge operator}
        \label{fig:compass_frei_chen_results}
    \end{minipage}\hfill
\end{figure}

We fine tune the parameters of the algorithm for a range of the threshold between $30$ and $160$ and sigma value of the Gaussian Filter Blur between $0.25$ and $3.2$. As we observe the best tuned Frei-Chen operator is for threshold of $50$ and a sigma of $2.5$, see Figure \ref{fig:compass_sig_thr_frei_chen_tunning}.

In Figure \ref{fig:compass_frei_chen_results} we can observe the \textit{F1-measure} results of all the edge operators presented in Section \ref{Sec:results_frei_chen} using the values found from our experiments, for the threshold value and sigma value. For detailed observation of the results we presented them in Table \ref{table:frei_chen_results}.

Even if we can consider that the line detection part of the algorithm is not in scope for our paper we decided to add it to the simulation just to observe the effects which the dilation brings to it. As we can observe, similar to edge detection, line detection has a certain improvement gain, see the Table \ref{table:frei_chen_results} last row. We can not pass this topic without remarking the aspect of artifacts that are created when processing the line evaluation of the operator that are not resolved by dilating the kernels.

\subsection{Laplacian edge operator}

Laplace edge operator is one of the most popular edge detector based on second order derivative discrete kernels, described in Section \ref{Sec:preliminari_laplacian}. Following the Algorithm \ref{algorithm:laplace_steps} we attempt to standardize the evaluation of the edge detection, even if threshold of the second derivatives operator is not common practice in the field. 

\begin{figure}[H]
    \centering
    \begin{minipage}{0.45\textwidth}
        \centering
        \includegraphics[scale=0.2]{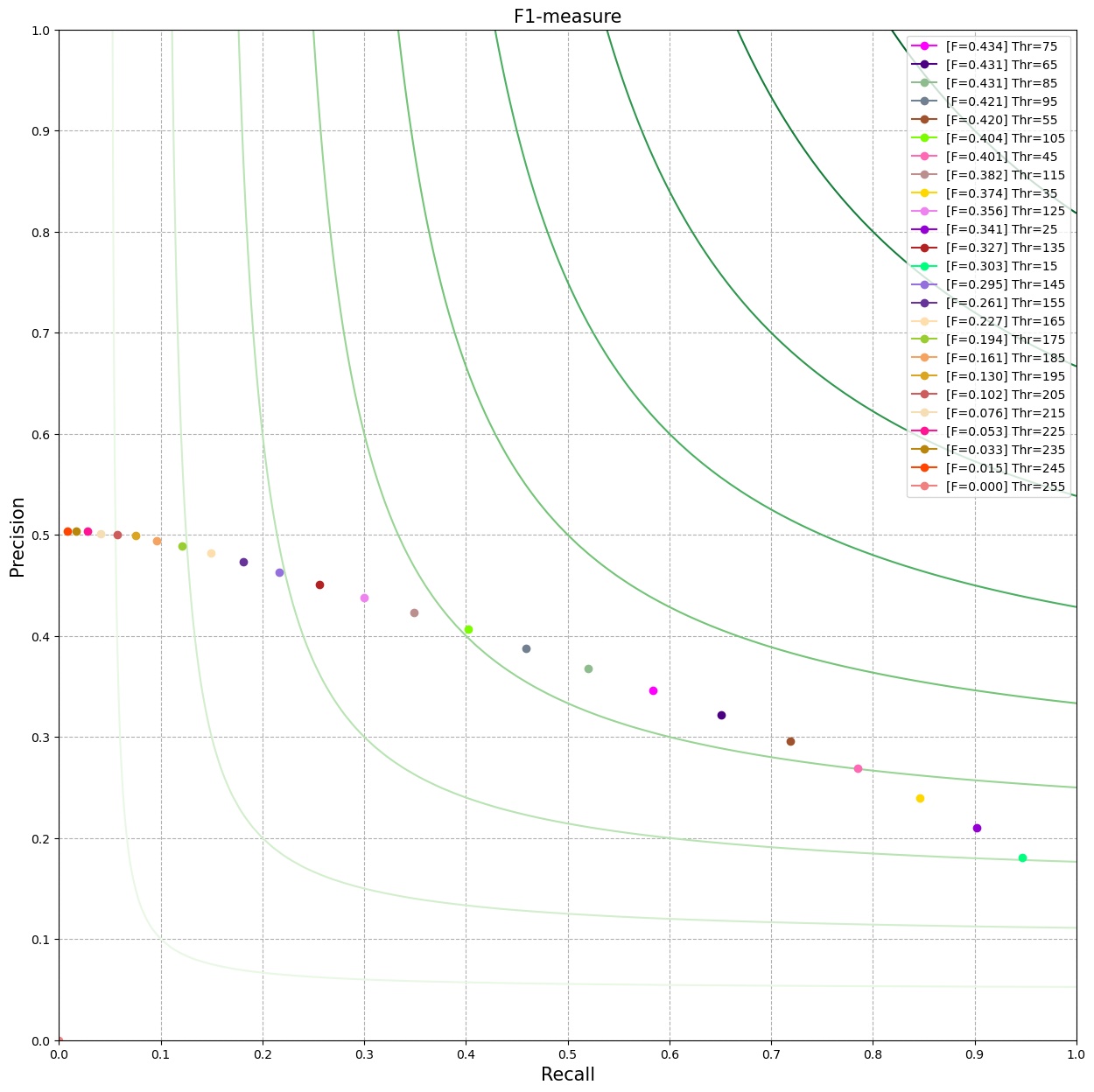} 
        \caption{Parameter tuning for Laplace edge operator}
        \label{fig:laplace_thr_tunning}
    \end{minipage}\hfill
    \begin{minipage}{0.45\textwidth}
        \centering
        \includegraphics[scale=0.2]{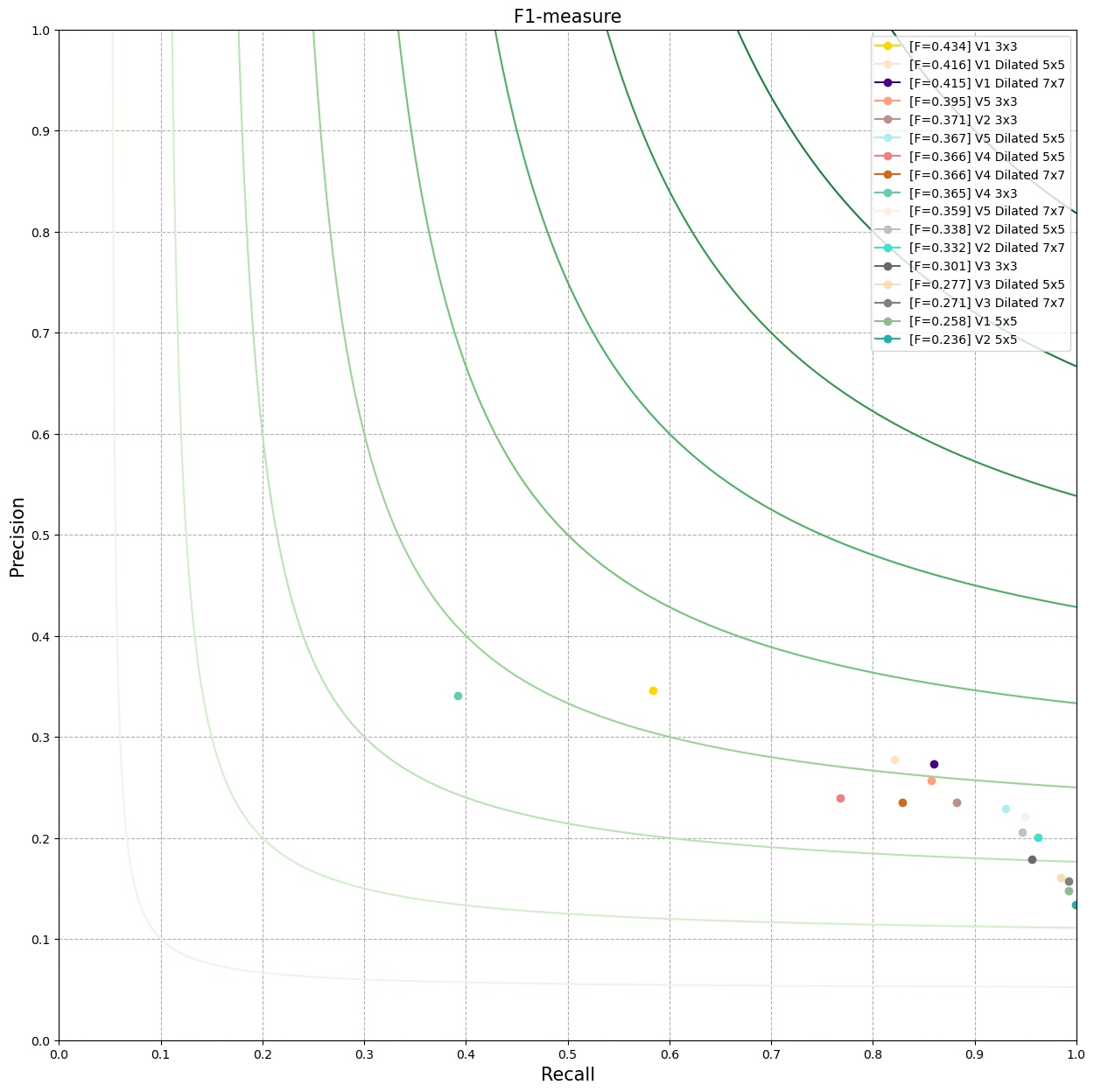} 
        \caption{Results Laplace edge operator}
        \label{fig:laplace_results}
    \end{minipage}\hfill
\end{figure}

For our simulation results and evaluation we first searched the best threshold value by using the Laplacian kernel V1, that we can find in Figure \ref{fig:laplace_kernel_masks}. We have chosen to vary the threshold from $15$ till $245$ and observed that we obtain the best $F1-measure$ measure for the value $75$, the results can be observed in Figure \ref{fig:laplace_thr_tunning}.

\begin{figure}[h!]
\begin{center}
\begin{minipage}{0.99\textwidth}
\centering
\setlength{\tabcolsep}{0.7pt}
\begin{tabular}{cccccccccc} 
        \includegraphics[scale=0.14]{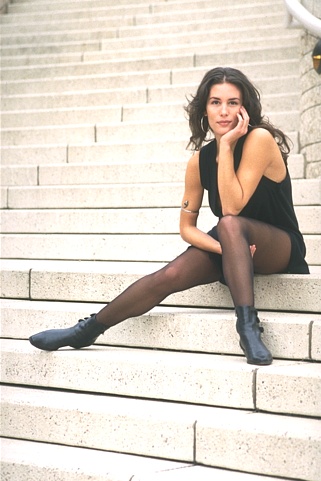}
    &
        \includegraphics[scale=0.14]{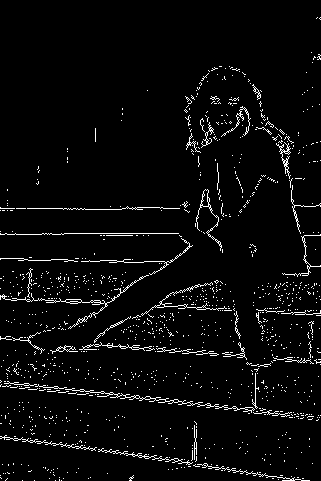} 
%   &
        % \includegraphics[scale=0.14]{result_images/laplace/FINAL_LAPLACE_V1_5x5_GREY_L0_181021.png} 
    & 
        \includegraphics[scale=0.14]{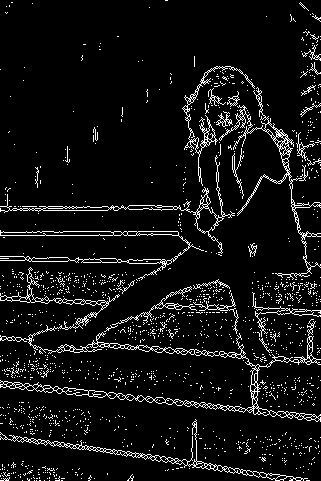} 
    &
        \includegraphics[scale=0.14]{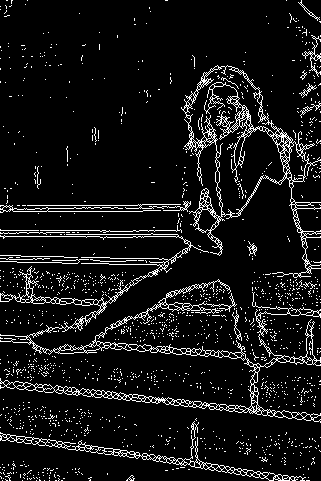} 
    &
        \includegraphics[scale=0.14]{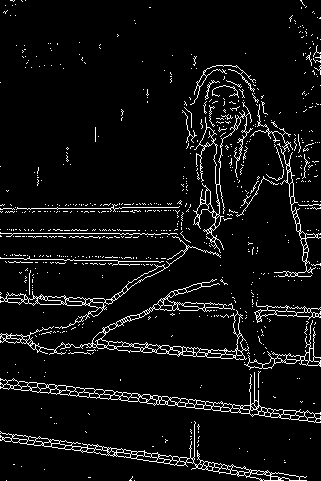} 
%   &
        % \includegraphics[scale=0.14]{result_images/log/FINAL_THR_5_LOG_LAPLACE_V1_5x5_S_1_8_GREY_L0_181021.png} 
    & 
        \includegraphics[scale=0.14]{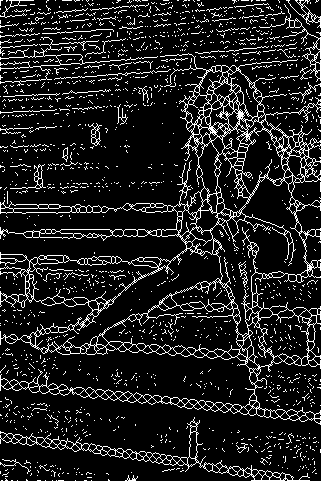} 
    &
        \includegraphics[scale=0.14]{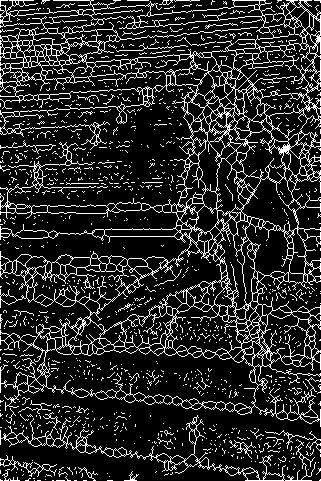} 
\\
        \includegraphics[scale=0.14]{result_images/laplace/181021.jpg}
    &
        \includegraphics[scale=0.14]{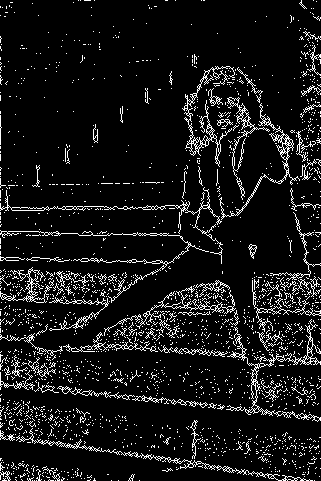} 
%   &
        % \includegraphics[scale=0.14]{result_images/laplace/FINAL_LAPLACE_V2_5x5_GREY_L0_181021.png} 
    & 
        \includegraphics[scale=0.14]{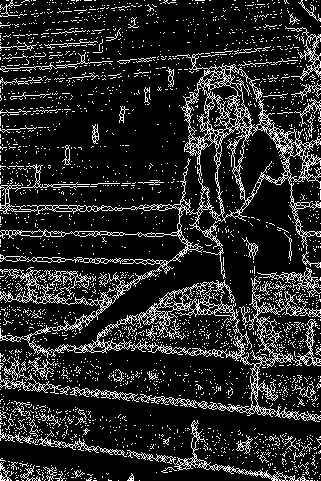} 
    &
        \includegraphics[scale=0.14]{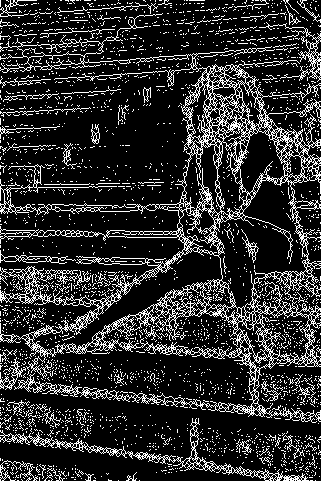} 
    &
        \includegraphics[scale=0.14]{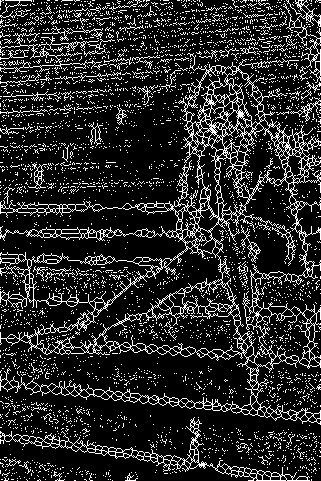} 
%   &
        % \includegraphics[scale=0.14]{result_images/log/FINAL_THR_5_LOG_LAPLACE_V2_5x5_S_1_8_GREY_L0_181021.png} 
    & 
        \includegraphics[scale=0.14]{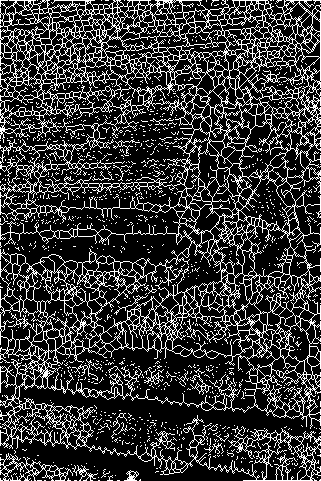} 
    &
        \includegraphics[scale=0.14]{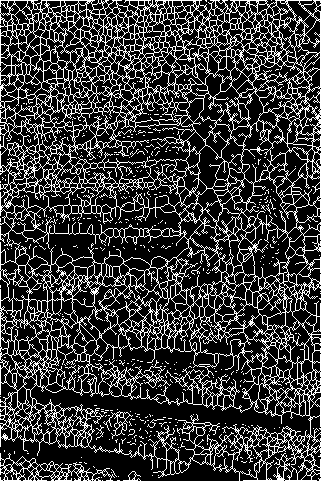} 
\\
        \includegraphics[scale=0.14]{result_images/laplace/181021.jpg}
    &
        \includegraphics[scale=0.14]{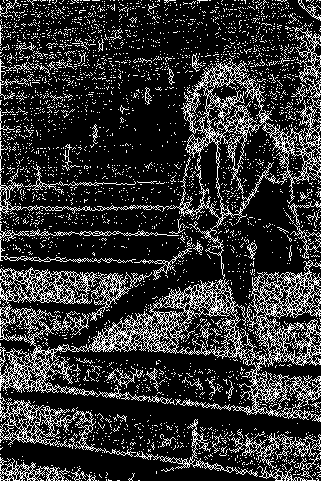} 
%   &
%         ~
    & 
        \includegraphics[scale=0.14]{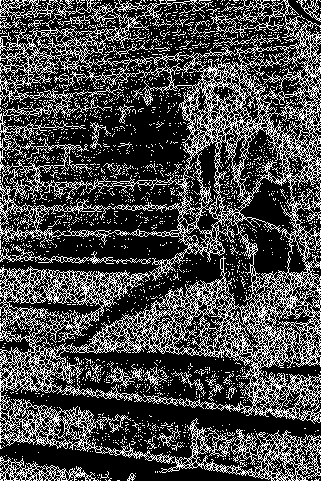} 
    &
        \includegraphics[scale=0.14]{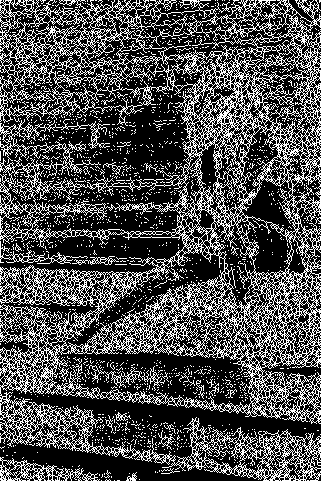} 
    &
        \includegraphics[scale=0.14]{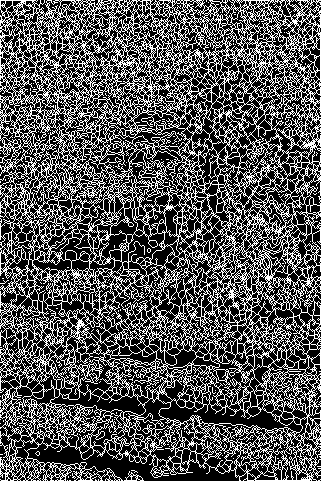} 
%   &
%         ~ 
    & 
        \includegraphics[scale=0.14]{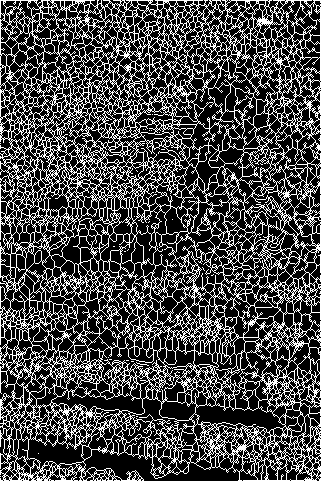} 
    &
        \includegraphics[scale=0.14]{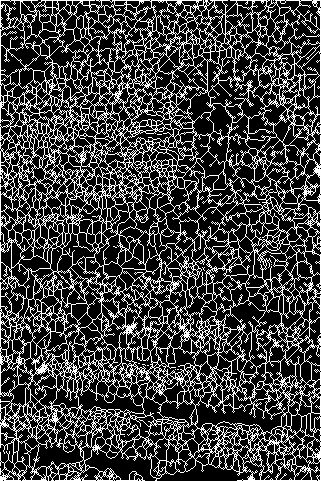} 
\\
        \includegraphics[scale=0.14]{result_images/laplace/181021.jpg}
    &
        \includegraphics[scale=0.14]{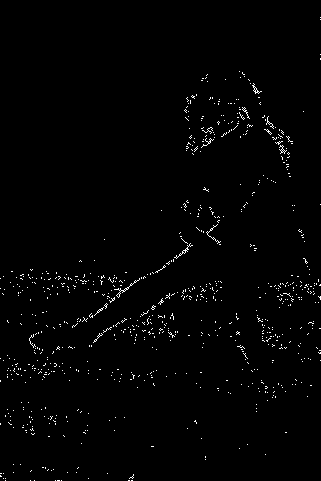} 
%   &
%         ~
    & 
        \includegraphics[scale=0.14]{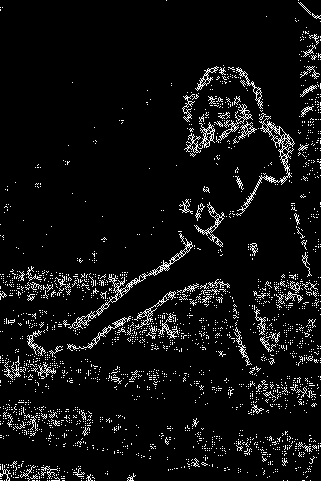} 
    &
        \includegraphics[scale=0.14]{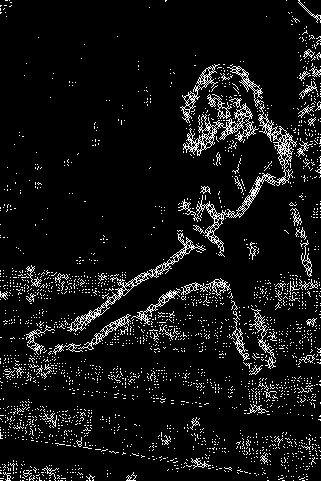} 
    &
        \includegraphics[scale=0.14]{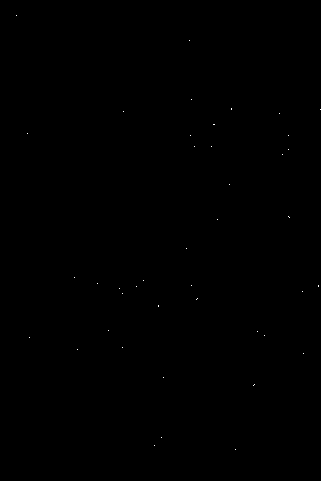} 
%   &
%         ~
    & 
        \includegraphics[scale=0.14]{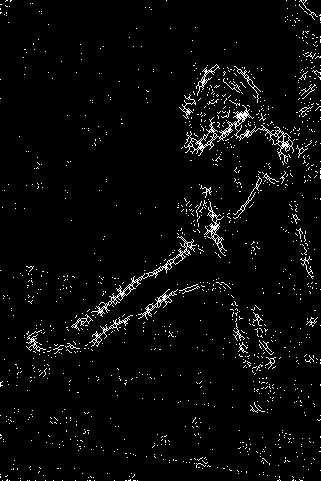} 
    &
        \includegraphics[scale=0.14]{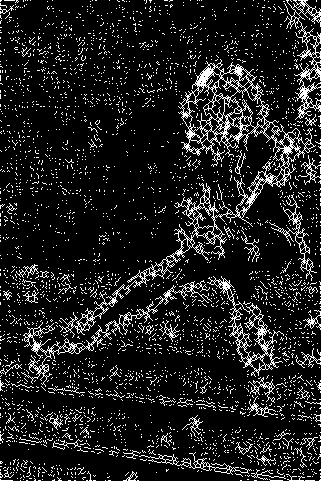} 
\\
        \includegraphics[scale=0.14]{result_images/laplace/181021.jpg}
    &
        \includegraphics[scale=0.14]{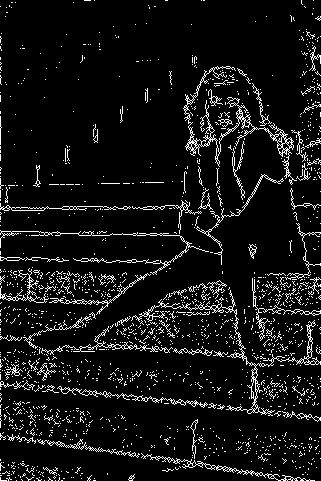} 
%   &
%         ~
    & 
        \includegraphics[scale=0.14]{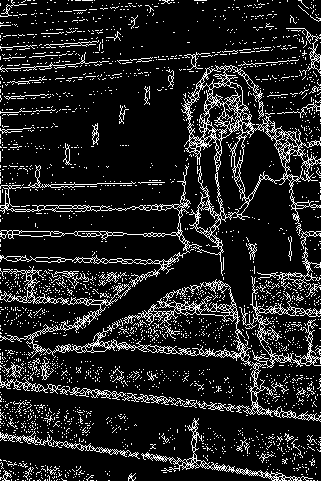} 
    &
        \includegraphics[scale=0.14]{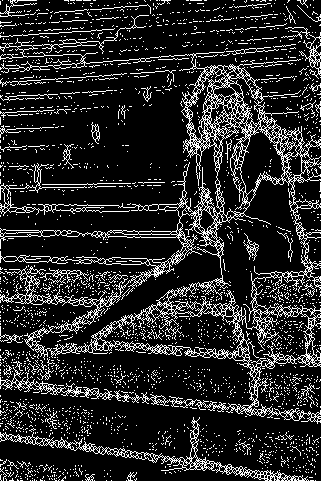} 
    &
        \includegraphics[scale=0.14]{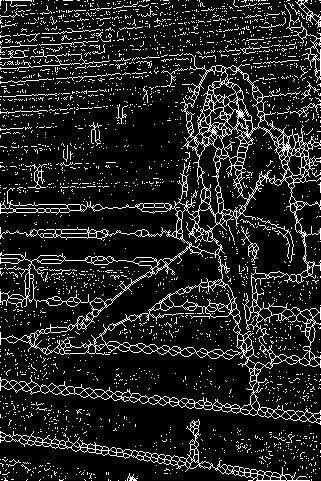} 
%   &
%         ~ 
    & 
        \includegraphics[scale=0.14]{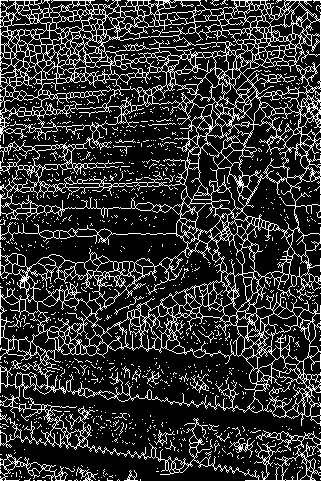} 
    &
        \includegraphics[scale=0.14]{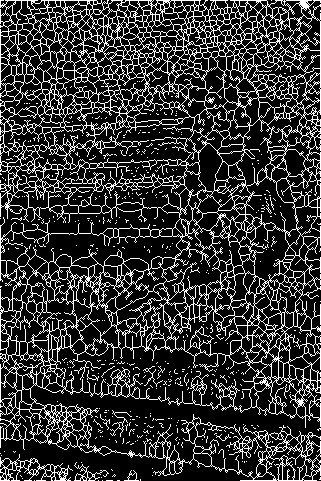} 
\end{tabular}

\caption{Laplace operators and Laplace of Gaussian results.\\ Columns: Original, Laplace 3x3, Laplace Dilated 5x5, Laplace Dilated 7x7, LoG 3x3, LoG 5x5, LoG Dilated 5x5, LoG Dilated 7x7 Rows: Laplace kernel V1, Laplace kernel V2, Laplace kernel V3, Laplace kernel V4, Laplace kernel V5}
\label{fig:laplace_img_results}
\label{fig:log_img_results}
\end{minipage}
\end{center}
\end{figure}

In Figure \ref{fig:laplace_img_results} we present the visual results for our experiments, the Figure \ref{fig:laplace_results} and Table \ref{table:laplace_results} contains the benchmark over all results. We can observe that in this case, dilation of kernel technique does not bring improvements to the edge output. We can see a duplication of the edge points detected that seems to be caused by the a number of edge points with low value found that are being highlighted more when we threshold, normalize and thin the edge map.

Even all the kernels found in literature are different approximation of the Laplace function we would like to see what effect this different kernels bring in the edge-maps produced. For all our experimentes we will use only the positive variants of the operator, so we will produce inwards edges.

\subsection{Laplacian of Gaussian - LoG - or Mexican Hat Operator}

A natural extension of the Laplace Operator is the Laplacian of Log, described in Section \ref{Sec:preliminari_log}. 

\begin{figure}[!h]
    \centering
    \begin{minipage}{0.45\textwidth}
        \centering
        \includegraphics[scale=0.2]{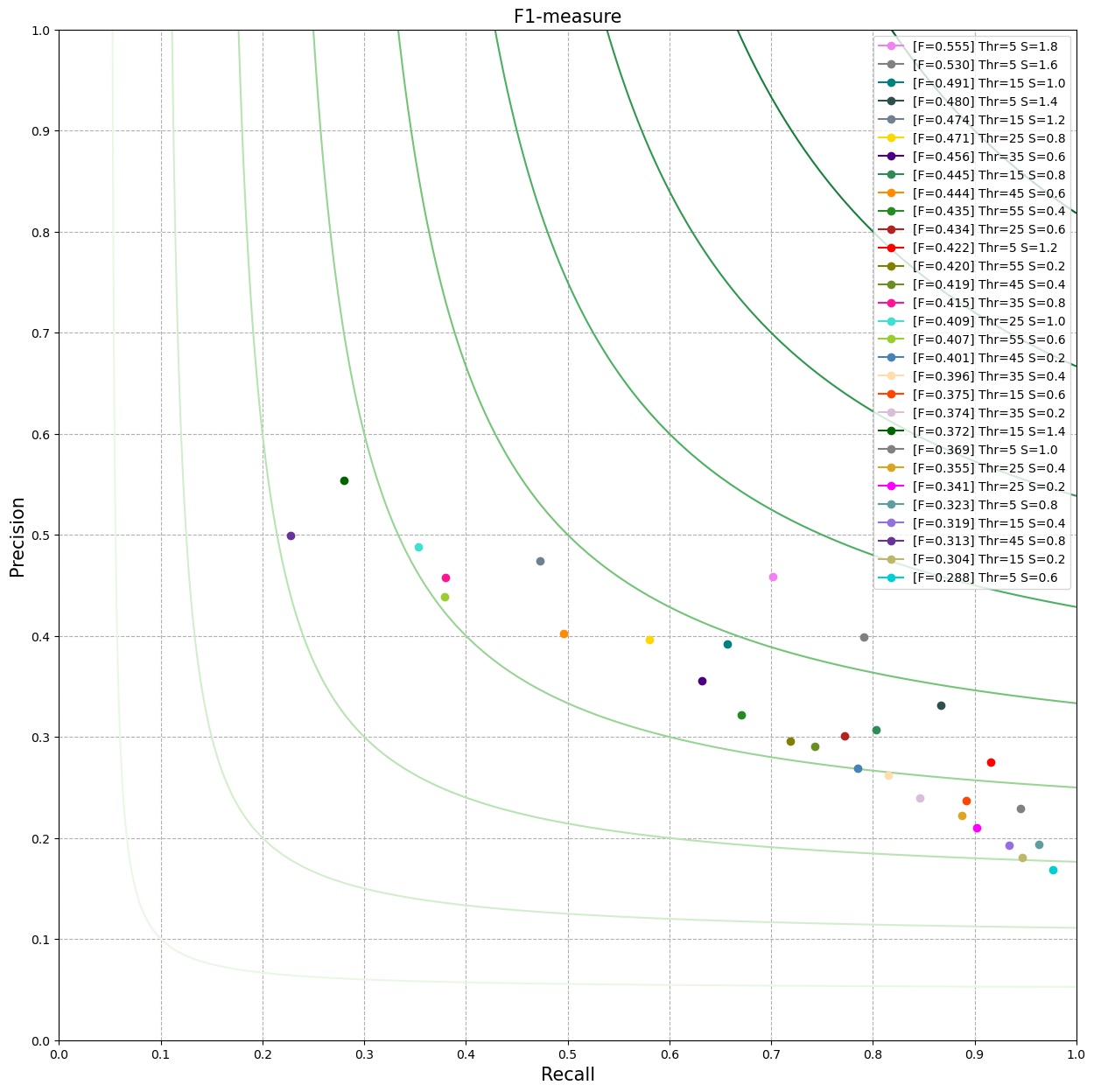} 
        \caption{Parameter tuning for LoG edge operator}
        \label{fig:log_tunning}
    \end{minipage}\hfill
    \begin{minipage}{0.45\textwidth}
        \centering
        \includegraphics[scale=0.2]{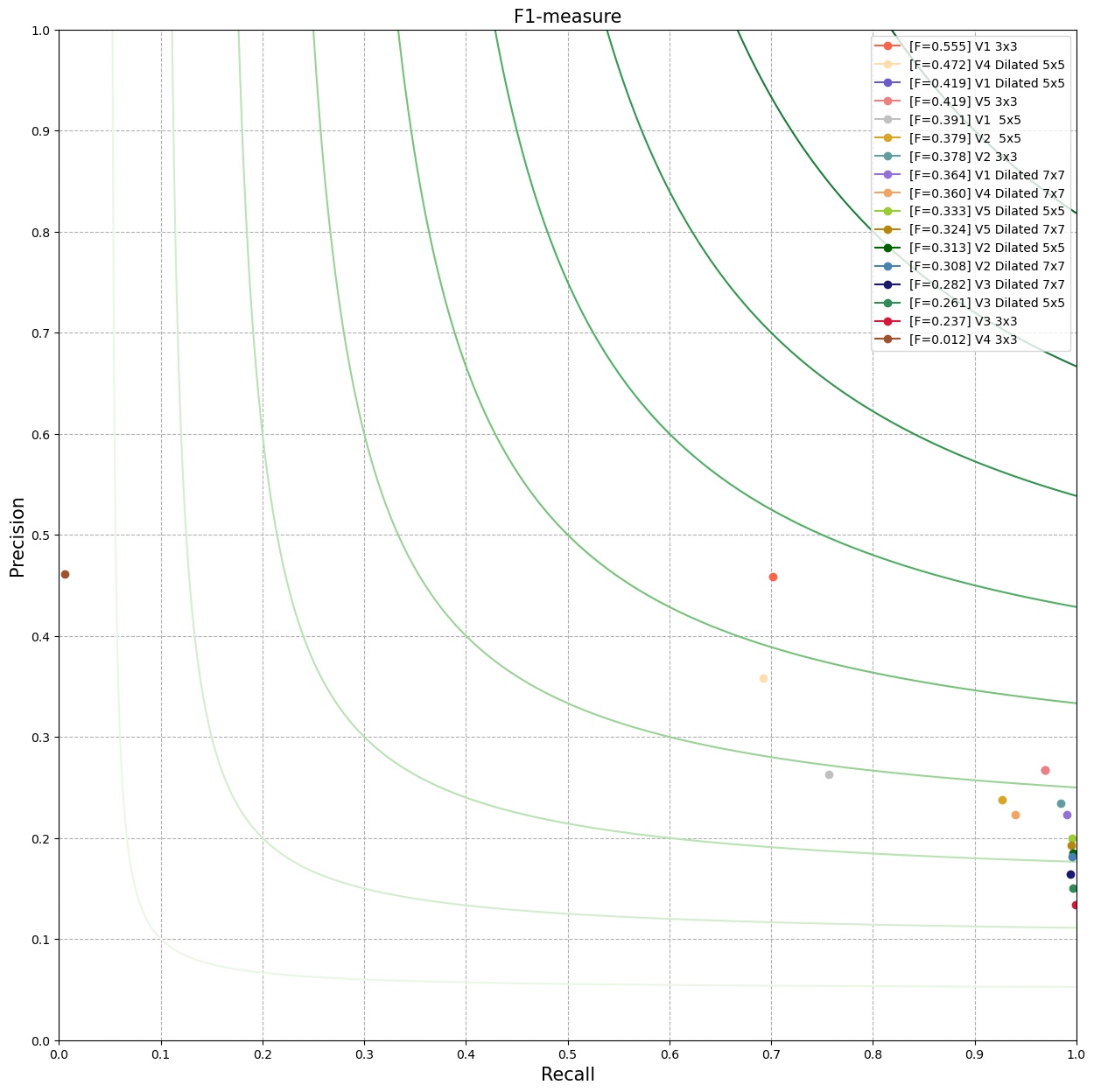} 
        \caption{Results for LoG edge operator}
        \label{fig:log_results}
    \end{minipage}\hfill
\end{figure}

For our simulation results and evaluation concerning Mexican Hat Operator we described in section \ref{Sec:preliminari_log} we first searched for the best combination of sigma for constructing the LoG kernel and the threshold. The results can be observed in Figure \ref{fig:log_tunning}. We have chosen to vary the threshold from $5$ till $60$, avoiding going for a bigger threshold. Bigger threshold values would not generate better results because of the small intensity resulted after the convolution. Regarding the sigma for the Gaussian Blur kernel, we chose to vary between $0.2$ and $2.0$, while going higher would generate a lack of resulting edge points.

\begin{table}[H]
\centering
\setlength{\tabcolsep}{2pt}
\scalebox{0.68}
{
\begin{tabular}{|l|c|cccc|cccc|}
\hline
\multicolumn{2}{|c|}{\bfseries Operator} &\multicolumn{4}{c|}{\bfseries Laplace}&\multicolumn{4}{c|}{\bfseries LoG}\\
\hline
	&	&3x3	&5x5	&Dilated 5x5	&Dilated 7x7	&3x3	&5x5	&Dilated 5x5	&Dilated 7x7	\\
\hline
	&R	&0.584	&0.992	&0.821	&0.860	&0.702	&0.757	&0.969	&0.991	\\
V1	&P	&0.346	&0.148	&0.278	&0.273	&0.459	&0.263	&0.267	&0.223	\\
	&F1	&\textbf{0.434}	&0.258	&0.416	&0.415	&\textbf{0.555}	&0.391	&0.419	&0.364	\\
\hline
	&R	&0.882	&0.999	&0.947	&0.962	&0.985	&0.927	&0.997	&0.996	\\
V2	&P	&0.235	&0.134	&0.206	&0.201	&0.234	&0.238	&0.185	&0.182	\\
	&F1	&\textbf{0.371}	&0.236	&0.338	&0.332	&0.378	&\textbf{0.379}	&0.313	&0.308	\\
\hline
	&R	&0.956	&-	&0.985	&0.992	&0.999	&-	&0.997	&0.994	\\
V3	&P	&0.179	&-	&0.161	&0.157	&0.134	&-	&0.150	&0.164	\\
	&F1	&\textbf{0.301}	&-	&0.277	&0.271	&0.237	&-	&0.261	&\textbf{0.282}	\\
\hline
	&R	&0.392	&-	&0.768	&0.829	&0.006	&-	&0.692	&0.940	\\
V4	&P	&0.341	&-	&0.24	&0.235	&0.461	&-	&0.358	&0.223	\\
	&F1	&0.365	&-	&\textbf{0.366}	&0.366	&0.012	&-	&\textbf{0.472}	&0.360	\\
\hline
	&R	&0.857	&-	&0.93	&0.949	&0.969	&-	&0.996	&0.995	\\
V5	&P	&0.257	&-	&0.229	&0.221	&0.267	&-	&0.200	&0.193	\\
	&F1	&\textbf{0.395}	&-	&0.367	&0.359	&\textbf{0.419}	&-	&0.333	&0.324	\\
\hline
\end{tabular}}
\vspace{1.5pt}
\caption{Results of Laplace Operator and LoG Operator}
\label{table:laplace_results}
\label{table:log_results}
\end{table}%

We observed that we obtain the best $F1-measure$ measure by using the values of $5$ for the threshold and $1.8$ for sigma. For a better visualization in Figure \ref{fig:log_tunning} we have chosen to shown only the best results.

As we can see in Figure \ref{fig:log_results} and Figure \ref{fig:log_img_results} dilating the filters does not bring forward better results for most of the kernel variants but if we look closer in Table \ref{table:log_results} we see cases like in $V3$ and $V4$ dilating the kernels bring much better results. 

\subsection{Marr–Hildreth Operator}

\begin{figure}[H]
    \centering
    \begin{minipage}{0.45\textwidth}
        \centering
        \includegraphics[scale=0.2]{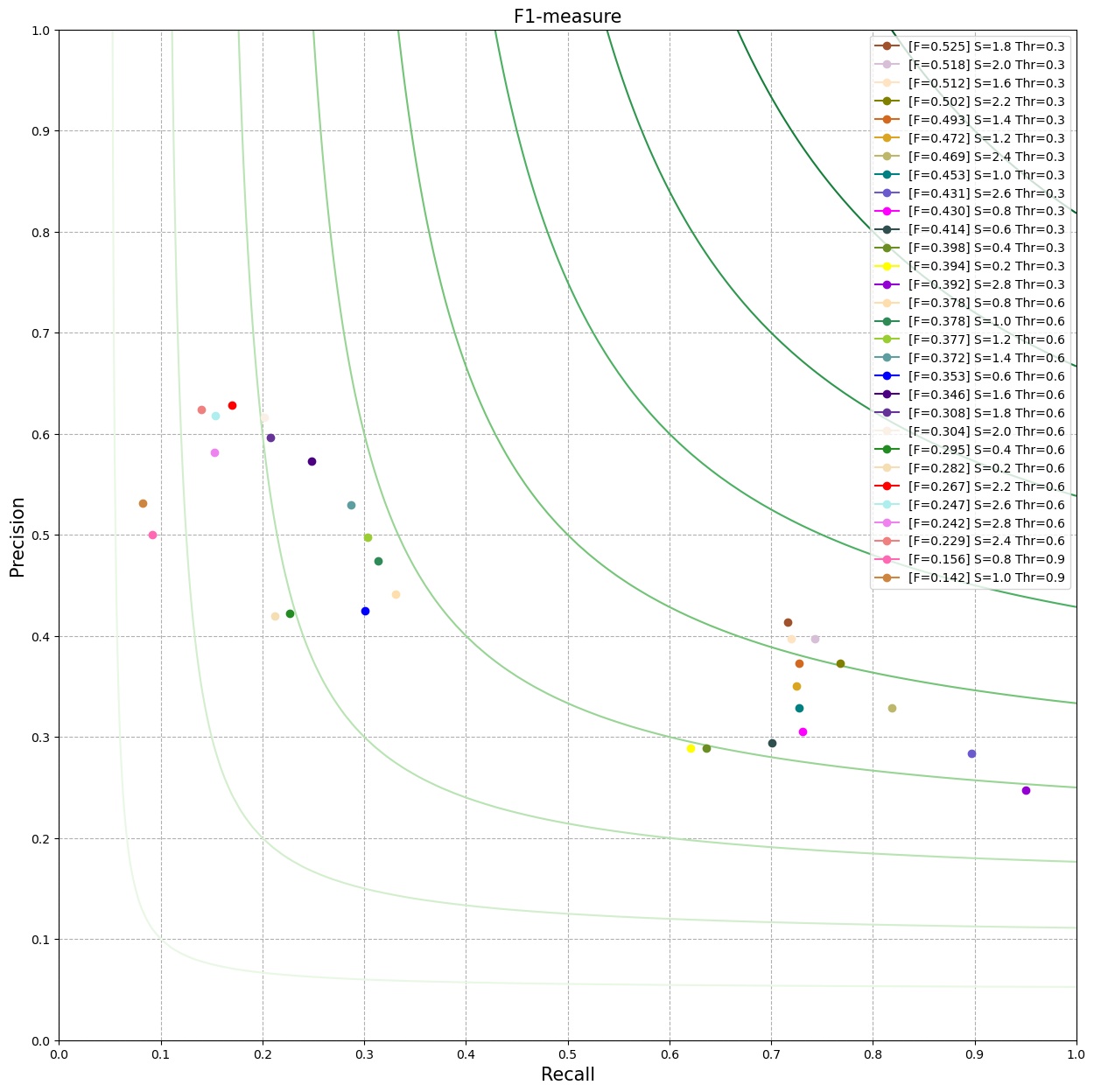} 
        \caption{Parameter tuning for Marr-Hildreth edge operator}
        \label{fig:marr_sigma_results}
    \end{minipage}\hfill
    \begin{minipage}{0.45\textwidth}
        \centering
        \includegraphics[scale=0.2]{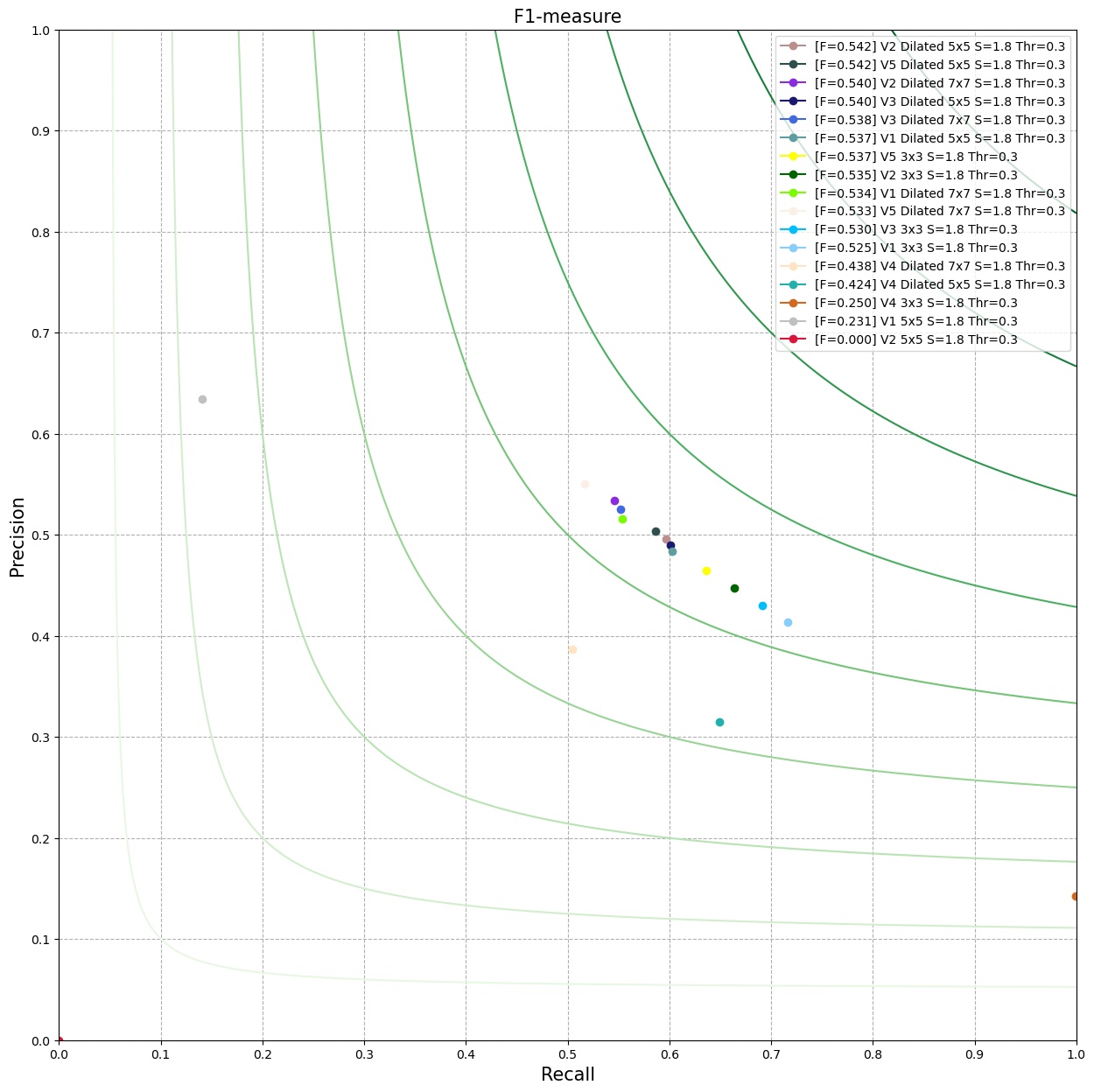} 
        \caption{Results of Marr-Hildreth edge operator}
        \label{fig:marr_edge_results}
    \end{minipage}\hfill
\end{figure}

Evaluation of Marr-Hildreth Operator, presented in section \ref{Sec:preliminari_marr}, are similar to the Log Operator. We first searched the best combination of sigma for constructing the LoG kernel and the threshold of Zero Crossing. The results can be observed in Figure \ref{fig:marr_edge_results} and for a better visualization, we have chosen to show only the best results in $F1$ order.

We have tried the combination of $sigma$ value from $0.2$ to $3.0$ with gradient threshold of $30$ to $90$ of the gradient image, considering that is the interval where these parameters bring good results to the output. We obtain the best results  when $sigma$ is $1.8$ and the threshold is $0.3$ * $255$ = $85$.

As we can see in Figure \ref{fig:marr_sigma_results} and Figure \ref{fig:marr_edge_results} dilating the Laplace kernels brings improvements to all variants. The Table \ref{table:marr_results} contains the full Marr-Hildreth results were in this case the dilated filter has obtained better results, see the \textit{P} and \textit{F1-score} of variants 1 to 4. For V5 variant the results are closer to the original filter results.  

\begin{figure}[!h]
% \begin{minipage}{0.47\textwidth}
\centering 
\setlength{\tabcolsep}{0.5pt}
\begin{tabular}{cccccccccc} 
        \includegraphics[scale=0.13]{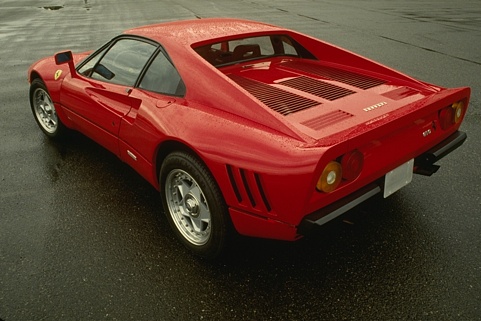}
    &
        \includegraphics[scale=0.13]{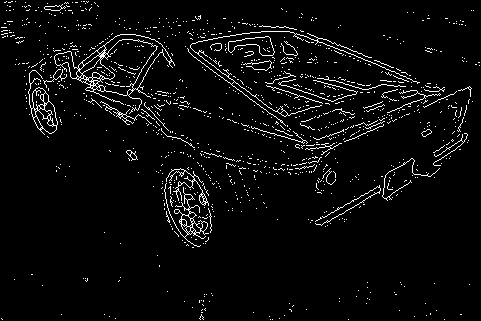} 
%   &
%         \includegraphics[scale=0.13]{result_images/marr/FINAL_MARR_HILDRETH_LAPLACE_V1_5x5_S_1_8_THR_0_3_GREY_L0_29030.png} 
    & 
        \includegraphics[scale=0.13]{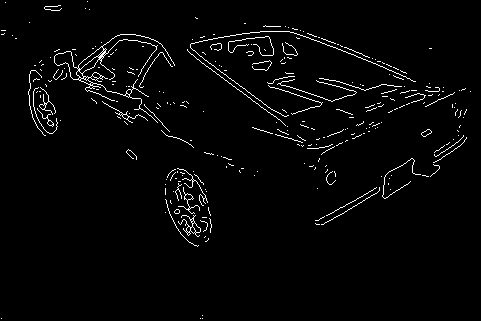} 
    &
        \includegraphics[scale=0.13]{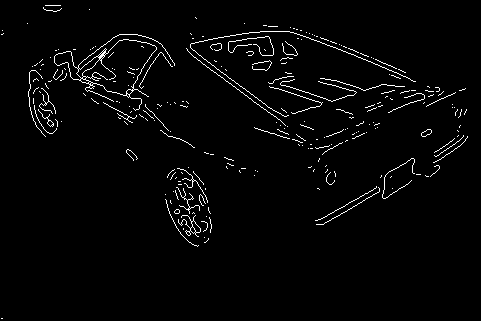} 
    &
        \includegraphics[scale=0.13]{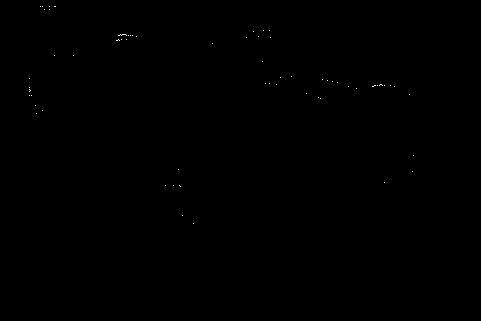} 
%   &
%         \includegraphics[scale=0.13]{result_images/shen/SHEN_CASTAN_LAPLACE_V1_5x5_L0_29030.png} 
    & 
        \includegraphics[scale=0.13]{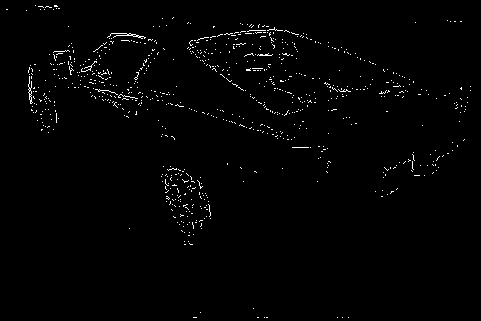} 
    &
        \includegraphics[scale=0.13]{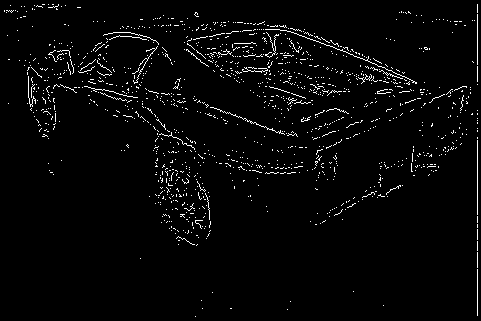} 
\\
        \includegraphics[scale=0.13]{result_images/marr/29030.jpg}
    &
        \includegraphics[scale=0.13]{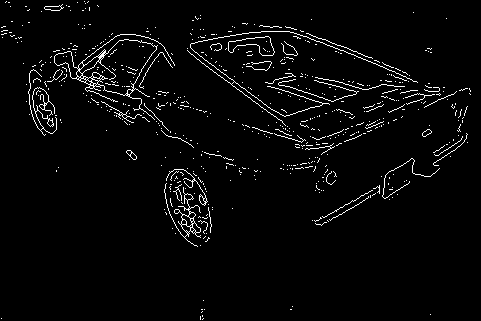} 
%   &
%         \includegraphics[scale=0.13]{result_images/marr/FINAL_MARR_HILDRETH_LAPLACE_V2_5x5_S_1_8_THR_0_3_GREY_L0_29030.png} 
    & 
        \includegraphics[scale=0.13]{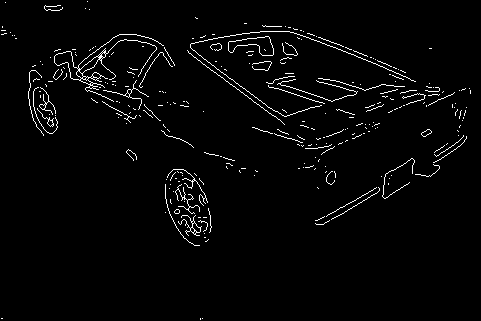} 
    &
        \includegraphics[scale=0.13]{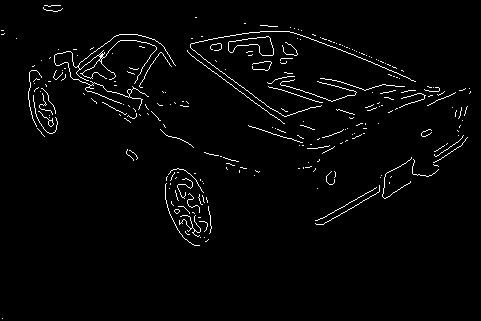} 
    &
        \includegraphics[scale=0.13]{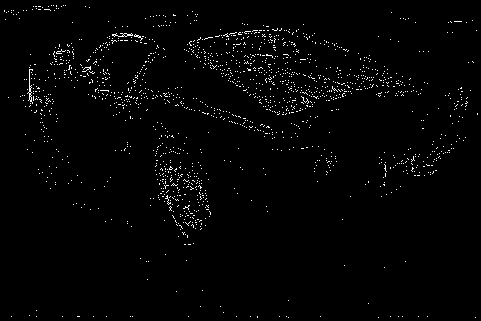} 
%   &
%         \includegraphics[scale=0.13]{result_images/shen/SHEN_CASTAN_LAPLACE_V2_5x5_L0_29030.png} 
    & 
        \includegraphics[scale=0.13]{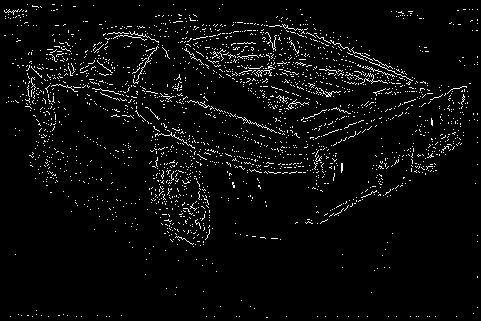} 
    &
        \includegraphics[scale=0.13]{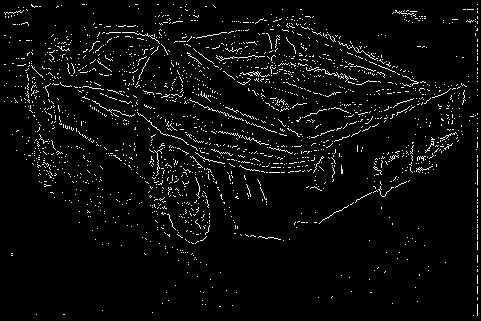} 
\\
        \includegraphics[scale=0.13]{result_images/marr/29030.jpg}
    &
        \includegraphics[scale=0.13]{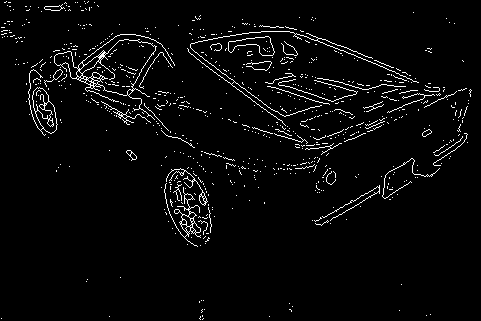} 
%   &
%         ~
    & 
        \includegraphics[scale=0.13]{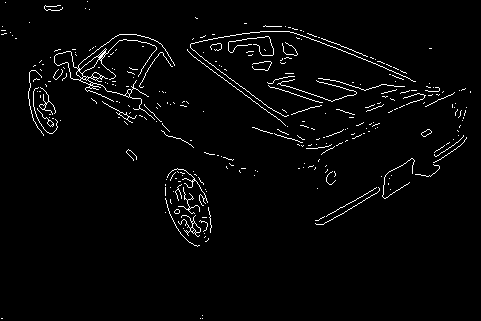} 
    &
        \includegraphics[scale=0.13]{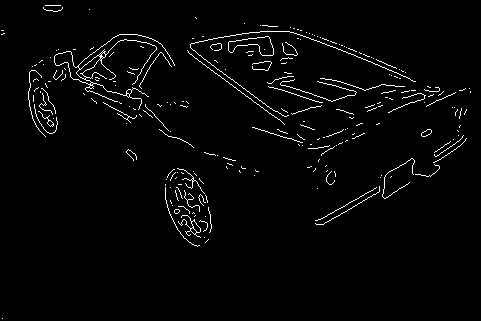} 
    &
        \includegraphics[scale=0.13]{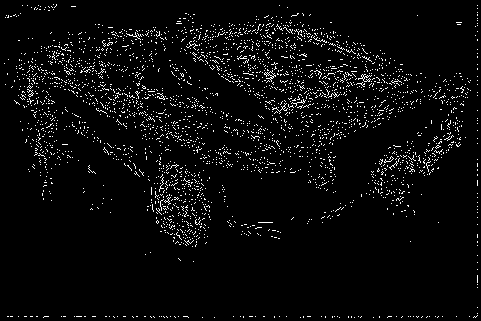} 
%   &
%         ~ 
    & 
        \includegraphics[scale=0.13]{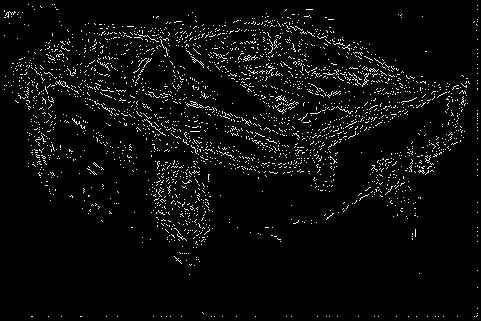} 
    &
        \includegraphics[scale=0.13]{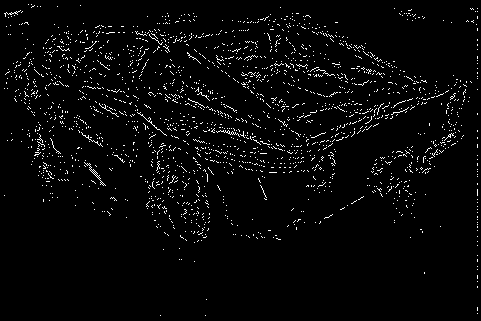} 
\\
        \includegraphics[scale=0.13]{result_images/marr/29030.jpg}
    &
        \includegraphics[scale=0.13]{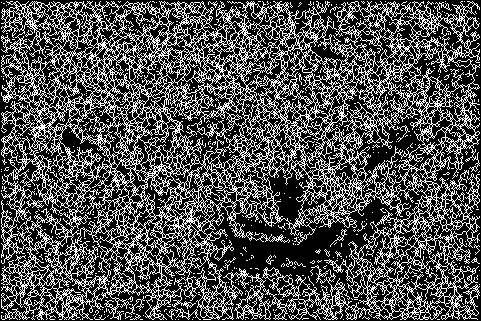} 
%   &
%         ~
    & 
        \includegraphics[scale=0.13]{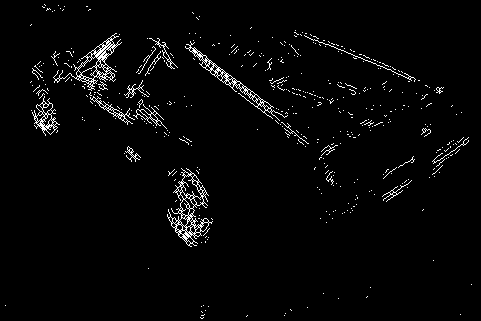} 
    &
        \includegraphics[scale=0.13]{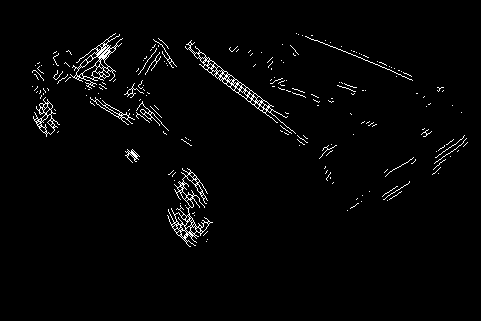} 
    &
        \includegraphics[scale=0.13]{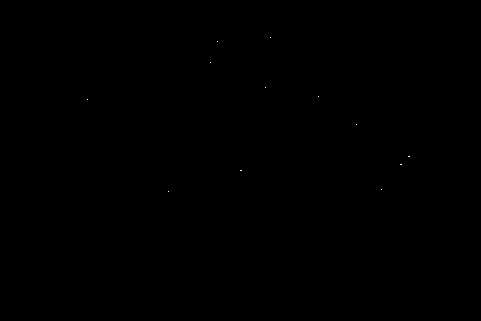} 
%   &
%         ~
    & 
        \includegraphics[scale=0.13]{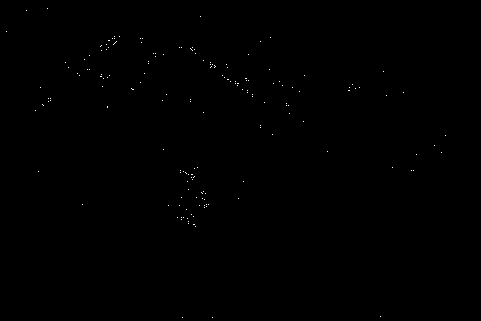} 
    &
        \includegraphics[scale=0.13]{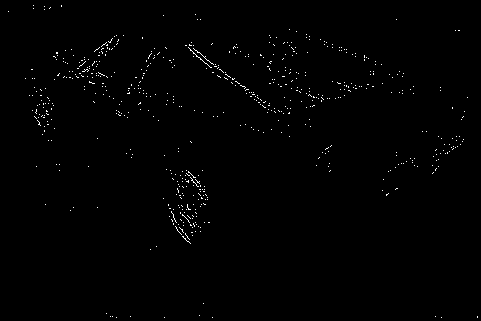} 
\\
        \includegraphics[scale=0.13]{result_images/marr/29030.jpg}
    &
        \includegraphics[scale=0.13]{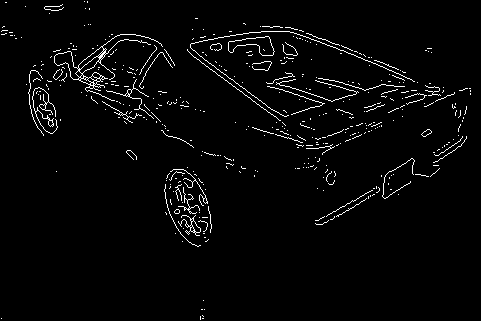} 
%   &
%         ~
    & 
        \includegraphics[scale=0.13]{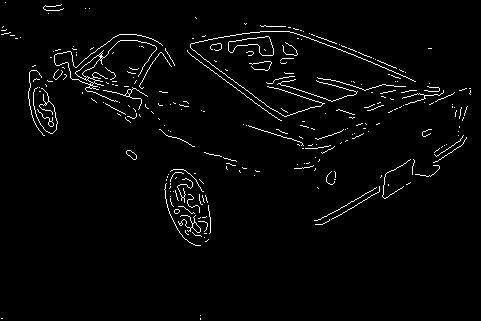} 
    &
        \includegraphics[scale=0.13]{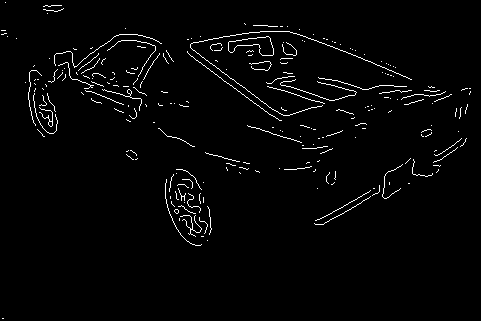}  
    &
        \includegraphics[scale=0.13]{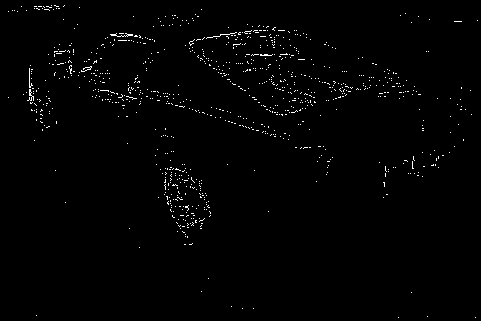} 
%   &
%         ~ 
    & 
        \includegraphics[scale=0.13]{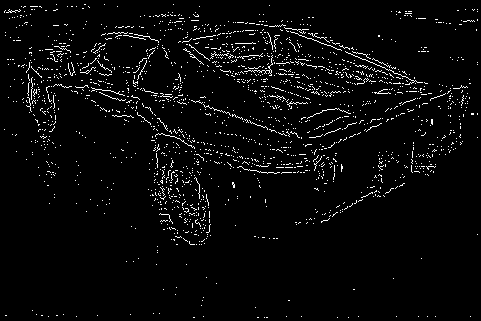} 
    &
        \includegraphics[scale=0.13]{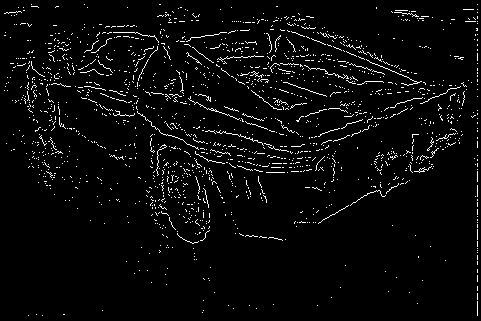} 
\end{tabular}

\caption{Marr-Hildreth operators and Shen-Castan operator results; Columns: Original, MH kernel 3x3, MH kernel dilated 5x5, MH kernel dilated 7x7, SC kernel 3x3, SC kernel dilated 5x5, SC kernel dilated 7x7; Rows: Laplace kernel V1, Laplace kernel V2, Laplace kernel V3, Laplace kernel V4, Laplace kernel V5}
\label{fig:marr_img_results}
\label{fig:shen_img_results}
% \end{minipage}
\end{figure}%

\subsection{Canny algorithm}

The following section consists of the analysis of the Canny results using first order derivative gradient operators, extended and dilated filters presented in Section \ref{Sec:preliminari_canny}. 

\begin{figure}[H]
    \centering
    \begin{minipage}{0.45\textwidth}
        \centering
        \includegraphics[scale=0.2]{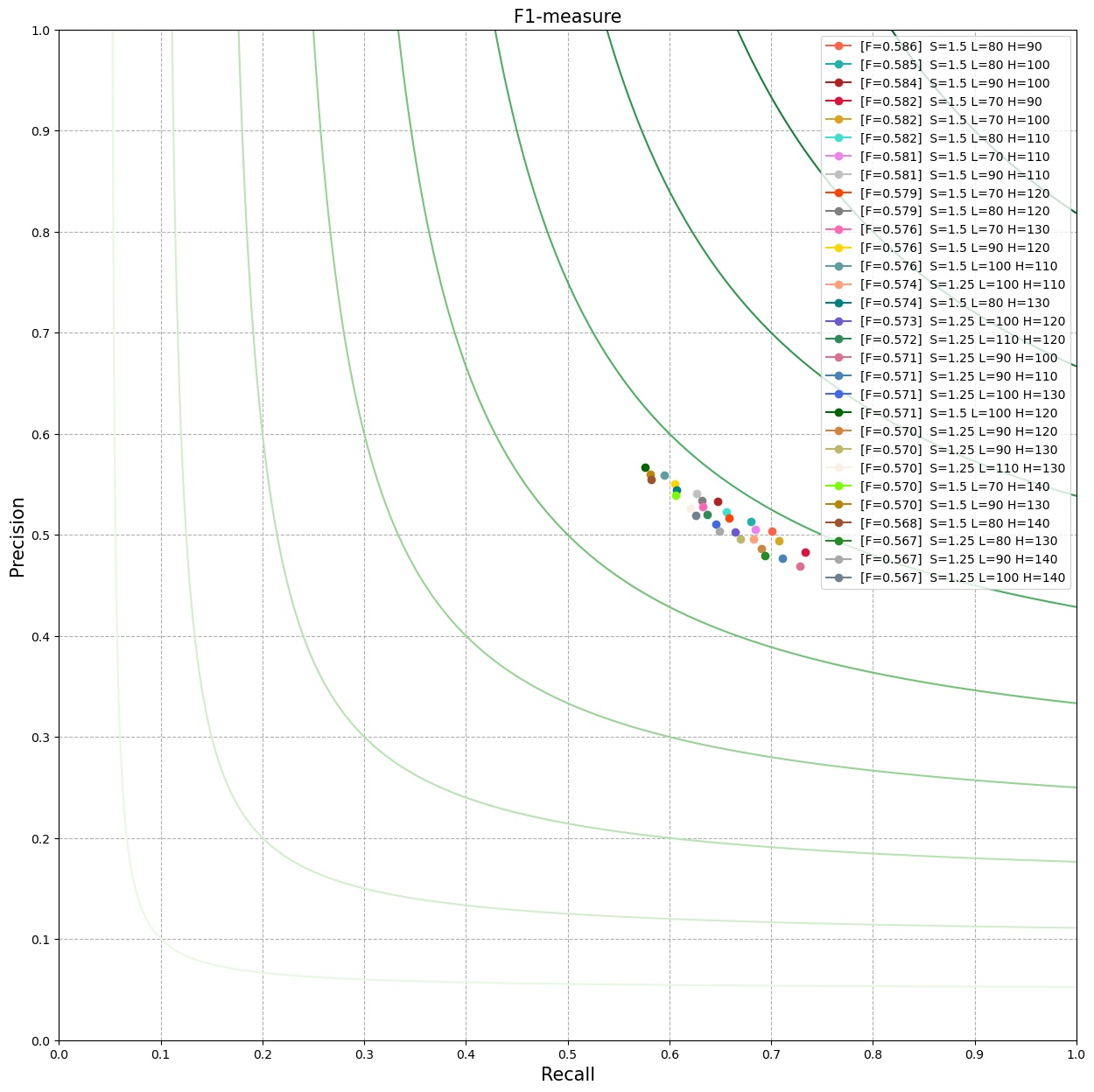} 
        \caption{Parameter tuning for Canny}
        \label{fig:canny_sigma_finder}
    \end{minipage}\hfill
    \begin{minipage}{0.45\textwidth}
        \centering
        \includegraphics[scale=0.2]{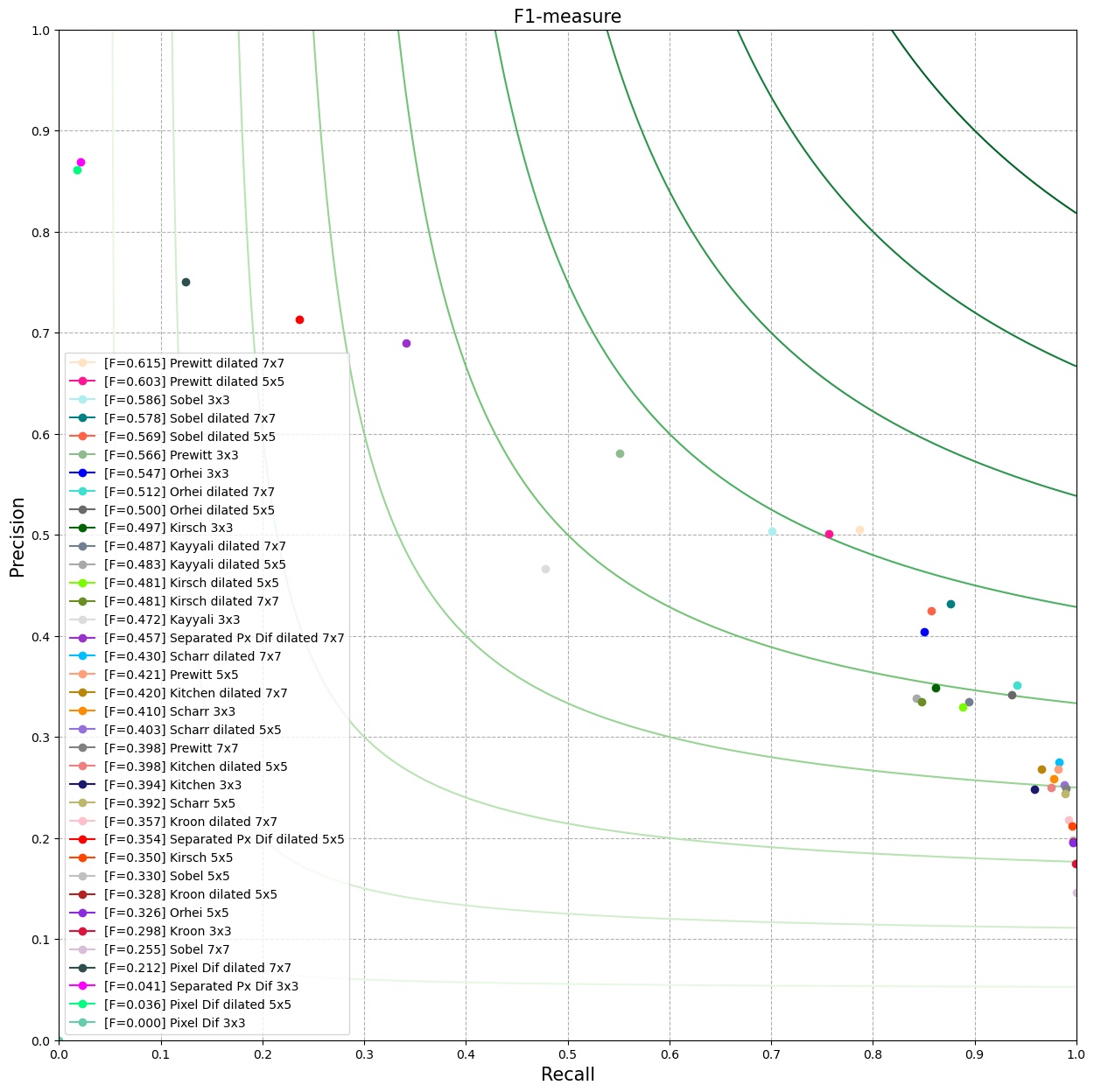} 
        \caption{Results of Canny}
        \label{fig:canny_results}
    \end{minipage}\hfill
\end{figure}

\begin{figure}[H]
\begin{minipage}{0.99\textwidth}
\centering
\setlength{\tabcolsep}{0.5pt}
\begin{tabular}{ccccccc} 
        \includegraphics[scale=0.12]{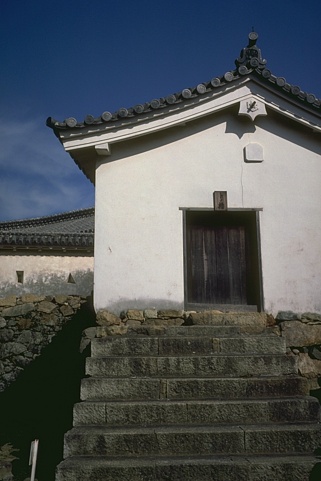} 
    &
        \includegraphics[scale=0.12]{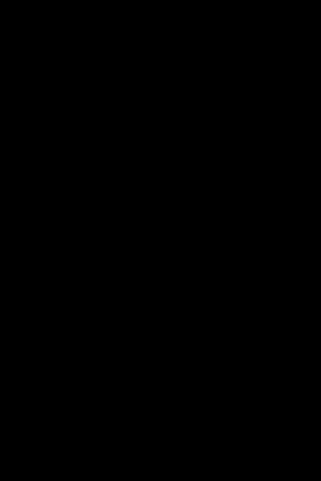} 
    & 
        \includegraphics[scale=0.12]{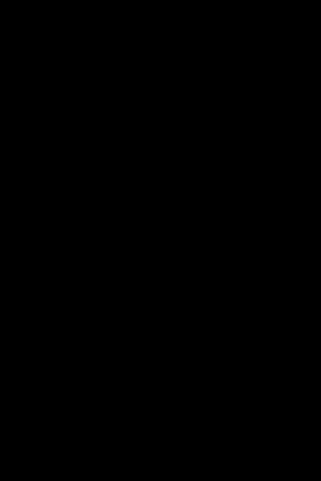}
    & 
        \includegraphics[scale=0.12]{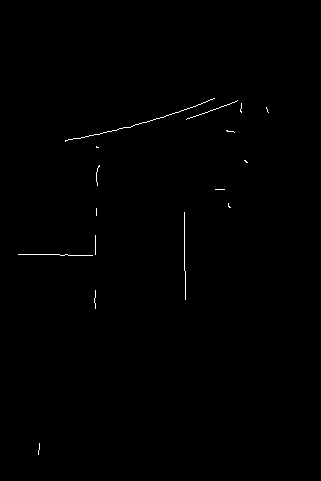}
    &
        \includegraphics[scale=0.12]{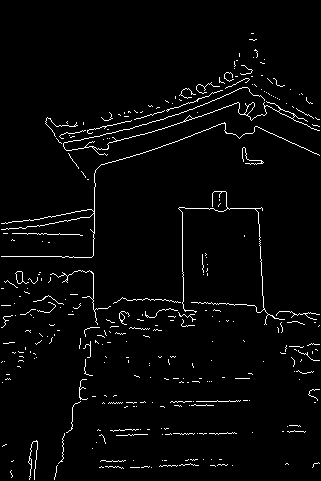} 
    & 
        \includegraphics[scale=0.12]{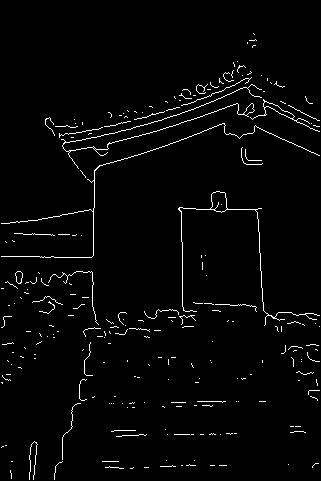}
    & 
        \includegraphics[scale=0.12]{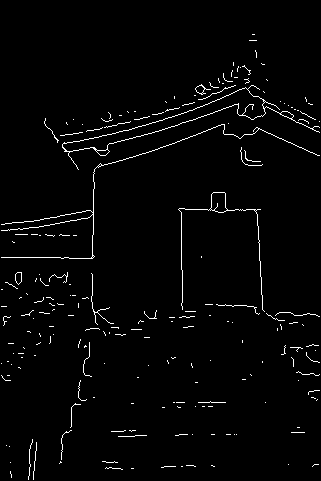}
\\
        \includegraphics[scale=0.12]{result_images/canny/334025.jpg} 
    &
        \includegraphics[scale=0.12]{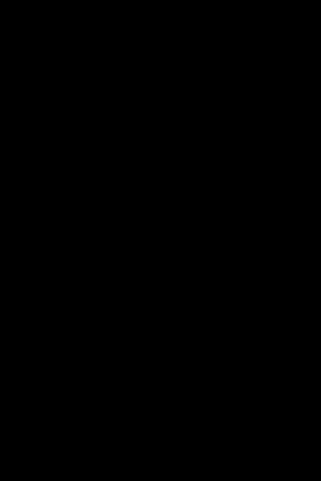} 
    & 
        \includegraphics[scale=0.12]{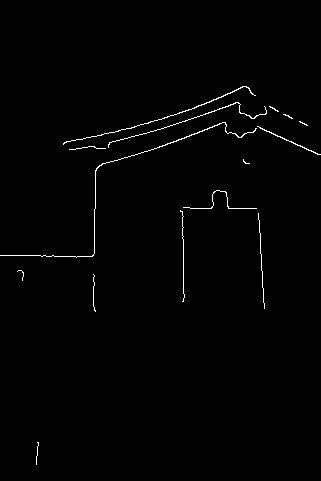}
    & 
        \includegraphics[scale=0.12]{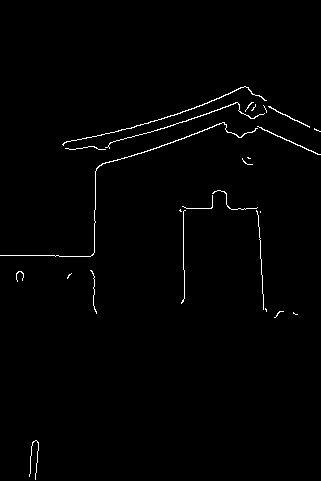}
    &
        \includegraphics[scale=0.12]{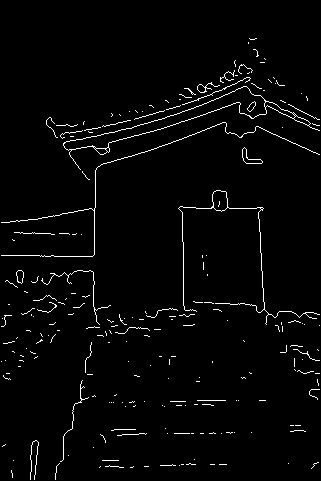} 
    & 
        \includegraphics[scale=0.12]{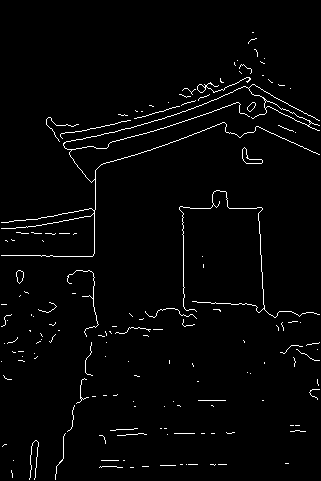}
    & 
        \includegraphics[scale=0.12]{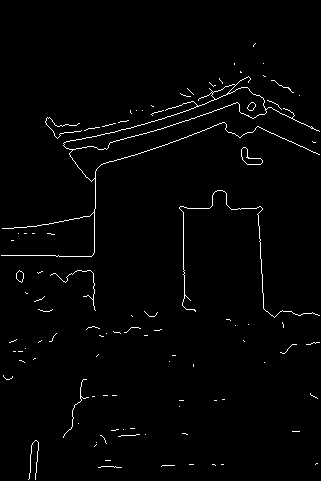}
\\
        \includegraphics[scale=0.12]{result_images/canny/334025.jpg} 
    &
        \includegraphics[scale=0.12]{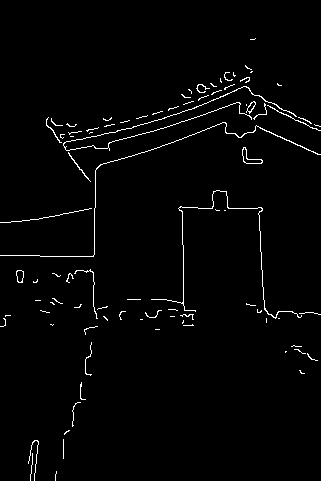} 
    & 
        \includegraphics[scale=0.12]{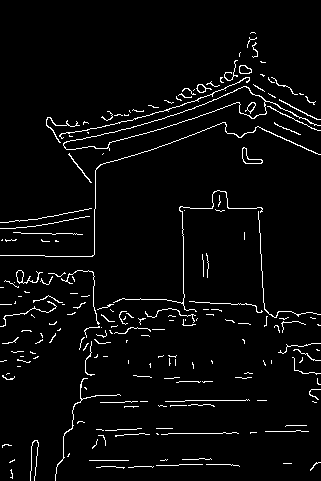}
    & 
        \includegraphics[scale=0.12]{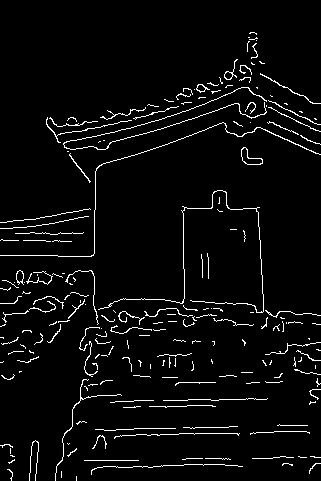}
    &
        \includegraphics[scale=0.12]{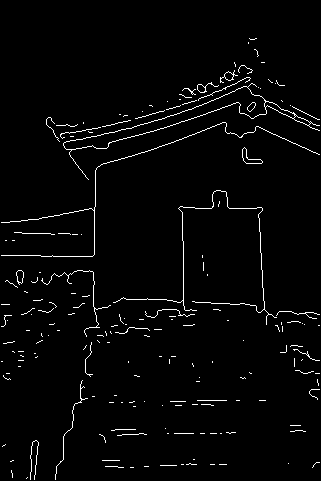} 
    & 
        \includegraphics[scale=0.12]{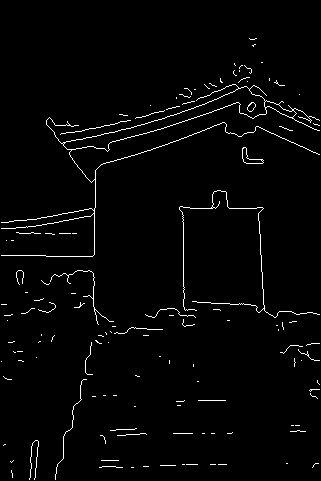}
    & 
        \includegraphics[scale=0.12]{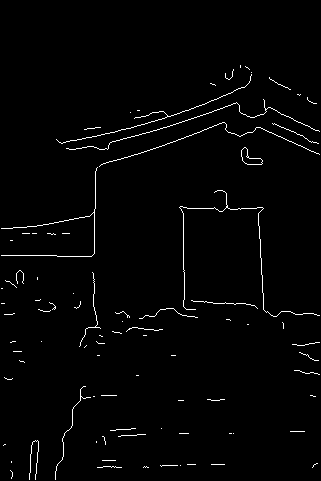}
\\
        \includegraphics[scale=0.12]{result_images/canny/334025.jpg} 
    &
      \includegraphics[scale=0.12]{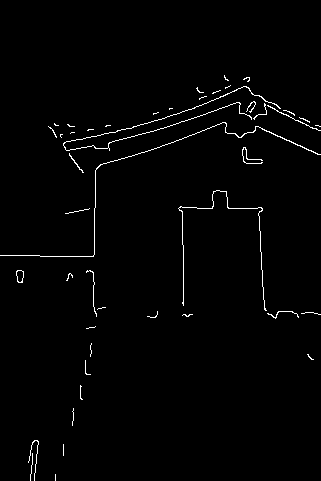} 
    & 
        \includegraphics[scale=0.12]{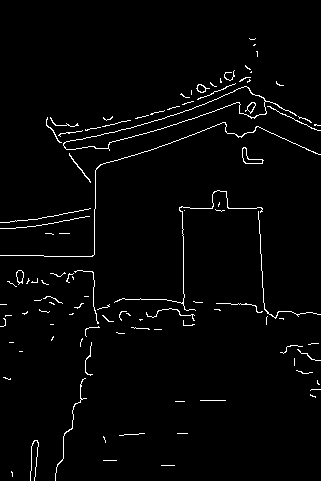}
    & 
        \includegraphics[scale=0.12]{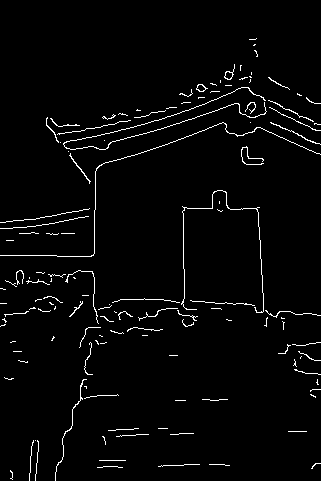}
    &
      \includegraphics[scale=0.12]{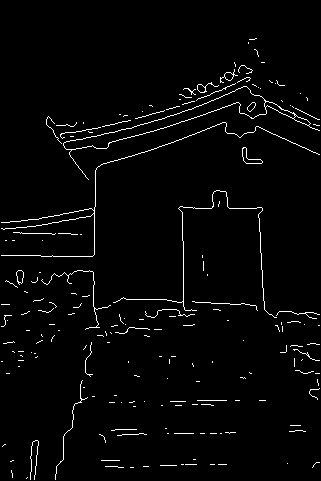} 
    & 
        \includegraphics[scale=0.12]{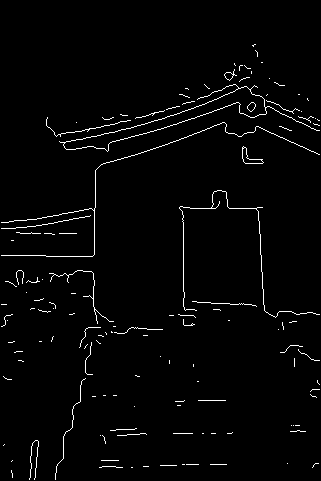}
    & 
        \includegraphics[scale=0.12]{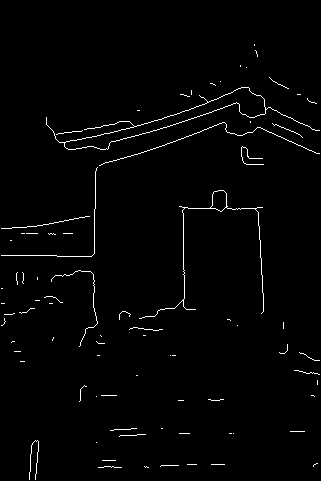}
\\
        \includegraphics[scale=0.12]{result_images/canny/334025.jpg} 
    &
        \includegraphics[scale=0.12]{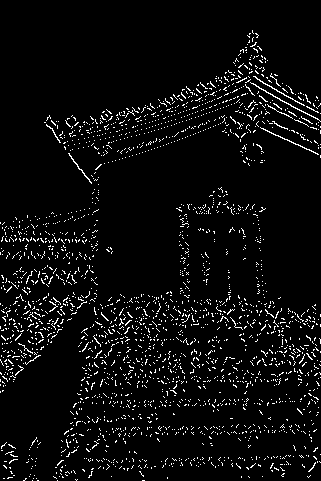} 
    & 
        \includegraphics[scale=0.12]{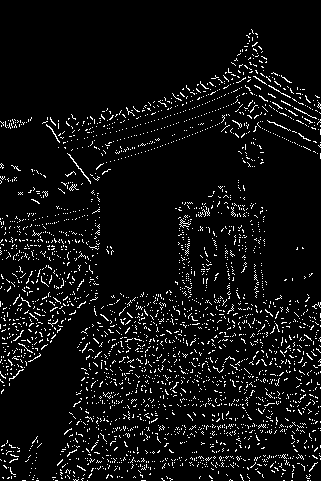}
    & 
        \includegraphics[scale=0.12]{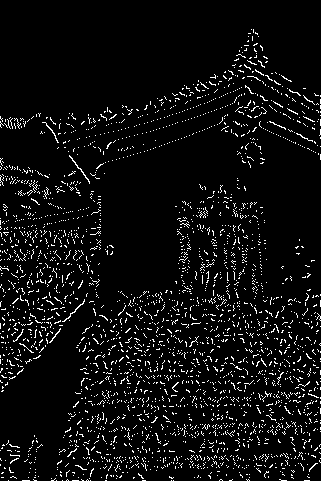}
    &
        \includegraphics[scale=0.12]{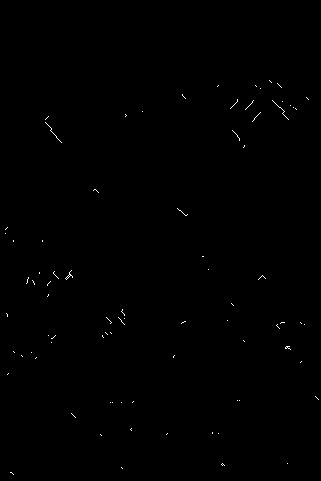} 
    & 
        \includegraphics[scale=0.12]{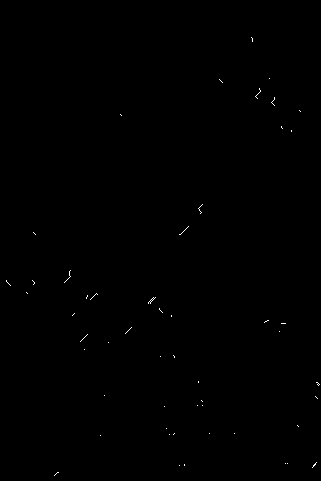}
    & 
        \includegraphics[scale=0.12]{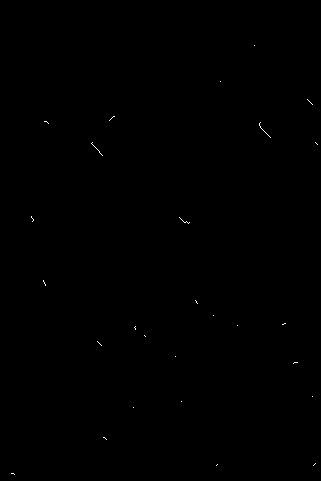}
\\
        \includegraphics[scale=0.12]{result_images/canny/334025.jpg} 
    &
        \includegraphics[scale=0.12]{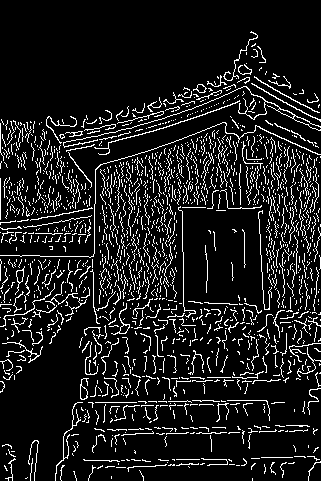} 
    & 
        \includegraphics[scale=0.12]{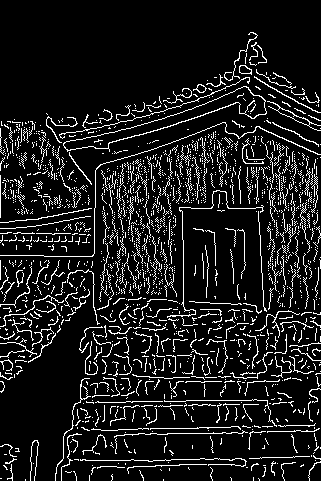}
    & 
        \includegraphics[scale=0.12]{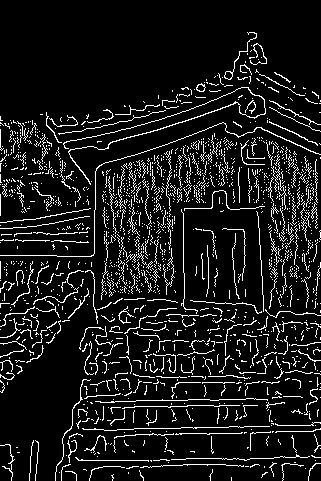}
    &
        \includegraphics[scale=0.12]{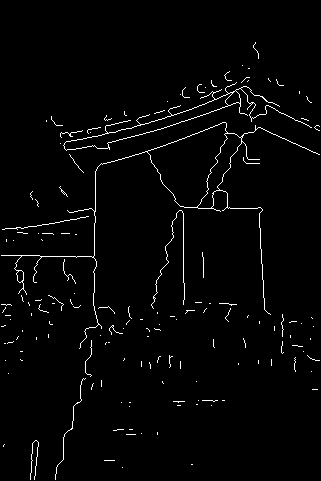} 
    & 
        \includegraphics[scale=0.12]{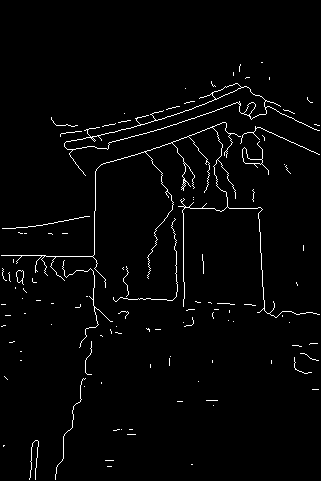}
    & 
        \includegraphics[scale=0.12]{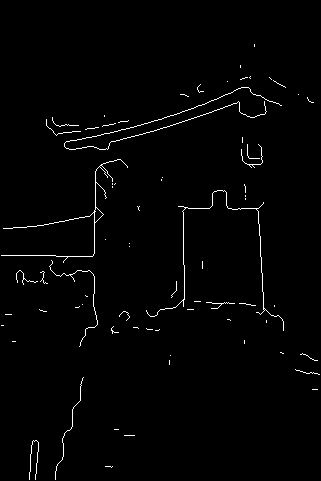}
\\
        \includegraphics[scale=0.12]{result_images/canny/334025.jpg} 
    &
        \includegraphics[scale=0.12]{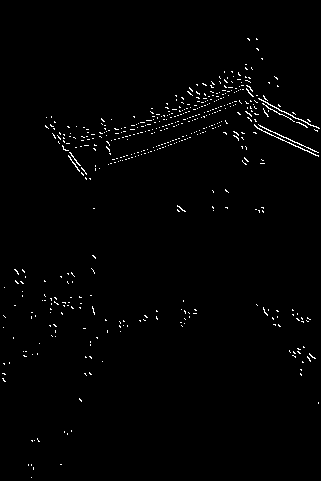} 
    & 
        \includegraphics[scale=0.12]{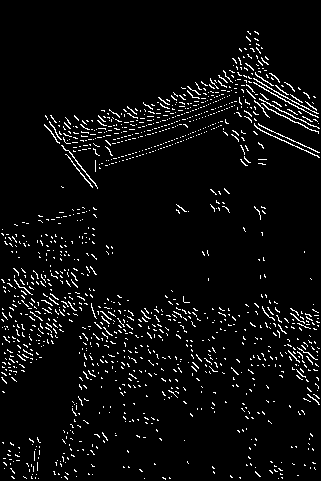}
    & 
        \includegraphics[scale=0.12]{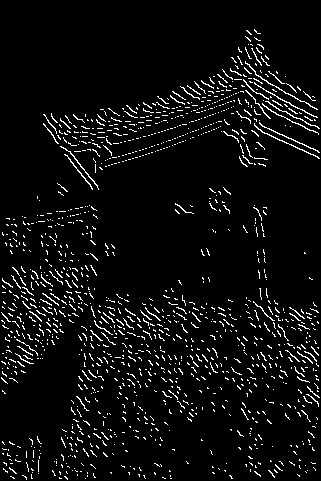}
    &
        \includegraphics[scale=0.12]{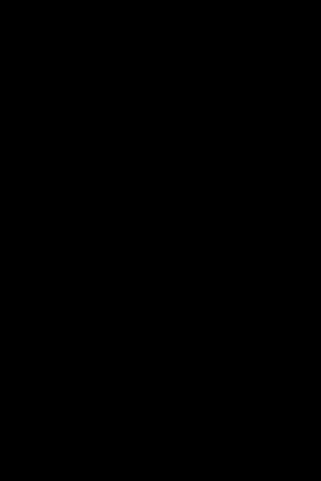} 
    & 
        \includegraphics[scale=0.12]{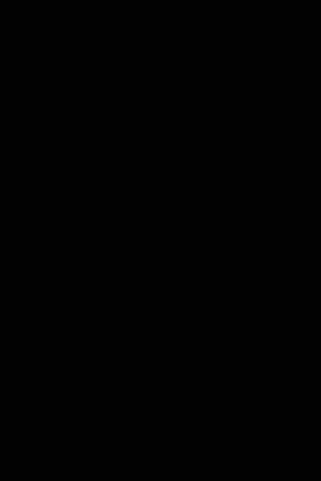}
    & 
        \includegraphics[scale=0.12]{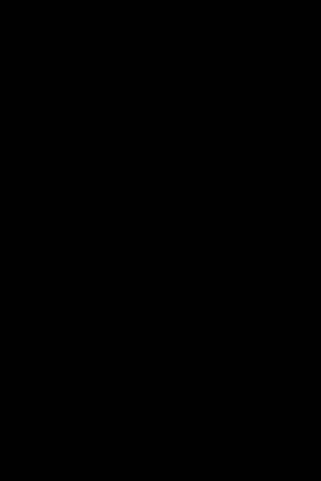}
\\
        \includegraphics[scale=0.12]{result_images/canny/334025.jpg} 
    &
        \includegraphics[scale=0.12]{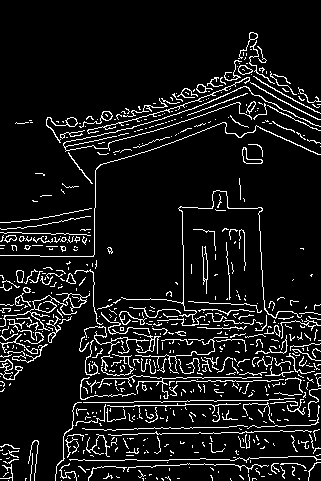} 
    & 
        \includegraphics[scale=0.12]{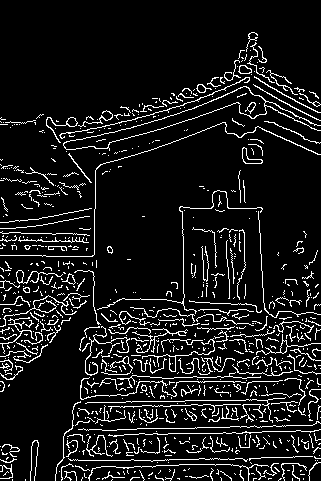}
    & 
        \includegraphics[scale=0.12]{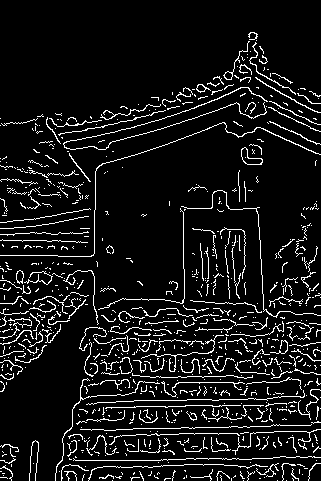}
    &
        \includegraphics[scale=0.12]{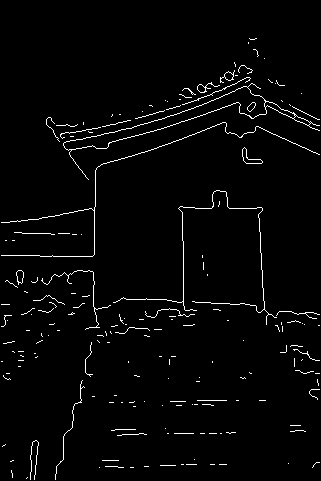} 
    & 
        \includegraphics[scale=0.12]{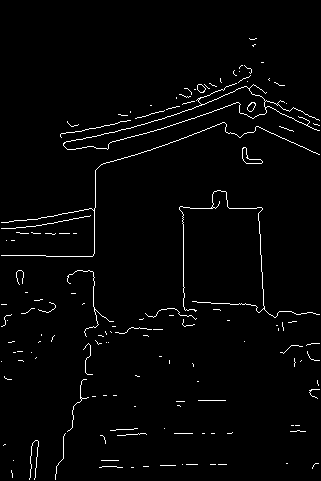}
    & 
        \includegraphics[scale=0.12]{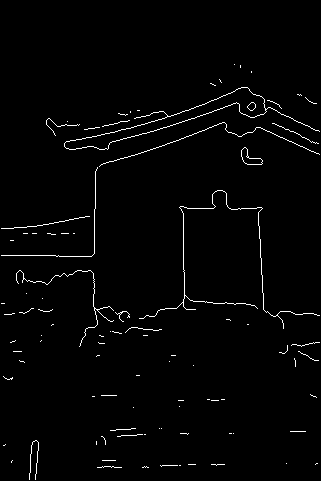}
\\
        \includegraphics[scale=0.12]{result_images/canny/334025.jpg} 
    &
        \includegraphics[scale=0.12]{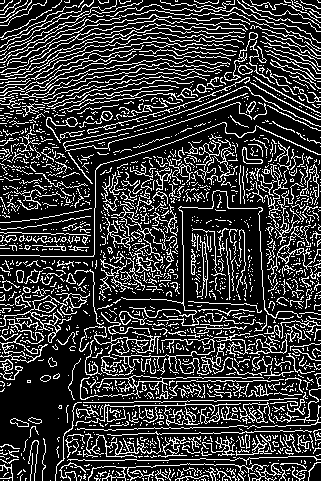} 
    & 
        \includegraphics[scale=0.12]{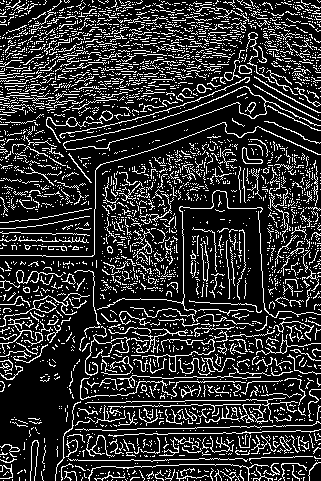}
    & 
        \includegraphics[scale=0.12]{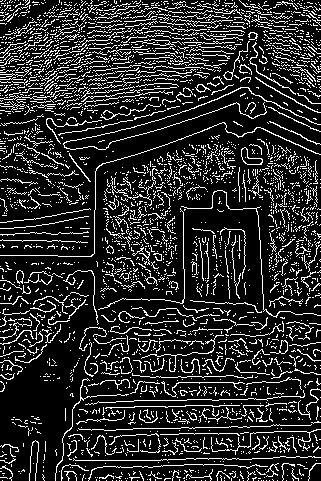}
    &
        \includegraphics[scale=0.12]{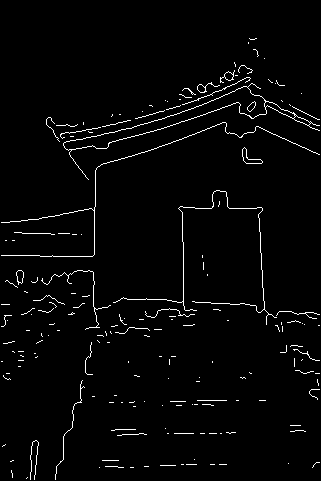} 
    & 
        \includegraphics[scale=0.12]{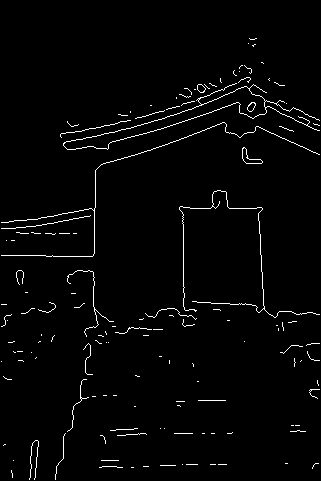}
    & 
        \includegraphics[scale=0.12]{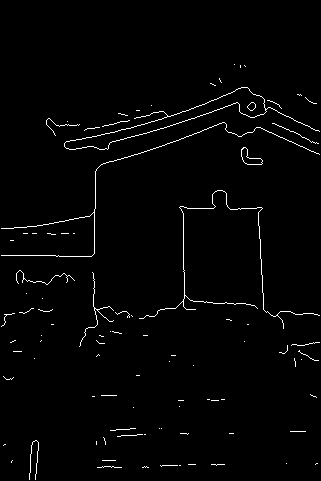}
\\
        \includegraphics[scale=0.12]{result_images/canny/334025.jpg} 
    &
        \includegraphics[scale=0.12]{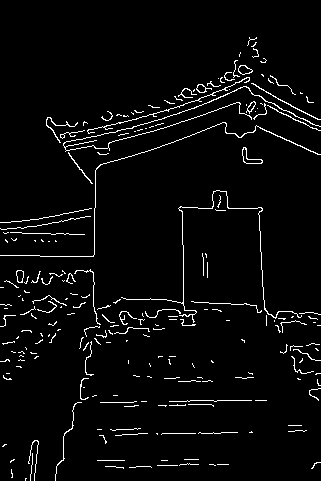} 
    & 
        \includegraphics[scale=0.12]{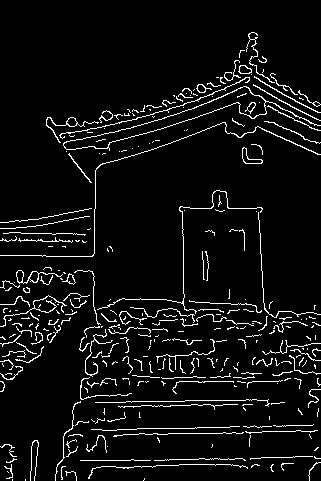}
    & 
        \includegraphics[scale=0.12]{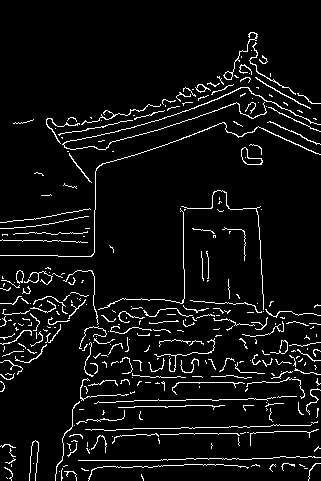}
    &
        \includegraphics[scale=0.12]{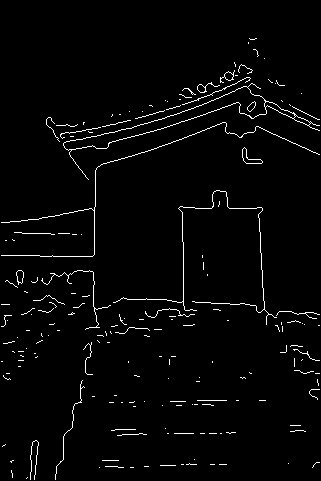} 
    & 
        \includegraphics[scale=0.12]{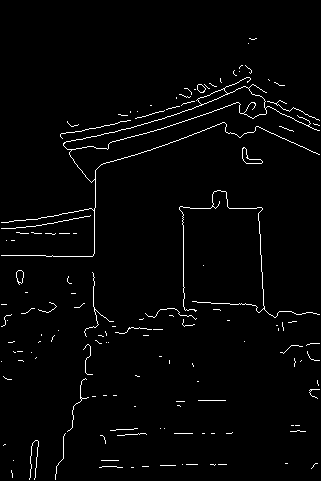}
    & 
        \includegraphics[scale=0.12]{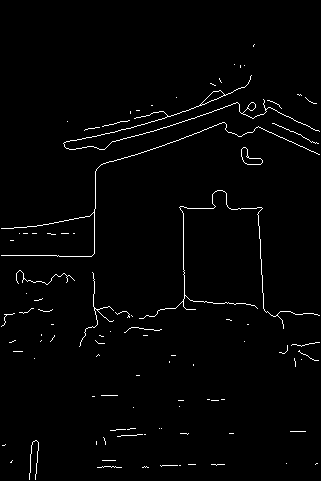}
\end{tabular}
\caption{Canny and ED results Columns: Original, Canny 3x3, Canny Dilated 5x5, Canny Dilated 7x7, ED 3x3, ED Dilated 5x5, ED Dilated 7x7; Rows: Pixel Differences, Separated Pixel Difference, Sobel, Prewitt, Kirsch, Kitchen, Kayyali, Scharr, Kroon, Orhei}
\label{fig:canny_img_results}
\label{fig:ed_img_results}
\end{minipage}
\end{figure}

The visual comparison results of the operators can be seen in Figures \ref{fig:canny_img_results}. We can observe from our simulation that the dilation of the kernels doesn't produce a degradation of the edge map. The dilation actually brings an improvement in most cases, which was already presented in \cite{Dilatetion2020} and \cite{DilateionVsExpansion2020}.

For parameter tuning we vary the following configurations: Gaussian sigma value from $0.2$ to $3.0$ with a step of $0.25$, low threshold from $70$ to $150$ with a step of $10$, high threshold from $90$ to $200$ with a step of $10$. As we can see in  Figure \ref{fig:canny_sigma_finder} the best results we obtain when Gaussian sigma is $1.5$, low threshold is $80$ and high threshold $90$. For the images we use the following notations: Gaussian sigma is $S$, low threshold is $L$, high threshold is $H$.

Table \ref{table:canny_filter_results} displays the results of the Canny Edge algorithm for all the operators which we analyzed. In most of the cases, the dilation of the initial operators led to higher $F1$ scores. In the case of the Prewitt, Kayyali or Scharr  filters, we can notice that the best F1 scores are obtained by the 7x7 dilated filters. Also, the dilated 5x5 and 7x7 filters obtain in both cases a significantly higher F1 score than the extended 5x5 and 7x7 filters. Similarly, in the case of the other operators, the dilation of the original filters obtained higher F1 scores. 

All these results are also highlighted in Figure \ref{fig:canny_results}, where we can notice that our dilation of the filters obtained more edges than the original filters.

When using Pixel Difference and Separated Pixel Difference kernels in Canny algorithm the benchmark results have not been improve by dilation. The edge pixels obtained by the magnitude calculation step in Canny are not strong enough to pass the double hysteresis thresholding and therefore the resulted edge map is blank (as we can see in Table \ref{table:canny_filter_results} last columns). When 7x7 dilated version is used, a limited number of edge pixels appears but not enough to be relevant in F1-score, even if the precision is good enough.

\subsection{Shen-Castan algorithm}

The results and evaluation of Shen-Castan Operator, described in section \ref{Sec:preliminari_shen_castan}, are presented in this section.  We searched again for the best combination of threshold of Laplace, smoothing factor, window size, thinning factor and ratio factor of the algorithm. We chose a threshold value of $40$ for the Laplace edge detector and vary the rest of the values as following: smoothing factor of ISEF filter($S$) from $0.5$ to $0.9$; adaptive zero crossing window size($W$) of $5$, $7$, $9$; threshold for zero crossing($R$) from $0.5$ to $0.9$; thinning factor($TH$) of $0$, $0.5$, $0.9$.  

\begin{figure}[H]
    \centering
    \begin{minipage}{0.45\textwidth}
        \centering
        \includegraphics[scale=0.2]{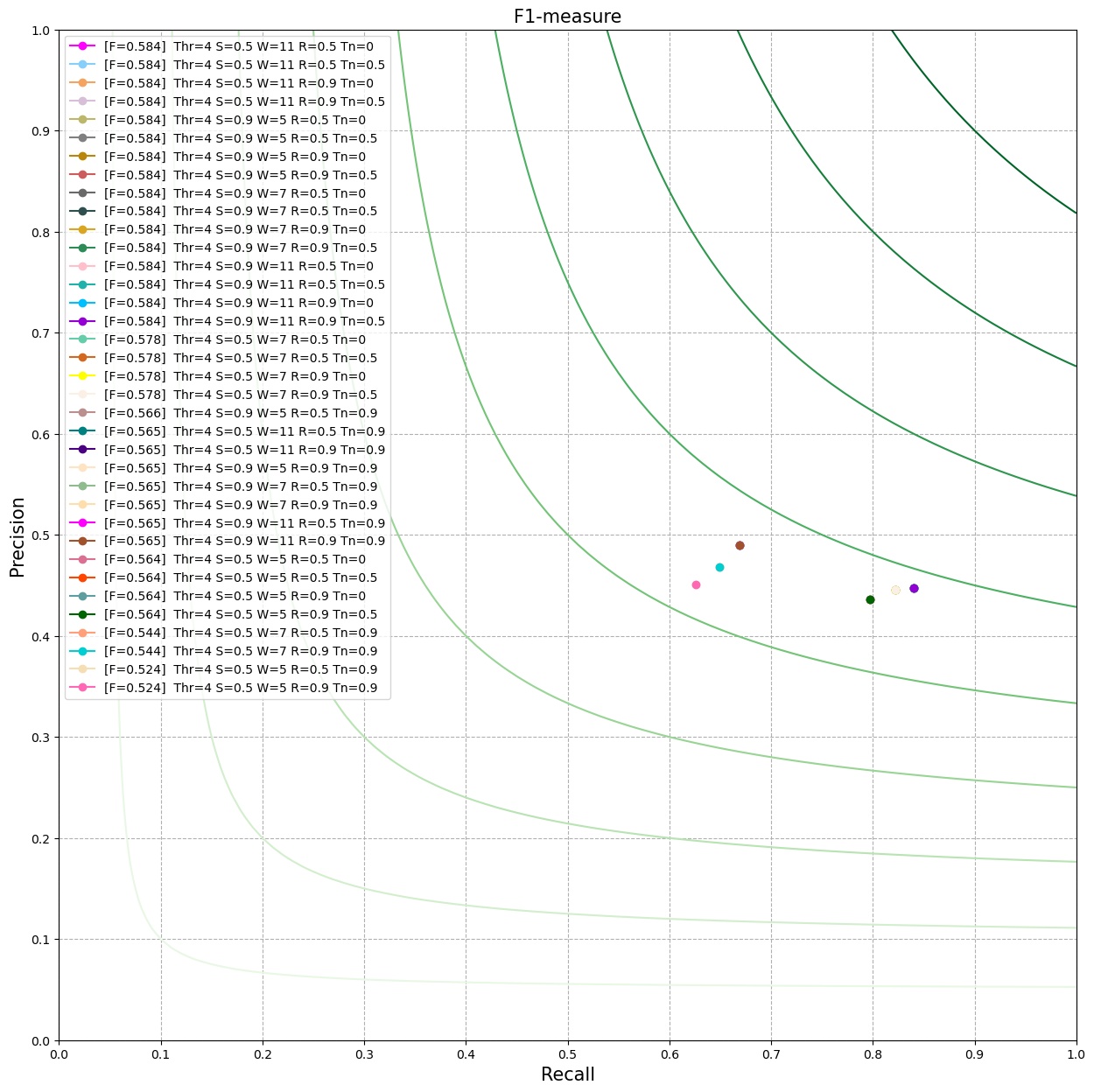} 
        \caption{Tuning for Shen-Castan edge operator}
        \label{fig:shen_tunning}
    \end{minipage}\hfill
    \begin{minipage}{0.45\textwidth}
        \centering
        \includegraphics[scale=0.2]{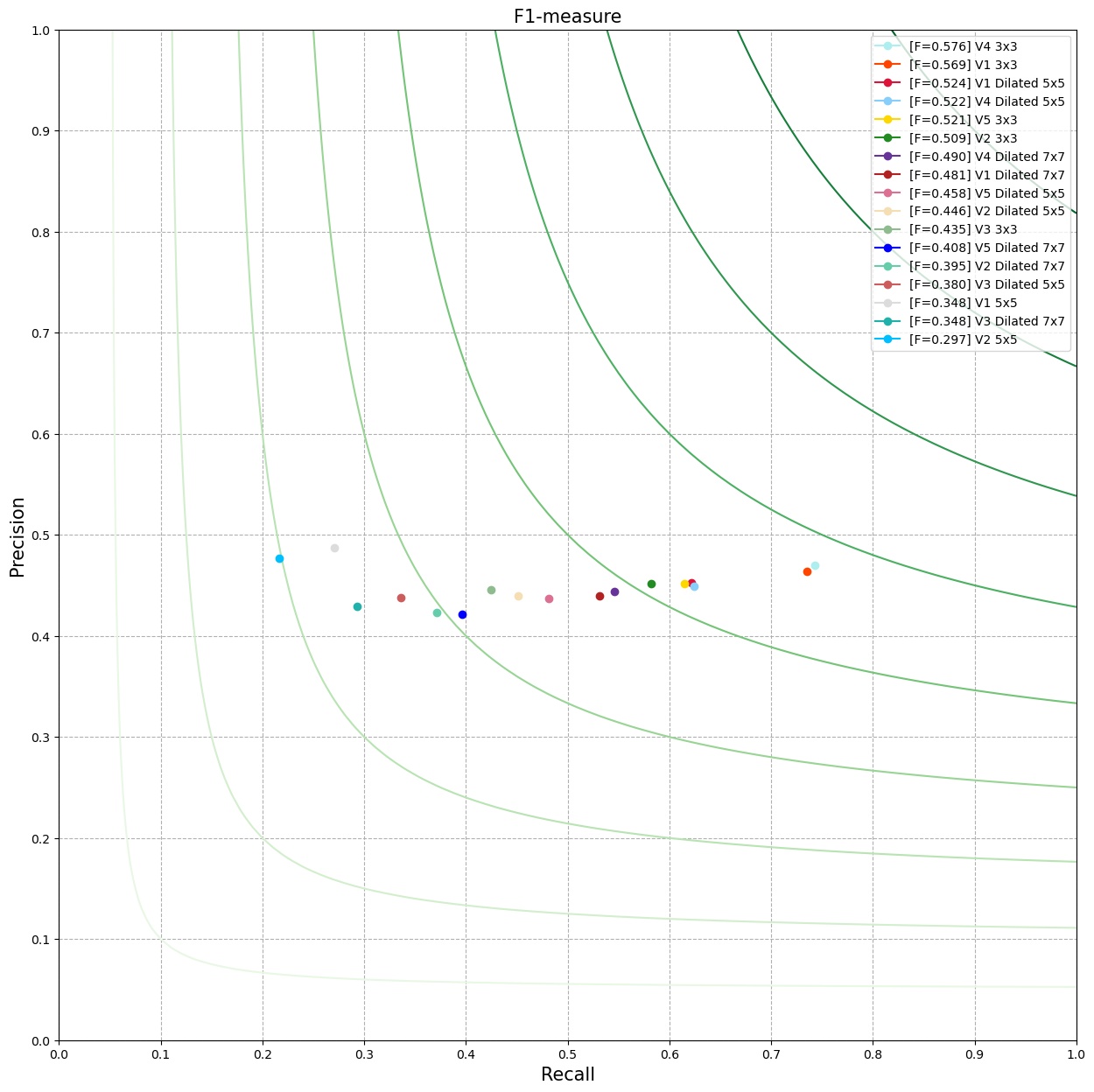} 
        \caption{Results for Shen-Castan edge operator}
        \label{fig:shen_results}
    \end{minipage}\hfill
\end{figure}

\begin{table}[H]
\centering
\setlength{\tabcolsep}{2pt}
\scalebox{0.68}
{
\begin{tabular}{|l|c|cccc|cccc|cccc|cccc|}
\hline
\multicolumn{2}{|c|}{\bfseries Operator} &\multicolumn{4}{c|}{\bfseries Marr-Hildreth}&\multicolumn{4}{c|}{\bfseries Shen-Castan}\\
\hline
	&	&3x3	&5x5	&Dilated 5x5	&Dilated 7x7	&3x3	&5x5	&Dilated 5x5	&Dilated 7x7	\\
\hline
	&R	&0.716	&0.141	&0.603	&0.554	&0.735	&0.271	&0.622	&0.531\\
V1	&P	&0.414	&0.634	&0.484	&0.516	&0.464	&0.487	&0.453	&0.440\\
	&F1	&0.525	&0.231	&\textbf{0.537}	&0.534	&\textbf{0.569}	&0.348	&0.524	&0.481\\
\hline
	&R	&0.664	&0.000	&0.597	&0.546	&0.582	&0.216	&0.451	&0.371\\
V2	&P	&0.447	&0.000	&0.496	&0.534	&0.452	&0.477	&0.440	&0.423\\
	&F1	&0.535	&0.000	&\textbf{0.542}	&0.540	&\textbf{0.509}	&0.297	&0.446	&0.395\\
\hline
	&R	&0.691	&-	&0.601	&0.552	&0.425	&-	&0.336	&0.293\\
V3	&P	&0.430	&-	&0.490	&0.525	&0.446	&-	&0.438	&0.429\\
	&F1	&0.530	&-	&\textbf{0.540}	&0.538	&\textbf{0.435}	&-	&0.380	&0.348\\
\hline
	&R	&0.999	&-	&0.649	&0.505	&0.743	&-	&0.624	&0.546\\
V4	&P	&0.143	&-	&0.315	&0.387	&0.470	&-	&0.449	&0.444\\
	&F1	&0.250	&-	&0.424	&\textbf{0.438}	&\textbf{0.576}	&-	&0.522	&0.490\\
\hline
	&R	&0.636	&-	&0.586	&0.517	&0.615	&-	&0.481	&0.396\\
V5	&P	&0.465	&-	&0.504	&0.550	&0.452	&-	&0.437	&0.421\\
	&F1	&0.537	&-	&\textbf{0.542}	&0.533	&\textbf{0.521}	&-	&0.458	&0.408\\
\hline

\end{tabular}}
\vspace{1.5pt}
\caption{Results of Marr-Hildreth operator and Shen-Castan operator}
\label{table:marr_results}
\label{table:shen_results}
\end{table}%

We can observe in Figure \ref{fig:shen_tunning} the best results we obtain when ISEF smoothing factor is $0.9$, ZC window is $7$, ZC threshold is $0.9$ and thinning factor of $0.5$. It looks that the figure doesn't show the correct value but in reality the values are overlapped. It can be seen in the details of the legend values from the figure.

In Figure \ref{fig:shen_results} we can see the results of simulating the algorithm using different kernels. We can observe that dilating the kernel size of the Laplace operator does not obtain better results in any case of the cases. If we look over the detailed results, see Table \ref{table:shen_results}, the \textit{precision} is not actually that affected by the dilation but \textit{recal} is strongly affected.

If we look over the visual results, see Figure \ref{fig:shen_img_results}, we can see that the results in some cases looks better but we can clearly see the increase in quantity of noise pixels appearing.

\begin{figure}[H]
    \centering
    \begin{minipage}{0.45\textwidth}
        \centering
        \includegraphics[scale=0.2]{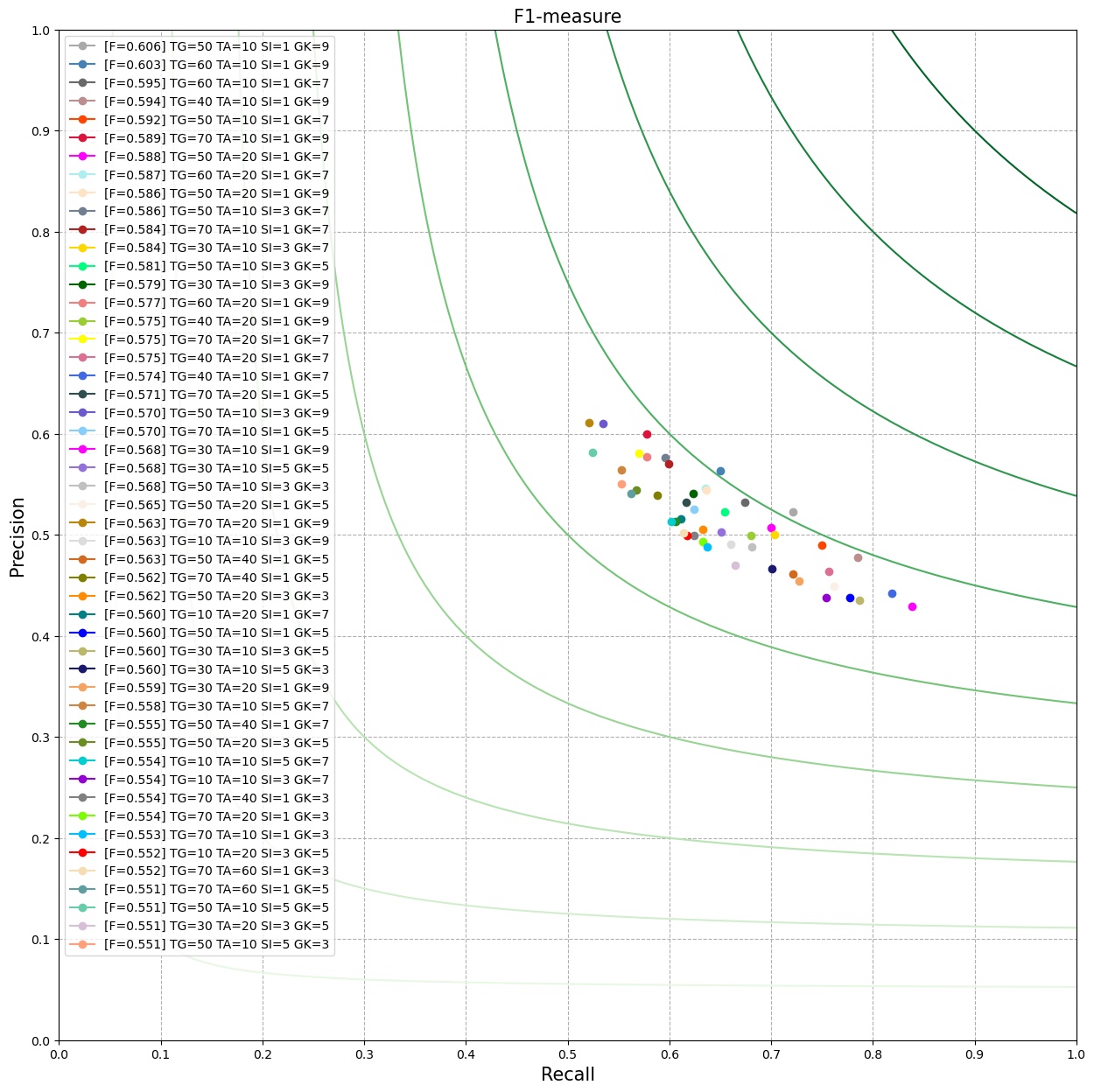} 
        \caption{Parameter tuning for ED edge operator}
        \label{fig:ed_tunning}
    \end{minipage}\hfill
    \begin{minipage}{0.45\textwidth}
        \centering
        \includegraphics[scale=0.2]{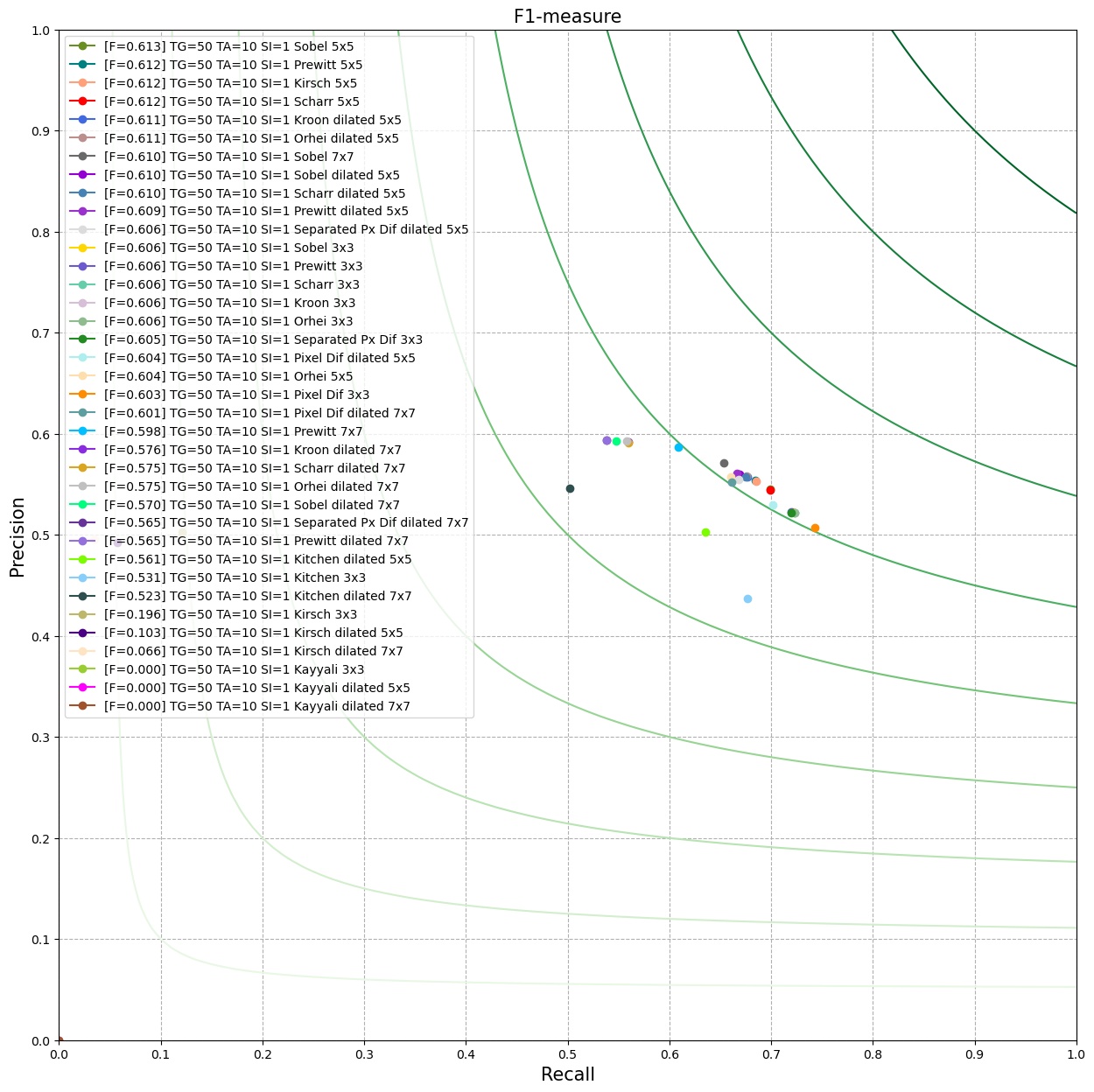} 
        \caption{Results for ED edge operator\\~}
        \label{fig:ed_results}
    \end{minipage}\hfill
\end{figure}

\begin{table}[H]
\begin{minipage}{.99\textwidth}
   \centering
        \setlength{\tabcolsep}{2pt}
        \scalebox{0.65}
        {
            \begin{tabular}{|l|c|ccccc|ccccc|}
            \hline
            \multicolumn{2}{|c|}{\bfseries Operator} &\multicolumn{5}{c|}{\bfseries ED}\\
            \hline
            	&	&3x3	&5x5	&Dilated 5x5	&7x7	&Dilated 7x7	\\
            \hline
            	                    &R	&0.743	&-	&0.702	&-	&0.661	            \\
            Pixel Diff	            &P	&0.507	&-	&0.530	&-	&0.552	            \\
            	                    &F1	&0.603	&-	&\textbf{0.604}	&-	&0.602	            \\
            \hline
            	                    &R	&0.720	&-	&0.668	&-	&0.538	            \\
            Separated Pixle Diff	&P	&0.522	&-	&0.555	&-	&0.594	            \\
            	                    &F1	&0.605	&-	&\textbf{0.606}	&-	&0.565	            \\
            \hline
            	                    &R	&0.721	&0.699	&0.669	&0.653	&0.548	    \\
            Sobel	                &P	&0.523	&0.545	&0.560	&0.572	&0.593	    \\
            	                    &F1	&0.606	&\textbf{0.613}	&0.610	&0.610	&0.570	    \\
            \hline
            	                    &R	&0.720	&0.684	&0.666	&0.609	&0.539	    \\
            Prewitt	                &P	&0.523	&0.554	&0.561	&0.587	&0.594	    \\
            	                    &F1	&0.606	&\textbf{0.613}	&0.609	&0.598	&0.565	    \\
            \hline
            	                    &R	&0.121	&0.686	&0.057	&-	&0.035	        \\
            Kirsch	                &P	&0.503	&0.553	&0.492	&-	&0.496	        \\
            	                    &F1	&0.196	&\textbf{0.612}	&0.103	&-	&0.066	        \\
            \hline
            	                    &R	&0.677	&-	&0.635	&-	&0.502	            \\
            Kitchen	                &P	&0.437	&-	&0.503	&-	&0.546	            \\
            	                    &F1	&0.531	&-	&\textbf{0.561}	&-	&0.523	            \\
            \hline
            	                    &R	&0.000	&-	&0.000	&-	&0.000	            \\
            Kayyali	                &P	&0.000	&-	&0.000	&-	&0.000	            \\
            	                    &F1	&0.000	&-	&0.000	&-	&0.000	            \\
            \hline
            	                    &R	&0.723	&0.699	&0.675	&-	&0.560	        \\
            Scharr	                &P	&0.522	&0.544	&0.557	&-	&0.590	        \\
            	                    &F1	&0.606	&\textbf{0.612}	&0.610	&-	&0.575	        \\
            \hline
            	                    &R	&0.723	&-	&0.677	&-	&0.561	            \\
            Kroon	                &P	&0.522	&-	&0.556	&-	&0.592	            \\
            	                    &F1	&0.606	&-	&\textbf{0.611}	&-	&0.576	            \\
            \hline
            	                    &R	&0.723	&0.660	&0.675	&-	&0.558	       \\
            Orhei	                &P	&0.522	&0.556	&0.558	&-	&0.593	        \\
            	                    &F1	&0.606	&0.604	&\textbf{0.610}	&-	&0.575	        \\
            \hline

            \end{tabular}}
        \vspace{1.5pt}
    \caption{F1-measure results for Edge Drawing algorithm}
    \label{table:ed_results}
\end{minipage}
\end{table}

\subsection{Edge Drawing algorithm}

For our simulation results and evaluation of ED algorithm, described in section \ref{Sec:preliminari_ed}, we first searched for the best combination of Gaussian smoothing kernel size, gradient threshold, anchor threshold and scan interval interval. We choose to vary the parameters as following: Gaussian kernel ($GK$) between $3$ to $9$ with a step of $2$; gradient threshold ($TG$) between $10$ and $150$ with a step of $10$; anchor threshold ($TA$) between $10$ and $60$ with a step of $10$; scan interval in range of $1$, $3$, $5$. Looking over the $F1$ results from Figure \ref{fig:ed_tunning} we can observe that the best results are obtained using $GK$ = $9$, $TG$ = $50$, $TA$ = $10$ and scan interval is $1$.

In Figure \ref{fig:ed_img_results} we present the visual results of using the parameters found and the edge operators from Section \ref{Sec:preliminaries} and in Figure \ref{fig:ed_results} we can see $F1$ results equivalent. We can conclude that, expanding and dilating techniques offer better results than using the classical kernels. 

In Table \ref{table:ed_results} we can see the edge drawing $R$, $P$ and $F1$ results. We can observe some interesting facts like that in the case of Kayyali\cite{Kayyali2000} ED algorithm does not bring any improvements. Another observations is that in general a bigger kernel is equivalent to better results. In this case, dilating the kernels with a factor of $1$ results in better $F1$ values, than by dilating with a factor of $2$.

\section{Conclusions and future work}
\label{sec:conclusions}

In this paper we extend our work (see the previous papers \cite{Dilatetion2020, DilateionVsExpansion2020}) regarding dilation of a classical convolution edge detection filters. In Section \ref{Sec:preliminaries} we present the theoretical background where we show the algorithms upon we conducted the simulation from Section \ref{Sec:simulation_results}. The experimental results confirm our intuition that dilation of filters have positive impact for edge detection.

Dilating the second order discrete approximation filter did not bring considerable improvements to the edge map resulted. As we see in Figure \ref{table:marr_results} when using the Marr-Hildreth algorithm we can spot small improvments when using the $5x5$ filter. But when applying the Shen-Castan algorithm , see Table \ref{table:shen_results}, we do not see improvements from dilation.

Statistical and visual we could observe that by dilating the filters we find more edge pixels than by the classical operators. By dilation we obtained a better precision and F1-score which can be observed in the results we presented. By the simple structure of the dilated filters, they are also a good choice when the runtime matters. The other classical filter extensions from \cite{GuptaSobelExtension2013, ScharrExtendedChen2017, KekreSobelExtend2010, BandaiSobelExtension2003, LevkineSobelPrewitScharrExtend2012} require a larger number of operations in order to return the resulting edge pixels, whereas the custom dilated filters have always the same number of operations for any extension. It seems that the gaps imply a speed up.

On the other hand, the dilated filters might have better results on bigger images because they contain more information and therefore, the dilated filters can use pixels from a higher distance for computing the gradient and thus obtaining more edge pixels.

In our research we focused on dilating $3x3$ filters but in our future work we can concern ourselves in dilating bigger kernels as starting point. This can be analyzed from the definition of the dilation. Another aspect worth exploring in the future would be to run experiments in automated threshold version of the algorithms. We concerned our experiments on the classical version of edge detection algorithms but looking to the combined effects of dilating and automated threshold could enhance the resulted edge-map.

In our approach, for the comparison purposes, the focus was to fine-tune the classical edge detection algorithms in order to obtain the optimal threshold and sigma values. We are certain that if we would have fine-tuned the algorithms for the dilated kernel, the results would have been better. 

In general we can state that using the dilated kernels brings benefits regarding the edge-map resulting from edge detection classical algorithm and run-time needed. We obtain similar or better results, taking in consideration a bigger neighborhood.

\iffalse

\section{Acknowledgements}
This work was partially supported by a grant of Ministry of Research and Innovation, CNCS - UEFISCDI, project number
PN-III-P4-ID-PCE-2016-0842, within PNCDI III.
\fi

%\bibliographystyle{apalike}
%\bibliography{References}

%\appendix
%\newpage
\label{appendix:A}
\section*{Appendix A}

\begin{figure}[!h]
%\begin{minipage}{0.45\textwidth}
\centering
\tiny{
\begin{tabular}{ccccc}
        $\begin{bmatrix}
            0&0&0 \\
            0&-1&1\\
            0&0&0 \\
        \end{bmatrix}$
    &
        $\begin{bmatrix}
            0&0&0 \\
            -1&0&1\\
            0&0&0 \\
        \end{bmatrix}$
    &
        $\begin{bmatrix}
        -1 & 0 & 1 \\
        -2 & 0 & 2 \\
        -1 & 0 & 1 \\
        \end{bmatrix}$
    &
        $\begin{bmatrix}
            -1 & 0 & 1 \\
            -1 & 0 & 1 \\
            -1 & 0 & 1 \\
        \end{bmatrix}$
    &
        $\begin{bmatrix}
            -3 & -3 & 5 \\
            -3 &  0 & 5 \\
            -3 & -3 & 5 \\
        \end{bmatrix}$
    \\~
    \\
        Pixel Difference 
    & 
        Separated Pixel 
    &
        Sobel 
    &         
        Prewitt
    &
        Kirsch
    \\~
    \\
        $\begin{bmatrix}
             -2 & 0 & 2 \\
             -3 & 0 & 3 \\
             -2 & 0 & 2 \\
        \end{bmatrix}$
    &
        $\begin{bmatrix}
             -6 & 0 &  6 \\
              0 & 0 &  0 \\
              6 & 0 & -6 \\
        \end{bmatrix}$
    &
        $\begin{bmatrix}
             -3 & 0 &  3 \\
            -10 & 0 & 10 \\
             -3 & 0 &  3 \\
        \end{bmatrix}$
    &
         $\begin{bmatrix}
            -17 & 0 & 17 \\
            -61 & 0 & 61 \\
            -17 & 0 & 17 \\
        \end{bmatrix}$
    &
        $\begin{bmatrix}
            -1 & 0 & 1 \\
            -4 & 0 & 4 \\
            -1 & 0 & 1 \\
        \end{bmatrix}$
   \\~
   \\

        Kitchen-Malin
    &
        Kayyali
    &
        Scharr
    &
        Kroon
    &
        Orhei
    \\ 
\end{tabular}
}

%\end{minipage}
\caption{3$x$3 kernels masks \label{fig:k3_kernel_masks}}
\end{figure}

\begin{figure}[!h]
%\begin{minipage}[c]{0.45\textwidth}
\centering
\tiny{
\begin{tabular}{ccc}
    $\begin{bmatrix}
        -1&1&1 \\
        -1&-2&1\\
        -1&1&1 \\
    \end{bmatrix}$
    &
    $\begin{bmatrix}
        -1&0&1 \\
        -2&0&2\\
        -1&0&1 \\
    \end{bmatrix}$
    &
    $\begin{bmatrix}
        -3 & -3 & 5 \\
        -3 & 0 & 5 \\
        -3 & -3 & 5 \\
    \end{bmatrix}$
    \\
    ~
    \\
        Prewitt Compass
    &
        Robinson Compass
    &
        Kirsch Compass
\end{tabular}
}
\caption{3$x$3 compass kernels masks}
\label{fig:compass_kernel_masks}
%\end{minipage}
\end{figure}

\begin{figure}[!h]
%\begin{minipage}[c]{0.45\textwidth}
\centering
\tiny{
\begin{tabular}{ccc}
        $\begin{bmatrix}
             -5 &  -4 & 0 &  4 & 5  \\
             -8 & -10 & 0 & 10 & 8  \\
            -10 & -20 & 0 & 20 & 10 \\
             -8 & -10 & 0 & 10 & 8  \\
             -5 &  -4 & 0 &  4 & 5  \\
        \end{bmatrix}$
    &
        $\begin{bmatrix}
            -2 & -1 & 0 & 1 & 2  \\
            -2 & -1 & 0 & 1 & 2  \\
            -2 & -1 & 0 & 1 & 2  \\
            -2 & -1 & 0 & 1 & 2  \\
            -2 & -1 & 0 & 1 & 2  \\
        \end{bmatrix}$
    &
        $\begin{bmatrix}
        -7 & -7 & -7 & 9 & 9  \\
        -7 & -3 & -3 & 5 & 9  \\
        -7 & -3 &  0 & 5 & 9  \\
        -7 & -3 & -3 & 5 & 9  \\
        -7 & -7 & -7 & 9 & 9  \\
    \end{bmatrix}$
    \\
        ~
    \\
        Sobel Extended
    &
        Prewitt Extended
    &
        Kirsch Extended
    \\
    ~
    \\

    $\begin{bmatrix}
        -1 & -1 & 0 & 1 & 1  \\
        -2 & -2 & 0 & 1 & 2  \\
        -3 & -6 & 0 & 6 & 3  \\
        -2 & -2 & 0 & 2 & 2  \\
        -1 & -1 & 0 & 1 & 1  \\
    \end{bmatrix}$
    &
    $\begin{bmatrix}
            -2 & -1 & 0 & 1 & 2  \\
            -2 & -1 & 0 & 1 & 2  \\
            -8 & -4 & 0 & 4 & 8  \\
            -2 & -1 & 0 & 1 & 2  \\
            -2 & -1 & 0 & 1 & 2  \\
    \end{bmatrix}$
    \\
        ~
    \\

        Scharr Extended
    &
        Orhei
\end{tabular}
}
\caption{$5x5$ kernels masks}
\label{fig:k5_kernel_masks}
%\end{minipage}
\end{figure}

\begin{figure}[!h]
%\begin{minipage}[c]{0.45\textwidth}
\centering
\tiny{
\begin{tabular}{ccccc}
        $\begin{bmatrix}
            0 &  1 & 0 \\
            1 & -4 & 1 \\
            0 &  1 & 0 \\
        \end{bmatrix}$
    &
        $\begin{bmatrix}
            1 &  1 & 1 \\
            1 & -8 & 1 \\
            1 &  1 & 1 \\
        \end{bmatrix}$
    &
        $\begin{bmatrix}
            -1 &  2 & -1 \\
             2 & -4 &  2 \\
            -1 &  2 & -1 \\
        \end{bmatrix}$
    % \\
    % ~
    % \\
    %     V1 & V2 & V3
    % \\
    &
        $\begin{bmatrix}
            1 &   4 & 1 \\
            4 & -20 & 4 \\
            1 &   4 & 1 \\
        \end{bmatrix}$
    &
        $\begin{bmatrix}
            2 & -1 & 2 \\
           -1 & -4 & 1 \\
            2 & -1 & 2 \\
        \end{bmatrix}$
    \\
        V1 & V2 & V3 & V4 & V5
    \\
\end{tabular}
}
\caption{Laplace discrete approximations masks 3x3}
\label{fig:laplace_kernel_masks}
%\end{minipage}
\end{figure}

\begin{figure}[!h]
%\begin{minipage}[c]{0.45\textwidth}
\centering
\tiny{
\begin{tabular}{ccccc}
    $\begin{bmatrix}
        0 & 0 & 1 & 0 & 0  \\
        0 & 1 & 2 & 1 & 0  \\
        1 & 2 & -17 & 2 & 1  \\
        0 & 1 & 2 & 1 & 0  \\
        0 & 0 & 1 & 0 & 0  \\
    \end{bmatrix}$
    &
    $\begin{bmatrix}
            1 & 1 & 1 & 1 & 1  \\
            1 & 1 & 1 & 1 & 1  \\
            1 & 1 & -18 & 1 & 1  \\
            1 & 1 & 1 & 1 & 1  \\
            1 & 1 & 1 & 1 & 1  \\
    \end{bmatrix}$
    \\
        V1 & V2 
    \\
\end{tabular}
}
\caption{Laplace discrete approximations masks 5x5}
\label{fig:laplace_kernel_masks}
%\end{minipage}
\end{figure}

\begin{figure}[!h]
%\begin{minipage}[c]{0.45\textwidth}
\centering
\tiny{
\begin{tabular}{cc}
    $\begin{bmatrix}
         -780 &  -720 &  -468 & 0 &  468 &  720 &  780 \\
        -1080 & -1170 &  -936 & 0 &  936 & 1170 & 1080 \\
        -1404 & -1872 & -2340 & 0 & 2340 & 1872 & 1404 \\
        -1560 & -2340 & -4680 & 0 & 4680 & 2340 & 1560 \\
        -1404 & -1872 & -2340 & 0 & 2340 & 1872 & 1404 \\
        -1080 & -1170 &  -936 & 0 &  936 & 1170 & 1080 \\
         -780 &  -720 &  -468 & 0 &  468 &  720 &  780 \\
    \end{bmatrix}$
    &
    $\begin{bmatrix}
        -3 & -2 & -1 & 0 & 1 & 2 & 3 \\
        -3 & -2 & -1 & 0 & 1 & 2 & 3 \\
        -3 & -2 & -1 & 0 & 1 & 2 & 3 \\
        -3 & -2 & -1 & 0 & 1 & 2 & 3 \\
        -3 & -2 & -1 & 0 & 1 & 2 & 3 \\
        -3 & -2 & -1 & 0 & 1 & 2 & 3 \\
        -3 & -2 & -1 & 0 & 1 & 2 & 3 \\
    \end{bmatrix}$
    \\
    ~
    \\
        Sobel Extended
    &
        Prewitt Extended
    \\
\end{tabular}
}
\caption{7$x$7 kernels masks}
\label{fig:k7_kernel_masks}
%\end{minipage}
\end{figure}

\begin{figure}[!h]
%\begin{minipage}[c]{0.45\textwidth}
\centering
\tiny{
\begin{tabular}{ccccc}
        $\begin{bmatrix}
            -1        & 0 & 1 \\
            -\sqrt{2} & 0 & \sqrt{2} \\
            -1        & 0 & 1 \\
        \end{bmatrix}$
    &
        $\begin{bmatrix}
            -1 & -\sqrt{2} & -1 \\
             0 &         0 & 0  \\
            1 &   \sqrt{2} & 1  \\
        \end{bmatrix}$
    &
        $\begin{bmatrix}
            0        &  -1 & \sqrt{2} \\
            1        &   0 & -1       \\
            \sqrt{2} &   1 & 0        \\
        \end{bmatrix}$
    &
        $\begin{bmatrix}
            \sqrt{2} & -1 & 0        \\
                  -1 &  0 & 1        \\
                   0 &  1 & \sqrt{2} \\
        \end{bmatrix}$
    &
        $\begin{bmatrix}
             0 & 1 &  0 \\
            -1 & 0 & -1 \\
             0 & 1 &  0 \\
        \end{bmatrix}$

    \\
    ~
    \\
        $G_1$
        &
        $G_2$
        &
        $G_3$
        &
        $G_4$
        &
        $G_5$

    \\
            $\begin{bmatrix}
            -1 & 0 &  1 \\
             0 & 0 &  0 \\
             1 & 0 & -1 \\
        \end{bmatrix}$
        &
            $\begin{bmatrix}
             1 & -2 &  1 \\
            -2 &  4 & -2 \\
             1 & -2 &  1 \\
        \end{bmatrix}$
        &
        $\begin{bmatrix}
            -2 & 1 & -2 \\
             1 & 4 &  1 \\
            -2 & 1 & -2 \\
        \end{bmatrix}$
    &
        $\begin{bmatrix}
            1 & 1 & 1 \\
            1 & 1 & 1 \\
            1 & 1 & 1 \\
        \end{bmatrix}$
    \\
    ~
    \\
        $G_6$
        &
        $G_7$
        &
        $G_8$
        &
        $G_9$
\end{tabular}
}
\caption{Frei-Chen kernels \label{fig:frei_kernel_masks}}
\end{figure}
\vspace{-1cm}

\end{document}